\documentclass[a4paper,fleqn]{cas-sc}

\usepackage[numbers]{natbib}
\usepackage{amsmath}
\usepackage{amssymb}
\usepackage{multirow}
\usepackage{framed,color}
\definecolor{shadecolor}{rgb}{0.9,0.9,0.9}

\newcommand{\R}{\,{\mathbb R}}
\newcommand{\C}{\,{\mathbb C}}
\newcommand{\Z}{\,{\mathbb Z}}
\newcommand{\T}{\,{\mathbb T}}
\newcommand{\diag}{\mbox{diag}}
\newcommand{\supp}{\mbox{supp}}
\newcommand{\sign}{\mbox{sign}}
\newcommand{\asin}{\mbox{asin}}
\newcommand{\acos}{\mbox{acos}}
\newcommand{\bolda}{{\mathbf a}}
\newcommand{\boldd}{{\mathbf d}}
\newcommand{\boldf}{{\mathbf f}}
\newcommand{\boldg}{{\mathbf g}}
\newcommand{\boldq}{{\mathbf q}}
\newcommand{\bolds}{{\mathbf s}}
\newcommand{\boldx}{{\mathbf x}}
\newcommand{\boldy}{{\mathbf y}}
\newcommand{\boldA}{{\mathbf A}}
\newcommand{\boldE}{{\mathbf E}}
\newcommand{\boldF}{{\mathbf F}}
\newcommand{\boldG}{{\mathbf G}}
\newcommand{\boldH}{{\mathbf H}}
\newcommand{\boldI}{{\mathbf I}}
\newcommand{\boldItilde}{\tilde {\mathbf I}}
\newcommand{\boldL}{{\mathbf L}}
\newcommand{\boldM}{{\mathbf M}}
\newcommand{\boldN}{{\mathbf N}}
\newcommand{\boldP}{{\mathbf P}}
\newcommand{\boldQ}{{\mathbf Q}}
\newcommand{\boldS}{{\mathbf S}}
\newcommand{\boldU}{{\mathbf U}}
\newcommand{\boldphi}{\boldsymbol{\phi}}
\newcommand{\boldpsi}{\boldsymbol{\psi}}
\newcommand{\boldPhi}{\boldsymbol{\Phi}}
\newcommand{\boldPsi}{\boldsymbol{\Psi}}
\newcommand{\<}{\langle}
\renewcommand{\>}{\rangle}

\date{}

 \newenvironment{reorganized}{\color{black}}{\color{black}}\date{}

\begin{document}
\shorttitle{New Orthogonal Multiwavelet Filters ...}
\shortauthors{Kolev et~al.}
\title[mode=title]{New Orthogonal Multiwavelet Filters Derived
             By Matrix Spectral Factorization}
\author{Vasil Kolev}
\cormark[1]
\ead{vasil.kolev@iict.bas.bg, kolev_acad@abv.bg}
\address{Institute of Information and Communication Technologies, 
Bulgarian Academy of Sciences, Bl. 2, Acad. G. Bonchev St., 1113 Sofia, Bulgaria}

\author{Todor Cooklev}[orcid=0000-0002-4619-3579]
\address{Wireless Technology Center, Purdue University, Fort Wayne, IN 46805, USA}
\ead{cooklevt@pfw.edu}

\author{Fritz Keinert}[orcid=0000-0003-3116-967X]
\address{Department of Mathematics, Iowa State University, Ames, IA 50011, USA} 
\ead{keinert@iastate.edu}
\begin{abstract}
   The paper considers the construction of two new orthogonal
  multiwavelets with supercompact support by using the Fast Bauer's
  method for matrix spectral factorization on the matrix product filter
  of the orthogonal CL multiwavelet filter.  The new multiwavelets
  possess orthogonality, symmetry/antisymmetry, and one of them provides
  better coding and smoothness than other supercompact
  multiwavelets. 
             \begin{reorganized}
  The performance of the new multiwavelet filters in
  subband-based edge detection, grayscale and color image compression
  and 1D and 2D signal denoising is compared with the GHM, SA4, CL,
  Integer Haar and Alpert multifilters. The comparative analysis shows
  that new multiwavelets can provides better human visual measures, SSIM
  and MS-SSIM in image compression and denoising applications.
             \end{reorganized}
\end{abstract}
\maketitle
\section{Introduction}\label{sec:intro}
Multiwavelets, as extensions of scalar wavelets, can possess important
features such as orthogonality, compact support, symmetry and high-order
vanishing moments. Their structure makes them useful in mathematical
theory and many applications. Supercompact multiwavelets have support
$[0,1]$ and retain the spatial compactness and orthogonality of Haar
wavelets \cite{BW-00}. They are simple to implement in applications on a
finite interval and/or for analysis of signals that are piecewise
continuous, since no special treatment of external or internal
boundaries is necessary.  This compactness becomes more important where
strong gradients occur, such as near shock waves and in boundary
regions. Numerical evidence also illustrates that these supercompact
multiwavelets could be utilized without modification even near
boundaries \cite{LBW-01}.

Supercompact multiwavelets are orthogonal vectors of $r$ scaling
functions $\boldphi(t) = [\phi_0, \phi_1, \ldots, \phi_{r-1}]^*$ and
$r$ wavelet functions $\boldpsi(t) = [\psi_0, \psi_1, \ldots,
  \psi_{r-1}]^*$, which belong to $L_2([0,1])^r$ \cite{SN-96} and
satisfy matrix recursion equations with only two coefficients
\begin{align*}
  \boldphi(t) &= \sqrt{2} \left[ H_0 \boldphi(2t) + H_1 \boldphi(2t-1)
    \right], \\
  \boldpsi(t) &= \sqrt{2} \left[ G_0 \boldphi(2t) + G_1 \boldphi(2t-1)
    \right].
\end{align*}
The multiscaling symbols $H(z) = H_0 + H_1 z^{-1}$ and $G(z) = G_0
+ G_1 z^{-1}$ are linear matrix polynomials.

The orthogonality conditions $\< \phi_i, \phi_j \> = \< \psi_i, \psi_j
\> = \delta_{ij}$ and $\< \phi_i, \psi_j \> = 0$ for $i, j = 0:r-1$,
can be expressed in terms of the coefficients as
\begin{equation*}
  H_0 H_0^* + H_1 H_1^* = G_0 G_0^* + G_1 G_1^* = I, \quad H_0 G_0^* +
  H_1 G_1^* = 0.
\end{equation*}

The Hermitian transpose of a matrix polynomial $A(z) = \sum_j A_j z^j$
is given by $A(z)^* = \sum_j A_j^* z^{-j}$.  The matrix product filter
$P(z)$ is defined as the autocovariance function
\begin{equation}
  P(z) = H(z) H(z)^* = P_{1} z^{-1} + P_0 + P_{-1} z,
\end{equation}
which is a positive semidefinite polynomial matrix. The matrix product
coefficients are
\begin{equation}\label{eq:pc}
  P_0 = H_0 H_0^* + H_1 H_1^*, \quad P_1 = H_0 H_1^*, \quad P_{-1} =
  P_1^* = H_1 H_0^*.
\end{equation}
It is Hermitian, and satisfies the half band condition $P(z) + P(-z) =
2I$ \cite{CNKS-96}.

The stability and causality conditions translate to requiring that
$P(z)$ has all its poles and zeros inside the unit circle.  {\em Matrix
  Spectral Factorization} (MSF) refers to finding a matrix spectral
factor $H(z)$, given $P(z)$. The existence of such a factor is
guaranteed under a condition on $P(z)$ due to Paley and Wiener,
requiring that $\log(\det(P(z)))$ have a finite integral over an interval of
length $2\pi$.  MSF is a key step for the construction of minimum phase
spectral factors from matrix product filters. Unlike the spectral
factorization of scalar Hermitian polynomial functions, where an
explicit formula exists, there is no explicit expression for matrix
spectral factorization. However, for low-degree para-Hermitian
polynomial matrix polynomials a fast algorithm has been constructed in
\cite{KCK-18,KCK-24}.

\begin{figure}
\centering 
  \includegraphics[width=6in]{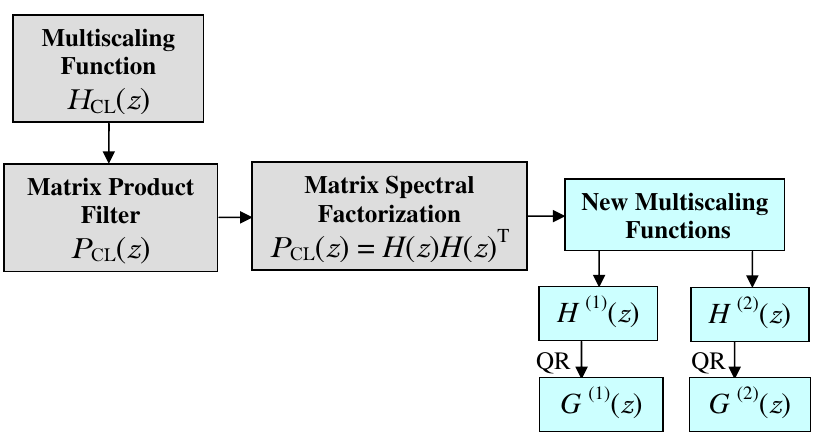}
  \caption{The main steps in the construction of new supercompact multiscaling and multiwavelet functions
  }
  \label{fig:01}
\end{figure}

Our main goal in this paper is to present a new approach for the
construction of new supercompact orthogonal multiwavelets by matrix
spectral factorization of a suitable product filter. The main steps are
shown in Fig.~\ref{fig:01}.

% Figure 1

The new multiwavelets are obtained from the product filter of the
Chui-Lian (CL) multiwavelet filter \cite{CL-96}. This immediately leads
to three main problems:

\begin{enumerate}
  \item Does the Fast Bauer’s method for matrix spectral factorization 
    converge in this case? 
  \item Is possible to obtain an orthogonal multiscaling filter with
    exact coefficients from existing multifilters?  
  \item Do the newly derived multiwavelets have better coding gain than
    other well-known orthogonal or biorthogonal multifilters? 
\end{enumerate}
    
We have the following main contributions:  
\begin{itemize}
  \item Obtain the matrix product filter of the CL orthogonal multiwavelet filter;
  \item Apply Fast Bauer’s method for spectral factorization to find 
        two new supercompact multiscaling functions with exact recursion coefficients;
  \item Find the complementary new supercompact multiwavelet functions;
  \item Compare the performance of new multiwavelet filters in subband-based edge detection, 
        1D signal denoising and grayscale and color image denoising and compression 
        with with the Integer Haar, Alpert, GHM, SA4 and CL multiwavelet filter banks. 
\end{itemize}

%\newpage

\section{Theory of Matrix Spectral Factorization}\label{sec:msf}

Spectral factorization is an important mathematical tool in signal processing, control theory, and other areas. The fundamental theorem of
the spectral factorization is the Fej{\'e}r-Riesz theorem for positive definite functions. The terminology comes from prediction theory, where
the nonnegative function plays the role of a spectral density for a multidimensional stationary stochastic process. Fej{\'e}r \cite{F-16}
first showed the importance of the class of trigonometric polynomials that admit only positive real values; as a theorem it was proved by
Riesz \cite{R-16}. The definition of a trigonometric univariate polynomial is
\begin{equation*}
  \nu(z) = \sum_{k=-N}^N \nu_k z^k.
\end{equation*}
When the function $\nu(z)$ is real for all $z \in \T$, where $\T$ is 
the unit circle in the complex plane, then the coefficients satisfy
$\overline\nu_k = \nu_{-k}$ for all $k$. If $\nu(z) \ge 0$ for 
all $z \in \T$, the {\em spectral factorization} of the function is
\begin{equation*}
  \nu(z) = p(z) p(z)^*,
\end{equation*}
where $p(z) = \sum_{k=0}^N p_k z^k$ is called a {\em scalar spectral factor}.

{\em Matrix spectral factorization} (MSF) refers to finding 
a matrix spectral factor $H(z)$, given $P(z)$, so that
\begin{equation*}
  P(z) = H(z) H(z)^*.
\end{equation*}
Such a factorization exists if $P(z)$ is positive definite on $\T$,
absolutely integrable and satisfies the Paley-Wiener condition
\cite{PW-33,W-55,SK-01}
\begin{equation*}
  \left| \int_{-\pi}^\pi \ln \det P(e^{i\omega}) \,d\omega \right| < \infty.
\end{equation*}

There is a unique minimal phase solution $H_M(z)$. All other solutions are of the form
\begin{equation*}
  H(z) = H_M(z) U,
\end{equation*}
where $U$ is a unitary matrix multiplier.

Usually, $P(z)$ is a half-band filter with $P_0 = I$, $P_{2k} = 0$ for
$k \ne 0$.

\subsection{Classical Bauer’s method of Matrix Spectral Factorization}
Let the multiscaling function $H(z)$ and its complex conjugate $H^*(z)$ be
\begin{equation}
   \begin{split}
     H(z) = H_0 + H_1 z^{-1} + \cdots + H_m z^{-m}, \\
     H^*(z) = H_0^* + H_1^* z + \cdots + H_m^* z^{m},
   \end{split}
\end{equation}
where $H_k$ are real matrices of size $r \times r$ and $z \in \C$. Let
$\T=\{z: |z| = 1\}$ be the unit circle in the complex plane. For $z \in \T$
we have $z^{-1} = \overline{z}$. Then the matrix product
\begin{equation}
  P(z) = H(z) H^*(z) = P_m^* z^{-m} + \cdots P_1^* z^{-1} + P_0 + P_1 z +
  \cdots P_m z^m
\end{equation}
is called the {\em matrix product filter}, where each matrix coefficient is determined by
\begin{equation}
  P_k = H_{m-k} H_m^* + H_{m-k-1} H_{m-1}^* + \cdots + H_0 H_k^*.
\end{equation}
Supercompact multifilters represent the simplest case $m=1$, with the coefficients $P_0$ and $P_1$ shown in~\eqref{eq:pc}.

Bauer’s matrix spectral factorization finds the Cholesky factor $L_n$ of the $(n+1) \times (n+1)$ banded block Toeplitz matrix
\begin{equation}\label{eq:5}
  T_n =
  \begin{pmatrix}
    P_0 & P_1 & 0 & 0 & 0 \\
    P_1^* & P_0 & P_1 & 0 & 0 \\
    0 & \ddots & \ddots & \ddots & 0 \\
    0 & 0 & 0 & P_1^* & P_0
  \end{pmatrix} = L_n L_n^*,
\end{equation}
where
\begin{equation*}
  L_n =
  \begin{pmatrix}
    H_0^{(0)} & 0 & 0 & 0 \\
    H_1^{(1)} & H_0^{(1)} & 0 & 0 \\
    0 & \ddots & \ddots & 0 \\
    0 & 0 & H_1^{(n)} & H_0^{(n)} 
  \end{pmatrix}
\end{equation*}
is a lower bidiagonal matrix.

The last row of $L_n$ converges to the true matrix coefficients $H_0^{(n)} \to H_0$, $H_1^{(n)} \to H_1$ as $n \to \infty$. 

The classical Bauer's method is very effective for nonsingular matrix product filters \textit{P}(z). 
Convergence can be very fast, so n does not need to be large to accurately estimate the spectral factors. 

	However, since in our applications the matrix product filters are required to possess multiple zeros on the unit circle,
the method converges very slowly and requires a very large Toeplitz matrix. 
In one example, it required a matrix of size around $10^6\times{10^6}$ for achieving a precision of $10^{-13}$. 
	
	In addition, for larger multiplicity \textit{r}$\geqslant{3}$ high precision often cannot be achieved. 
In one example in \cite{KCK-24} with \textit{r}=5, \textit{m}=1 and a decuple zero singularity,
the maximum achievable accuracy in the classical Bauer's method was around $10^{-3}$.

\subsection{Fast Bauer’s method of Matrix Spectral Factorization}

Obviously the Cholesky decomposition \eqref{eq:5} can be realized row by row. 
If we assume that $H_0^{(n)}$, $H_1^{(n)}$ have already been found,
with the initial value $H_0^{(0)}$ obtained as the Cholesky factor of $P_0$, then it follows that 
\begin{equation}\label{eq:7}
  H_1^{(n+1)} = P_1^* \left[ H_0^{(n)} \right]^{-*}
\end{equation}
from $P_1^* = H_1^{(n+1)} \left[ H_0^{(n)} \right]^*$, and
\begin{equation}\label{eq:8}
  H_0^{(n+1)} \left[ H_0^{(n+1)} \right]^* = P_0 - H_1^{(n+1)}  \left[ H_1^{(n+1)}  \right]^*
\end{equation}
from $P_0 = H_0^{(n+1)} \left[ H_0^{(n+1)} \right]^* + H_1^{(n+1)}  \left[ H_1^{(n+1)}  \right]^*$.

Then \eqref{eq:7} substituted into~\eqref{eq:8} leads to
\begin{equation}
  H_0^{(n+1)} \left[ H_0^{(n+1)} \right]^*  = P_0 - P_1^* \left[
    H_0^{(n)} \right]^{-*} \left[  H_0^{(n)} \right]^{-1} P_1.
\end{equation}
If we define $X^{(n+1)} = H_0^{(n+1)} \left[ H_0^{(n+1)} \right]^*$, then
\begin{equation}
  X^{(n+1)} = P_0 - P_1^* \left[  X_0^{(n)} \right]^{-1} P_1.
\end{equation}
The limit $n \to \infty$ leads to the solution $X^{(n+1)} \to X$ of the
nonlinear matrix equation (NME)
\begin{equation}\label{eq:11}
  X = P_0 - P_1^* X^{-1} P_1.
\end{equation}

Our focus in this paper is the supercompact case \textit{m}=1, but we note that the general case 
can be reduced to this by replacing the original coefficients with block matrices. For \textit{m}=3, 
for example, we set
\begin{equation*}
  \hat P_{-1} = \hat P_1^* =
  \begin{bmatrix}
    P_3^* & P_2^* & P_1^* \\[3pt]
    0 & P_3^* & P_2^* \\[3pt]
    0 & 0 & P_3^*
  \end{bmatrix}, \qquad
  \hat P_0 =
  \begin{bmatrix}
    P_0 & P_1 & P_2 \\[2pt]
    P_1^* & P_0 & P_1 & \\[2pt]
    P_2^* & P_1^* & P_0 
  \end{bmatrix}, \qquad
  \hat P_1 =
  \begin{bmatrix}
    P_3 & 0 & 0 \\
    P_2 & P_3 & 0 \\
    P_1 & P_2 & P_3
  \end{bmatrix}.
\end{equation*}
The resulting coefficients $\hat H_0$, $\hat H_1$ have the form
\begin{equation*}
  \hat H_1 =
  \begin{bmatrix}
    H_3 & H_2 & H_1 \\
    0 & H_3 & H_2 \\
    0 & 0 & H_3
  \end{bmatrix}, \qquad
    \hat H_0 =
  \begin{bmatrix}
    H_0 & 0 & 0 \\
    H_1 & H_0 & 0 \\
    H_2 & H_1 & H_0
  \end{bmatrix}.
\end{equation*}
Once $\hat H_0$, $\hat H_1$ have been computed, we can extract the  
original $H_0$, $H_1$, $H_2$, $H_3$ as submatrices.  

	Setting up the symmetrical nonsingular matrix $X$ from the above Fast Bauer's method with 
symbolic entries $x_{ij}$ converts \eqref{eq:11} into a system of nonlinear equations. 
For small matrix sizes this can be solved with exact values by a computer algebra system (CAS).

	For larger \textit{r}, \eqref{eq:11} could be solved numerically by fixed point iteration, 
which is identical to the original Bauer's method, but Newton's method converges much faster 
and gives more accurate results. See \cite{KCK-24} for more details.

\section{Design of the supercompact multiwavelet filter}
\subsection{The matrix product filter of the CL multiscaling function}
The matrix coefficients of the orthogonal CL scaling functions
\cite{CL-96}
\begin{equation*}
  H_{\mbox{\scriptsize CL}}(z) = \frac{1}{4} \left[
  \begin{pmatrix}
    0 & 2 + \sqrt{7} \\
    0 & 2 - \sqrt{7}
  \end{pmatrix} +
  \begin{pmatrix}
    3 & 1 \\
    1 & 3
  \end{pmatrix} z^{-1} +
  \begin{pmatrix}
    2 - \sqrt{7} & 0 \\
    2 + \sqrt{7} & 0
  \end{pmatrix} z^{-2} \right]
\end{equation*}
lead to the matrix product filter
\begin{equation*}
  P_{\mbox{\scriptsize CL}}(z) = \frac{1}{8} \left[
    \begin{pmatrix}
      4 & 1 + \sqrt{7} \\
      -(1 + \sqrt{7}) & -\sqrt{7}
    \end{pmatrix} z^{-1} + 
    \begin{pmatrix}
      8 & 0 \\
      0 & 8
    \end{pmatrix} + 
    \begin{pmatrix}
      4 & -(1 + \sqrt{7}) \\
      1 + \sqrt{7} & -\sqrt{7}
    \end{pmatrix} z    \right]
\end{equation*}
with
\begin{equation*}
  \det(P_{\mbox{\scriptsize CL}}(z)) = (4 - \sqrt{7}) \frac{(1+z)^4}{32 z^2},
\end{equation*}
which satisfies the regularity criterion that $\det(P(z))$ has a
quadruple zero at $z=-1$.

\subsection{New supercompact multiscaling functions from the Exact Bauer’s method}
% Figure 2
\begin{figure}
\centering 
  \includegraphics[width=6in]{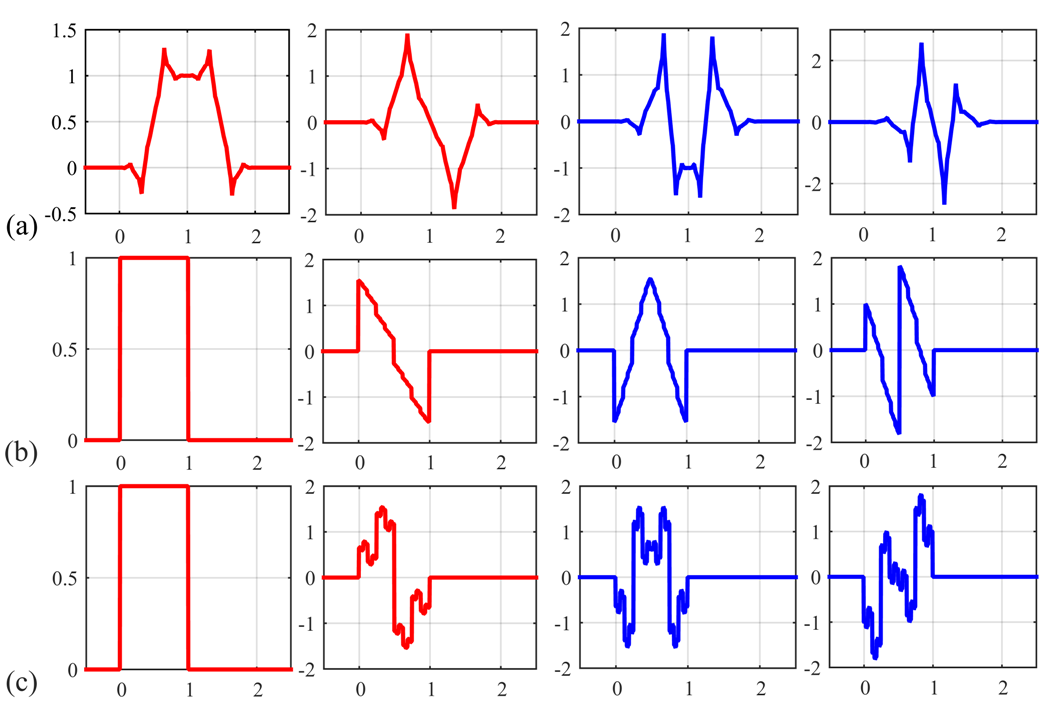} 
  \caption{(a) CL  multiscaling function  $\boldphi_{CL}$ (red) and
    multiwavelet function  $\boldpsi_{CL}$ (blue); (b) New supercompact
    multiscaling function $\boldphi^{(1)}$  (red) and multiwavelet
    function $\boldpsi^{(1)}$  (blue);  (c) New supercompact multiscaling function $\boldphi^{(2)}$   (red) and  multiwavelet
    function $\boldpsi^{(2)}$  (blue);
  }
  \label{fig:02}
\end{figure}

In order to obtain multiscaling functions with exact coeficients, we solve  eq.\eqref{eq:11} by an appropriate
CAS, for example the Matlab Symbolic Toolbox. The solution is a symmetric matrix 

\begin{equation*}
  X = \frac{1}{8}
  \begin{pmatrix}
    4 & \sqrt{7} + 1 \\
    \sqrt{7} + 1 & 4
  \end{pmatrix}.
\end{equation*}

	The standard Cholesky factor is

\begin{equation*}
  H_0^{(1)} = \frac{\sqrt{2}}{8}
  \begin{pmatrix}
          4      &      0      \\
    \sqrt{7} + 1 & \sqrt{8-2\sqrt{7}}
  \end{pmatrix}
\end{equation*}

The square root has two possible values
\begin{equation*}
       \sqrt{8-2\sqrt{7}} = \pm{(\sqrt{7} - 1)}
\end{equation*}

which leads to two solutions

\begin{equation*}
  H_0^{(1)} = \frac{\sqrt{2}}{8}
  \begin{pmatrix}
               4 &  0 \\
    \sqrt{7} + 1 & \sqrt{7} - 1
  \end{pmatrix}
\end{equation*}
and 
\begin{equation*}
  H_0^{(2)} = \frac{\sqrt{2}}{8}
  \begin{pmatrix}
    4 & 0 \\
    \sqrt{7} + 1 & -(\sqrt{7} - 1)
  \end{pmatrix}
\end{equation*}

The second coefficients $H_1^{(i)}$ is found from \eqref{eq:7} as
$H_1^{(i)} = P_1^* \left[ H_0^{(i)} \right]^{-*}$.  We call the
resulting supercompact multiscaling function {\bf\em New1} and {\bf\em
  New2}.

The resulting coeficients, as well as the coeficients of the
corresponding multiwavelet functions and the determinants of the symbols
are tabulated in Tab~\ref{tab:01}.  It is important to note that the
determinants of both multiscaling functions include a factor of
$(1+z^{-1})/2$, which according to \cite{CNKS-96} leads to regular
scaling functions and a desirable supercompact multiscaling function.

% Table 1
\begin{table}
\centering 
  \caption{The matrix coefficients of the two new multiscaling and multiwavelet functions {\bf\em New1} and {\bf\em New2}.} %
  {\label{tab:01}
    \begin{tabular}{c|c}
      \multicolumn{2}{c}{\ } \\
      \hline
      \hline
      \multicolumn{2}{c}{Multiscaling Functions} \\
%       \hline
      $H^{(1)}(z)$ & $H^{(2)}(z)$ \\
      \hline
      \hline
      $H_0^{(1)} = \frac{\sqrt{2}}{8}
      \begin{pmatrix}
        4 & 0 \\
        \sqrt{7} + 1 & \sqrt{7} - 1
      \end{pmatrix}$ &
      $H_0^{(2)} = \frac{\sqrt{2}}{8}
      \begin{pmatrix}
        4 & 0 \\
        \sqrt{7} + 1 & 1 - \sqrt{7}
      \end{pmatrix}$  \\
      $H_1^{(1)} = \frac{\sqrt{2}}{8}
      \begin{pmatrix}
        4 & 0 \\
        -(\sqrt{7} + 1) & \sqrt{7} - 1
      \end{pmatrix}$ &
      $H_1^{(2)} = \frac{\sqrt{2}}{8}
      \begin{pmatrix}
        4 & 0 \\
        -(\sqrt{7} + 1) & 1 - \sqrt{7}
      \end{pmatrix}$  \\[10pt]
      $\det(H^{(1)}(z)) = \det(H_0^{(1)} + H_1^{(1)} z^{-1})$ & 
      $\det(H^{(2)}(z)) = \det(H_0^{(2)} + H_1^{(2)} z^{-1})$ \\
      $= \frac{\sqrt{7} - 1}{\sqrt{2}} (1 + z^{-1})^2$ & 
      $= \frac{1 - \sqrt{7}}{\sqrt{2}} (1 + z^{-1})^2$  \\[5pt]
      \hline
      \hline
      \multicolumn{2}{c}{\ } \\
      \hline
      \hline
      \multicolumn{2}{c}{Multiwavelet Functions} \\
%       \hline
      $G^{(1)}(z)$ & $G^{(2)}(z)$ \\
      \hline
      \hline
      $G_0^{(1)} = \frac{\sqrt{2}}{8}
      \begin{pmatrix}
        0 & -4 \\
         -(\sqrt{7} - 1) & \sqrt{7} + 1
      \end{pmatrix}$ &
      $G_0^{(2)} = \frac{\sqrt{2}}{8}
      \begin{pmatrix}
        0 & -4 \\
        -(\sqrt{7} - 1) & -(\sqrt{7} + 1)
      \end{pmatrix}$  \\
         $G_1^{(1)} = \frac{\sqrt{2}}{8}
      \begin{pmatrix}
        0 & 4 \\
        \sqrt{7} - 1 & \sqrt{7} + 1
      \end{pmatrix}$ &
      $G_1^{(2)} = \frac{\sqrt{2}}{8}
      \begin{pmatrix}
        0 & 4 \\
        \sqrt{7} - 1 & -(\sqrt{7} + 1)
      \end{pmatrix}$  \\[10pt]
      $\det(G^{(1)}(z)) = \det(G_0^{(1)} + G_1^{(1)} z^{-1})$ & $\det(G^{(2)}(z)) = \det(G_0^{(2)} + G_1^{(2)} z^{-1})$ \\
      $= \frac{\sqrt{7} - 1}{\sqrt{2}} (1 - z^{-1})^2$ & $ = \frac{\sqrt{7} - 1}{\sqrt{2}} (1 - z^{-1})^2$   \\[5pt]
      \hline
      \hline
    \end{tabular}}{}
\end{table}
\subsection{New supercompact multiwavelet functions}
In order to obtain the wavelet matrix coefficients $G_0$ and $G_1$ we
can employ the QR decomposition as in \cite{KCK-18}, or in the present
case, the orthogonal vectors method. This multiwavelet possesses the all
the desirable properties of the Alpert multiwavelet. The recursion
coefficients of the new orthogonal multiscaling and multiwavelet
functions obtained by QR decomposition are shown in Table \ref{tab:01};
the functions are shown in Fig.~\ref{fig:02}.
\section{Performance Analysis}
\subsection{Numerical errors}
\begin{figure}
\centering 
  \includegraphics[width=4in]{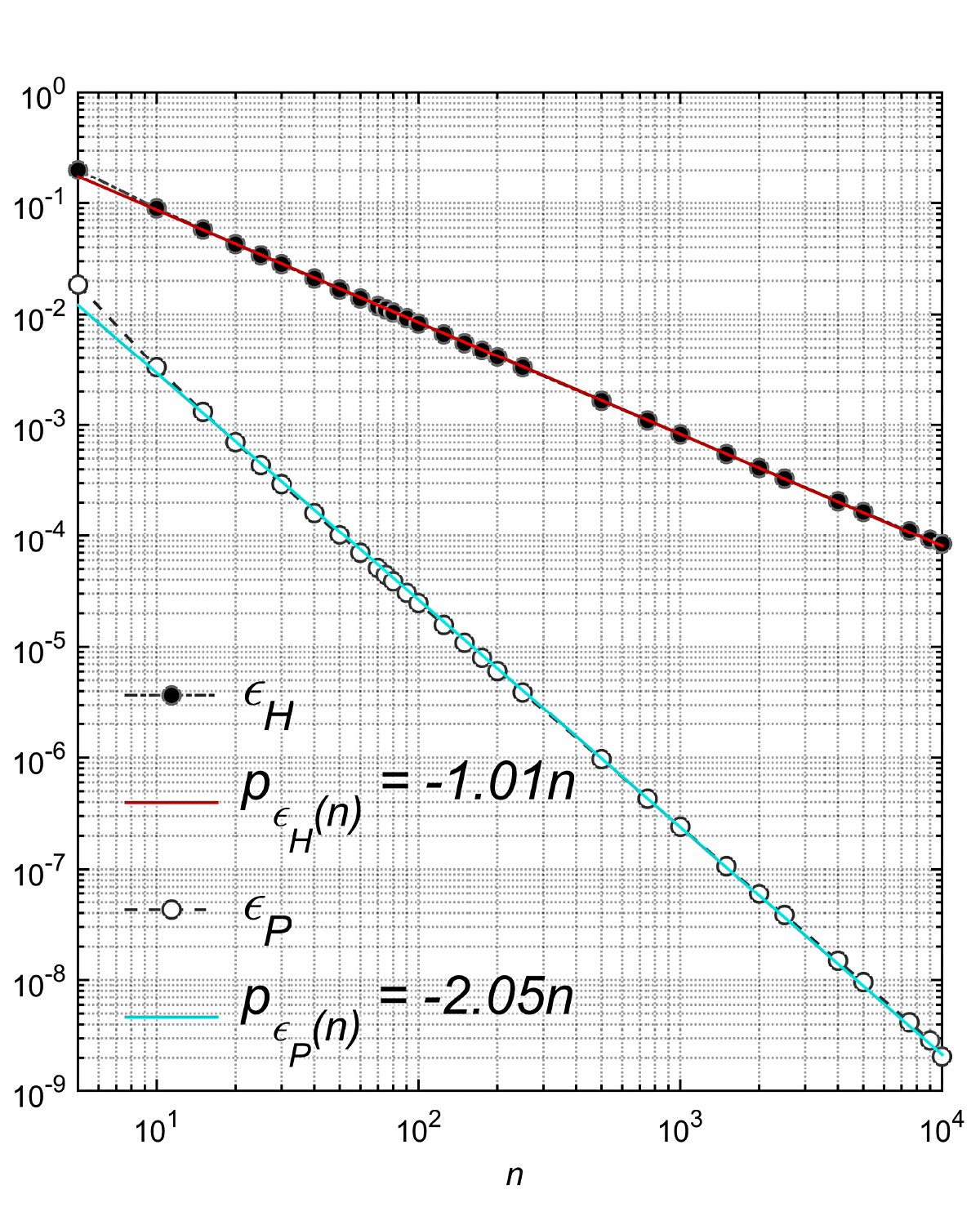} \\
  \caption{Influence of the size $n$ of the Toeplitz matrix on the precision 
of the matrix spectral factor (in log-log format): 
The numerical error of matrix spectral factor $\epsilon_H^{(n)}$(red) 
and of the matrix product filter $\epsilon_P^{(n)}$(cyan) }
  \label{fig:03}
\end{figure}

The classical Bauer’s method is applied to the spectral factorization of
the matrix product filter. The numerical errors obtained for the matrix
product filter $P(z)$ and the matrix spectral factor $H(z)$ are defined
as
\begin{equation}
   \begin{split}
     \epsilon_H^{(n)} = \left| H - H^{(n)} \right| \\
     \epsilon_P^{(n)} = \left| P - P^{(n)} \right|.
   \end{split}
\end{equation}
The values are shown in Fig.~\ref{fig:03}, in a log-log plot.  The
convergence of the two errors $\epsilon_H$ and $\epsilon_P$ is well
modeled by the linear fit $p_{\epsilon_H(x)} = -1.0101n$ and
$p_{\epsilon_P(x)} = -2.0454n$ respectively, where the logarithm of the
error $\log(\epsilon_H^{(n)})$ has slope $(-1)$ (a red line), while
$\log(\epsilon_H^{(n)})$ has a slope of $(-2)$ (cyan line).

In \cite{KCK-24}, after applying the fast Bauer’s method using a general
discrete algebraic Riccati equation (GDARE) in Matlab R2018a, the best
achieved precision for the matrix spectral factor
$\epsilon_H^{(n)} = 6.16 \times 10^{-5}$ and the matrix product filter
$\epsilon_P^{(n)} = 6.66 \times 10^{-14}$, while with Matlab R2010a,b
the best precisions are $\epsilon_H^{(n)} = 6.7077 \times 10^{-5}$ and
$\epsilon_P^{(n)} = 3.3307 \times 10^{-16}$, respectively.

It is important to note that the classical Bauer's method for spectral
factorization obtained errors for $n = 10,000$ of
$\epsilon_H^{(n)}= 8.396 \times 10^{-5}$ and
$\epsilon_P^{(n)} = 2.055 \times 10^{-9}$ shown in
Fig.~\ref{fig:03}. They can easily be modeled by regression
functions. The empirical results show minimal possible errors
$\epsilon_H^{(n)} = 9.6145 \times 10^{-17}$ and
$\epsilon_H^{(n)} = 6.979 \times 10^{-5}$ that can be achieved for
$n = 38.639$.
% Figure 3
\subsection{Statistical Characteristic Analysis of the Test Images}
\subsubsection{Histograms}
The histogram of an image is a one-dimensional statistical
characteristic of the given image and represents the frequency of
intensity in separate groups of pixels in the image. The groups usually
contain the same number of adjacent intensity levels.  The most
representative (precise) are histograms on all possible single levels.

As an application, the histogram values are in some (engineering) range
of intensities of useful (measured) quantities, of course appropriately
quantized. Visually, the histogram allows for different interpretations
by pixels and groups of pixels, for example how the brightness
(luminance) correlates with the number of pixels in it, as well as some
easily apparent characteristics such as contrast, dynamic range and
saturation effects, which make problems easily visible and facilitate
the analysis of the image (and the problem it reflects).

The color of images is represented by color models. The most popular of
them is RGB, but we can also mention many others, YUV, HSV, etc.,
emphasizing different characteristics of color. While a halftone image
is a scalar function over the discrete area of pixels, color images are
usually vector functions. RGB is a 3-dimensional function, one vector
for each of the 3 channels (all halftones).

For better decomposition at multiple levels of detail and better
encoding as well, in the future we could apply a suitable 3D transform
based on 3D convolution and extraction of color channels by a
\textit{inflated 3D convNet (I3D) model} \cite{huang2023fusion} instead
of extraction of RGB channels.
	
\begin{figure}
\centering 
  \includegraphics[width=6in]{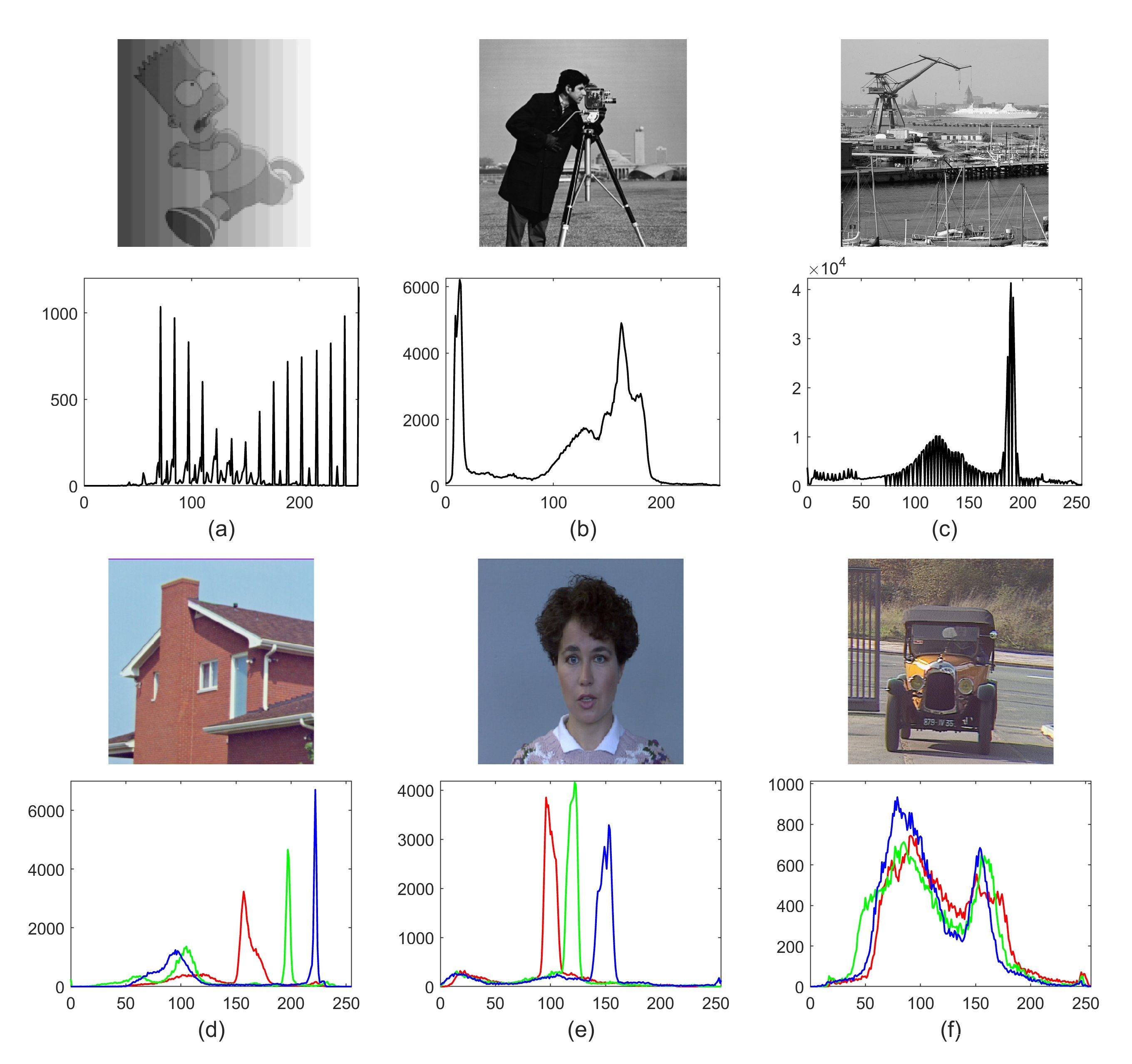}
  \caption{Test images and their histograms;
(a){\tt Bart Run}($128\times128$); (b){\tt Cameraman}($256\times256$);
(c){\tt Kiel}($512\times512$);(d){\tt House}($256\times256$); 
(e){\tt Cara}($256\times256$);(f){\tt Voit}($256\times256$).
}
  \label{fig:04}
\end{figure}

\begin{figure}
\centering 
  \includegraphics[width=6in]{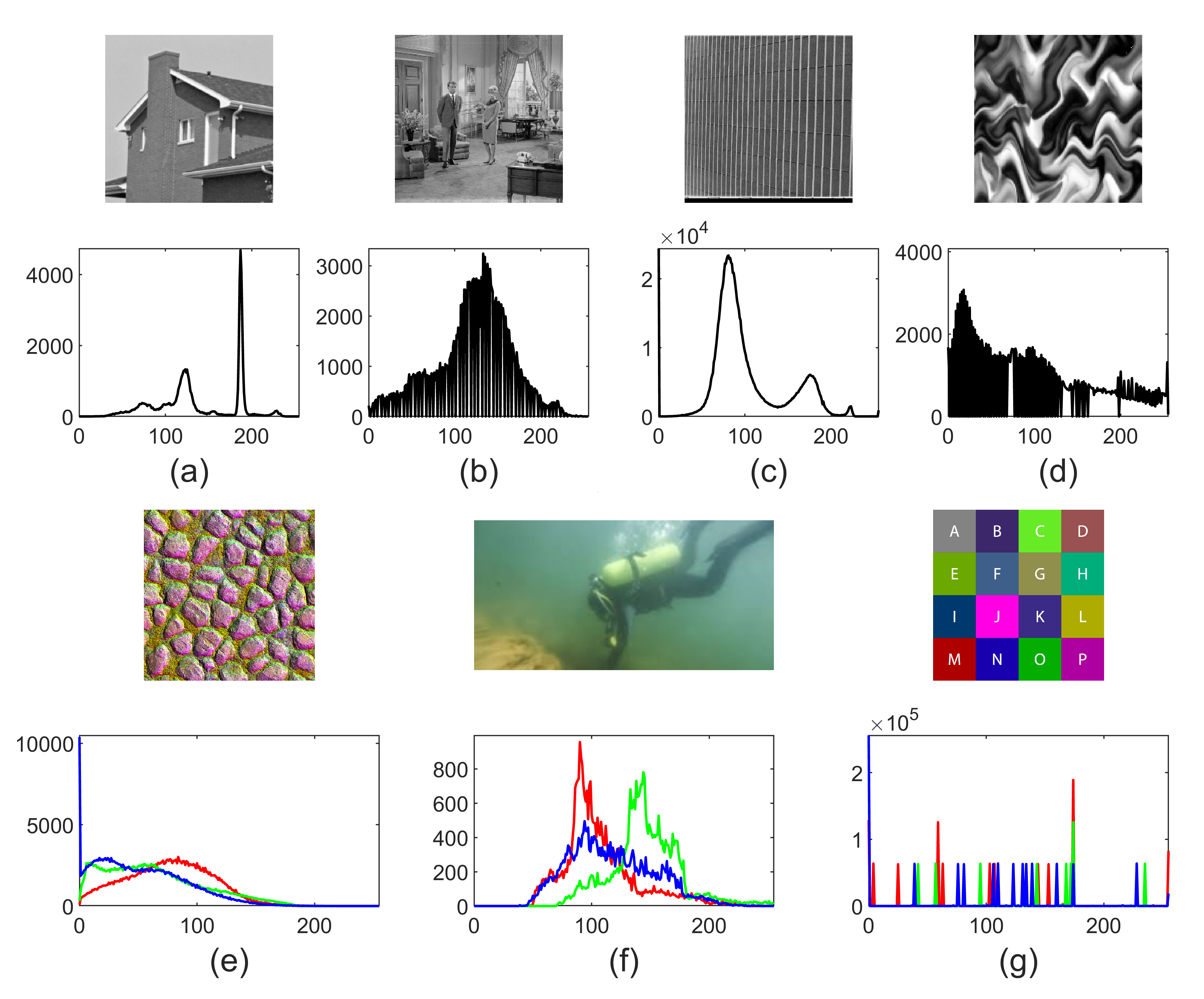}
  \caption{Test images and their histograms; (a){\tt House}($256\times256$); 
(b){\tt Couple}($256\times256$); (c){\tt Brick Wall}($1024\times1024$); 
(d) {\tt Texture 5} ($256\times256$); (e) {\tt Texture 1}($512\times512$); 
(f) {\tt Subwater}($256\times128$); (g){\tt Texture 4}($1024\times1024$).
 }
  \label{fig:05}
\end{figure}

The abscissa $Ox$ shows the pixel intensities in the range from 0 to 255
for each color channel (R, G, B). The value 0 corresponds to the "black"
level, and 255 to the "white" level. If the image is halftone,
successive levels of "gray" are placed between them. If the channel is
color, the "gray" level represents the corresponding base colors (R, G,
B) in the given color image.  The ordinate $Oy$ shows the frequency of
the corresponding pixel intensity levels.  Color histograms are triplets
of halftone histograms, one for each color (red (R), green (G), and blue
(B) channels. The size of the intervals in the figures is not specified
and the value of each pixel is displayed separately.  Some
characteristic tests for seven color and seven grayscale images with
different sizes and textures are illustrated in Fig.~\ref{fig:04}(a)–(f)
and Fig.~\ref{fig:05}(a)–(g).
	
\begin{itemize}
  \item {\tt Bart Run} has an extremely low dynamic range with only 14
    intensity values;
  \item {\tt Cameraman} has 2 isolated spikes, as can be see from the
    histogram in Fig.~\ref{fig:04}(b). The big peak on the far left
    side of the graph means that the image is made up of a lot of
    shadows or black tones. Therefore the image has low brightness. The
    peaks in the middle of the histogram indicate lots of midtones which
    mean muted color grays;
  \item {\tt Kiel} has distributed intensity values with 2 isolated
    peaks. The histogram in Fig.~\ref{fig:04}(c) shows peaks that are
    cut off on the left or right side, which means an underexposed
    image;
  \item {\tt House} is underexposed (intensity values are clustered) and
    has high contrast for all three colors. The blue channel, see
    Fig.~\ref{fig:04}(d), has many pixels in the high intensity region,
    while the red and green spectra remain mostly low. This means that
    the image has shades of blue, but not much red and green.
  \item {\tt Cara} has saturated and underexposed pixel values
    (intensity values are clustered) and has also an isolated peak for
    all three colors. This is shown in Fig.~\ref{fig:04}(e), where the
    histogram peaks in the middle of each color indicate a large
    presence of dark pixels for each color. Each color has peaks
    scattered around the middle, indicating a midtone, which means a
    muted color.
  \item {\tt Voit} has properly exposed normal contrast (intensity
    values are scattered).
% == Vasil Note 1 =========
% == Missed paragraphs ====
  \item {\tt House} has low contrast histogram by a slight increase in contrast. 
     The big peak on the right side shown in Fig.~\ref{fig:05}(a). 
  \item {\tt Couple} has low contrast histogram gaps caused by a slight increase in contrast shown in Fig.~\ref{fig:05}(b). 
  \item {\tt Brick wall} has a normal contrast and many pixels 
       in the low intensity region which means that the image 
        has shades or dark color tones shown in Fig.~\ref{fig:05}(c). 
  \item {\tt Texture 1}	has low high and overexposed contrast for 
        R, G and B channels respectively. The blue channel has many pixels 
        with in the lowest intensity region, while the red and green channels
        remain mostly low shown in Fig.~\ref{fig:05}(d). 
  \item {\tt Subwater} has normal contrast for channels,
        blur the red and blue channels and isolated peak shown in Fig.~\ref{fig:05}(e). 
  \item {\tt Texture 4} Texture 4 has extremely low dynamic range 
          for R, G, B channels with 2 isolated peaks shown in Fig.~\ref{fig:05}(f).
  \item {\tt Texture 5} histogram gaps caused by a slight increase in contrast shown in Fig.~\ref{fig:05}(g). 
% == END Vasil Note 1 =========
\end{itemize}
% Figure 4
% Figure 5

\subsubsection{Adjacent Pixel Correlations}
Neighboring pixel correlation refers to the relationship between two
adjacent pixels in horizontal and vertical coordinates, respectively.
The more the pixel values remain concentrated near the linear fit, the
stronger the correlation between neighboring pixels in the image;
conversely, the more pixels are scattered throughout the plot, the
weaker the correlation between neighboring pixels in the image.

\begin{figure}
\centering 
  \includegraphics[width=6in]{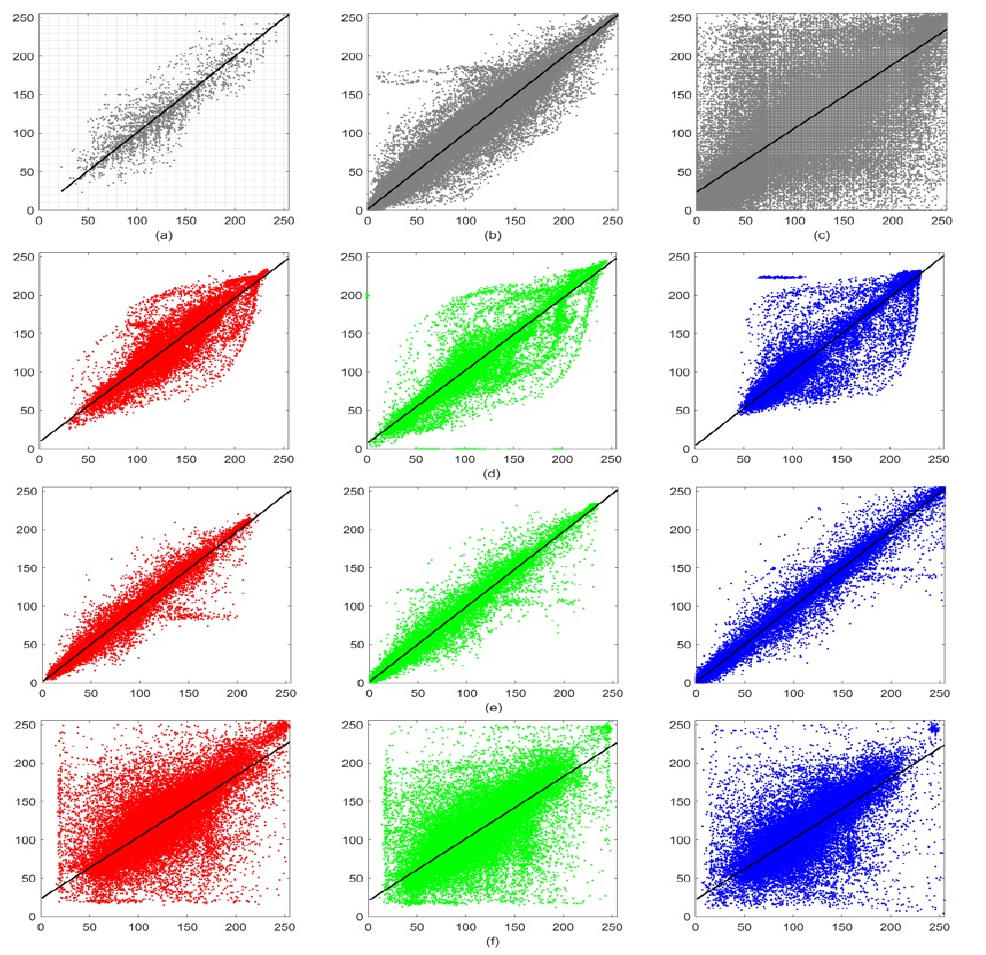}
  \caption{ Pixel correlations and linear fit lines for the gray and
    color (RGB channels) images; (a) {\tt Bart Run} (128x128) (b) {\tt
      Cameraman} (256x256) (c) {\tt Kiel} (512x512) (d) {\tt House}
    (256×256); (e) {\tt Cara} (256×256); (f) {\tt Voit} (256×256).  }
  \label{fig:06}
\end{figure}

The correlation coefficient $R^2$ is a primary measure of how well a
regression model fits the data. This correlation coefficient is simply
the square of Pearson’s correlation coefficient $R$.  To check the
existence of a high correlation among neighboring pixels in the test
images, a pixel correlation analysis test has performed. The correlation
analysis of adjacent pixels of this encryption scheme is shown in
Fig.~\ref{fig:06}(a)-(f). High correlation among pixels in the original
image is represented by the linear fit it converges to. The scatterplot
of Fig.~\ref{fig:06} leads to the values of $R^2$ for the regression
models shown in Table~\ref{tab:02}.

The correlation coefficient $R^2$, which is the square of the Pearson
correlation coefficient, is a basic measure of how well a regression
model fits the data; when $R^2=1$, the regression line predicts a
perfect fit of the pixel values.  The values of the regression lines for
all test images are shown in Table~\ref{tab:02}. Fig.~\ref{fig:06} shows
the scatter plot of all regression models from Table~\ref{tab:02}.

The images {\tt Bart Run} and {\tt Cara} have quite a bit of scatter,
but the scatter plot of Fig.~\ref{fig:06}(a) and (e) (red, green, and
blue channels) shows a clear linear trend with correlation
$88\%-98\%$. The images {\tt Kiel} and {\tt Voit} have a moderate
positive correlation in the interval $62\%-68\%$. It’s not a very strong
relationship, as some values show no correlation. Obviously, for these
two images there is quite a lot of scatter, but the trend is reasonably
linear, despite the fact that the big number of data points makes it
difficult to do a full correlation assessment.  Fig.~\ref{fig:05}(b) and
(f) (red, green, and blue channels) shows that there is no correlation
in some subset pixels and a strong correlation in the whole pixels data
set.

The images {\tt Bart Run} and {\tt Cara} have quite a large scatter, but
the scatter plot in Fig.~\ref{fig:06}(a) and (e) (RGB channels) shows a
clear linear trend with a correlation of $88\%-98\%$. The two images
{\tt Kiel} and {\tt Voit} have a moderate positive correlation in the
interval $62\%-68\%$. This is not a very strong relationship, as some
values show no correlation. It is clear that there is quite a bit of
scatter for these two images, but the trend is relatively linear,
despite the large number of data points, which makes it difficult to
fully assess the correlation. Fig.~\ref{fig:06}(b) and (f) (RGB channels)
show that there is no correlation in some subsets of pixels
and a strong correlation in the entire pixel dataset.

\begin{figure}
\centering 
  \includegraphics[width=6in]{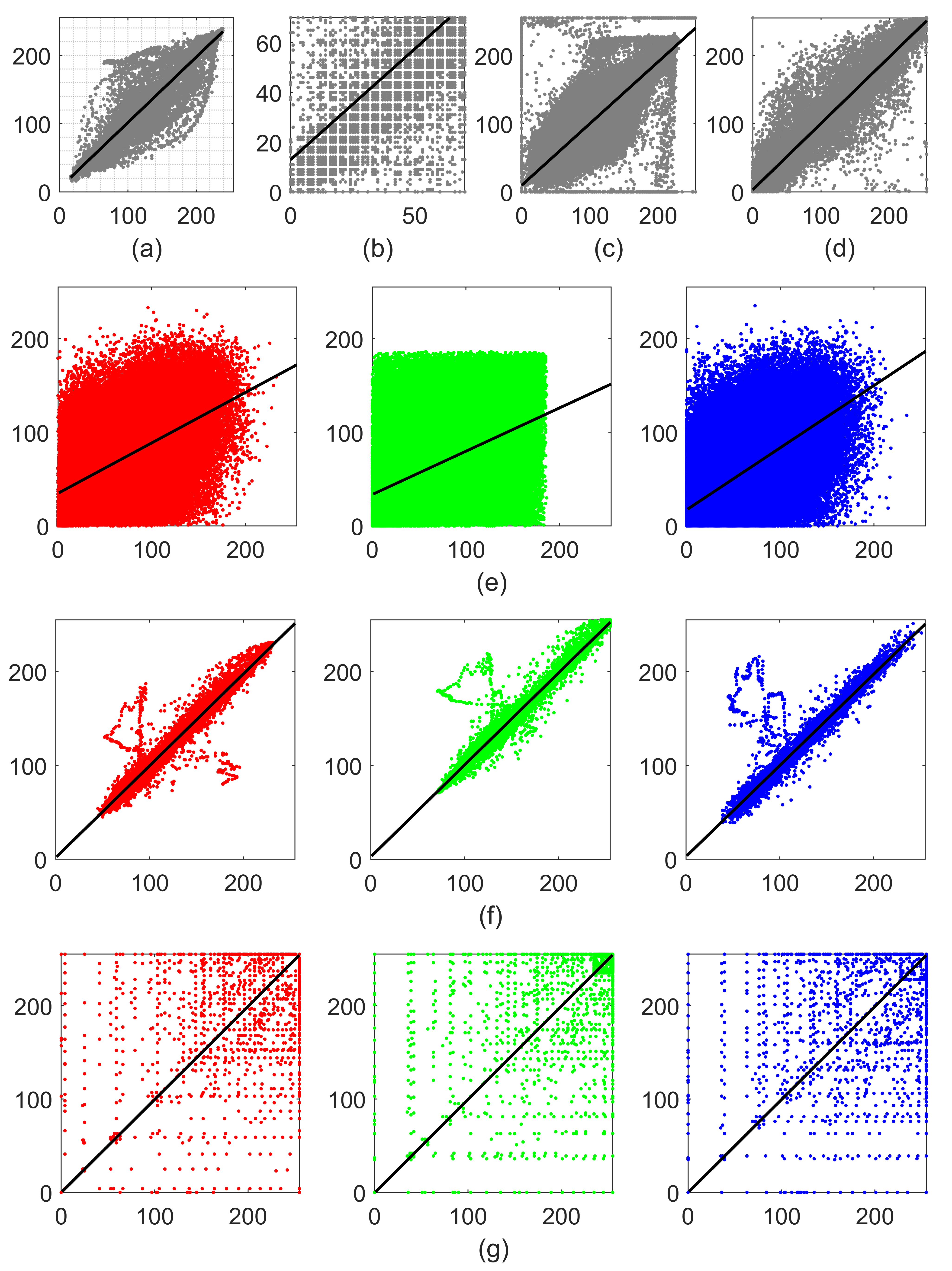}
  \caption{ Pixel correlations and linear fit lines for the gray and color (RGB channels) images; 
(a){\tt House} ($256\times256$); (b){\tt Couple}($256\times256$); (c){\tt Brick Wall}($1024\times1024$); 
(d){\tt Texture 5}($256\times256$); (e){\tt Texture 1}($512\times512$); 
(f){\tt Subwater}($256\times128$); (g){\tt Texture 4}($1024\times1024$).
 }
  \label{fig:07}
\end{figure}

% == Vasil Note 2 =========
% == Missed paragraphs ====
The grayscale images {\tt House},{\tt Cameraman}, {\tt Texture 5} and color image 
(for R,G,B channels){\tt Subwater} have a strong correlation shown of Fig.~\ref{fig:07}. 
The grayscale image {\tt House} and color images (for R,G,B channels) {\tt Texture 1}, {\tt Texture 4} 
there is no correlation of pixels and a strong correlation in the entire pixel dataset shown of Fig.~\ref{fig:07}.
% ===== End Vasil Note 2 ===========

% Figure 6

% Figure 7

% Table 2
\begin{table}
\centering 
  \caption{The coefficients of the linear regression lines for the gray and color images}
  {\label{tab:02}
    \begin{tabular}{ccccc}
      \hline
      \hline
      & &  \multicolumn{2}{c}{The coefficients of} \\
           \multicolumn{2}{c}{Images} & \multicolumn{2}{c}{the linear regression equation} & Coefficient of \\
      & &  \multicolumn{2}{c}                              {$y = p_1 x + p_2$}             & determination $R^2$ \\
      & & $p_1$ & $p_2$ \\
      \hline
      \hline
      {\tt Bart Run}  & & 0.9912 &  1.3412 & {\bf0.9824} \\
      {\tt Cameraman} & & 0.9887 &  1.3424 & {\bf 0.9774} \\
      {\tt Kiel}      & & 0.8271 & 23.5374 & 0.6841 \\
      {\tt House}     & & 0.9586 &  5.7160 & {\bf 0.9189}\\
      {\tt Couple}    & & 0.8927 &  12.9530 & 0.7970 \\
      {\tt Brick wall}& & 0.9053 &  9.5085 &  0.8195 \\
      {\tt Texture 5} & & 0.9727 &  2.9083 & {\bf 0.9461} \\
      \hline
				 & R & 0.9326 & 9.8814 & 0.8697 \\
	{\tt House}  & G & 0.9408 & 7.8723 & 0.8852 \\
                 & B & 0.9668 & 4.7158 & 0.9347 \\
       \hline
				 & R & 0.9763 & 2.2668 & {\bf 0.9531} \\
    {\tt Cara}   & G & 0.9797 & 2.1628 & {\bf 0.9598} \\
                 & B & 0.9861 & 1.8061 & {\bf 0.9723} \\
      \hline
				 & R & 0.7972 & 24.2278 & 0.6356 \\
    {\tt Voit}   & G & 0.8058 & 21.2176 & 0.6493 \\
                 & B & 0.7870 & 22.8056 & 0.6194 \\
    \hline
				 & R & 0.5361 & 35.3482 & 0.2874 \\
 {\tt Texture 1} & G & 0.4606 & 34.0091 & 0.2121 \\
                 & B & 0.6607 & 17.9257 & 0.4365 \\
     \hline
				 & R & 0.9761 & 2.5480 & {\bf 0.9527} \\
  {\tt Subwater} & G & 0.9762 & 3.4688 & {\bf 0.9530} \\
                 & B & 0.9689 & 3.6210 & {\bf 0.9387} \\
   \hline
				 & R & 0.9927 & 0.7649 & {\bf 0.9854} \\
 {\tt Texture 4} & G & 0.9947 & 0.5201 & {\bf 0.9894} \\
                 & B & 0.9923 & 0.7514 & {\bf 0.9846}\\
      \hline
      \hline
    \end{tabular}}{}
\end{table}

\subsection{General multifilter characteristics}
We compare the performance of the proposed new multifilters with some well-known 
orthogonal and biorthogonal filter banks.The results are given in Table~\ref{tab:03},
with the criteria being coding gain (CG), Sobolev smoothness (S) \cite{J-98}, 
symmetry/anti-symmetry (S/A) and length. 

% Table 3
\begin{table}
\centering 
  \caption{Comparison of CGs, S, S/A, for autocorrelation matrix $R_x$ of AR(1) and $\rho = 0.95$}%
  {\label{tab:03}
     \begin{tabular}{|c|c|c|c|c|}
       \hline
       \hline
       \multirow{1}{*}{Multifilter} & \multirow{1}{*}{CG} & \multirow{1}{*}{S} & \multirow{1}{*}{S/A} & \multirow{1}{*}{length}\\
       \hline
        \hline
       Biorthogonal Hermitian cubic spline &
       \multirow{2}{*}{1.51} &  \multirow{2}{*}{2.5} & \multirow{2}{*}{yes} & \multirow{2}{*}{3/5}\\
       ({\tt bih34n}) \cite{S-98a} & & & & \\
       \hline
       Biorthogonal $\left( \mbox{dual to Hermitian cubic spline} \right)	$ &
       \multirow{2}{*}{1.01} &  \multirow{2}{*}{2.5} &\multirow{2}{*}{yes} &  \multirow{2}{*}{3/5}\\
       ({\tt bih32s}) \cite{AZC-07} & & & & \\
       \hline
       Integer Haar \cite{RA-16} ({\em supercompact}) & 1.83 & 0.5 & yes & 2 \\
       \hline
       CL \cite{CL-96} & 2.06 & 1.06 & yes & 3 \\
       \hline
       Alpert \cite{SW-98} ({\em supercompact}) & 2.13 & 0.5 & yes & 2 \\
       \hline
        {\bf New2} ({\em supercompact}) & 2.24 & 1.28 & yes & 2 \\
       \hline 
       {\bf New1} ({\em supercompact}) & 2.38 & 1.28 & yes & 2 \\
       \hline
       SA4 \cite{STT-00} & 3.73 & 0.99 & yes & 4 \\
       \hline
       GHM \cite{GHM-94} & 4.41 & 1.5 & yes & 4  \\
       \hline
       \hline
   \end{tabular}}{}
 \end{table}

The CG is the ratio of arithmetic and geometric means of channel
variances and is a good indication of the performance in signal
processing. To estimate the CG, the variance is computed using a first
order Markov model AR(1) with intersample autocorrelation coefficient
$\rho = 0.95$ \cite{V-98}. As shown in Table~\ref{tab:03}, the
supercompact multiwavelets have smaller CGs than SA4 and
Geronimo-Hardin-Massopust (GHM), but better CGs, by at least a quarter
dB, than even CL and most other orthogonal and biorthogonal
multifilters, including designs with three or more matrix coefficients.

\subsection{Applications}
\subsubsection{Subband-based edge detection}
Edge detection is one of the fundamental operations in image processing
and computer vision. It is used to identify the edges in an image, for
image segmentation and data extraction, and for easy processing as well.

Since the choice of edge detection operator has to be responsive to
gradual changes, the Canny method is used, which provides good edge
detection. It uses two different threshold types to detect strong and
weak edges \cite{canny2009computational}.

The location and orientation of edges in an image can be determined from
the encoded subbands of the original image
\cite{mohsenian2002edge}. Note that in many edge detection algorithms,
the detail subbands which can be used in multiwavelet subband-based edge
detection are using vector quantization (VQ) via $k$-means-based deep
network architecture \cite{im2023vector} to encode the multiwavelet
coefficients. The main advantage of VQ is the construction of the code
book consisting of code vectors. The goal of quantization is to obtain
sufficiently small multiwavelet coefficients with a quantized
distribution so that they can be encoded at a low bit rate.  This
motivated us to consider a simple example of a multiwavelet
subband-based edge detection.

\begin{figure}
\centering 
  \includegraphics[width=6.5in]{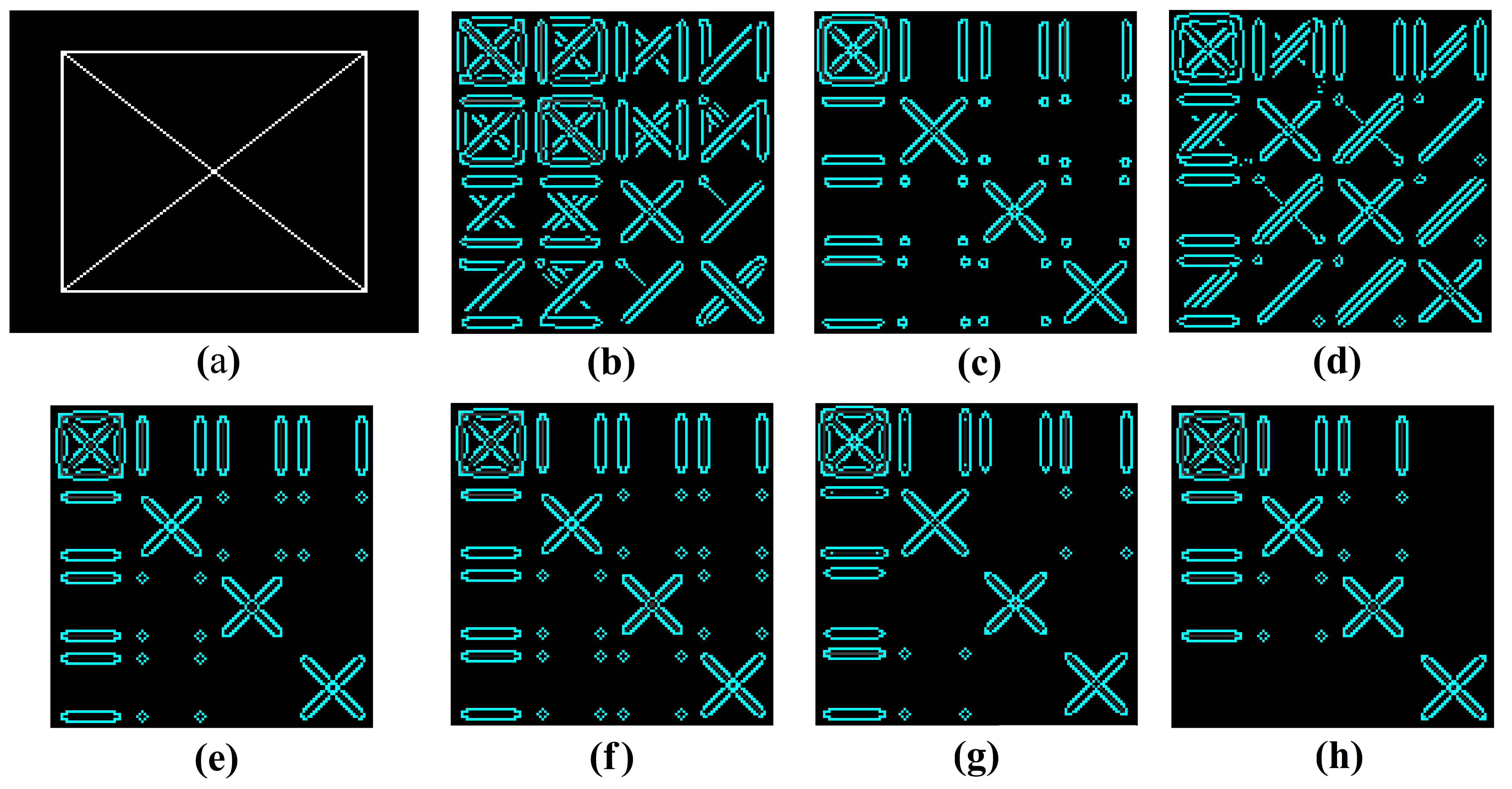}
  \caption{Subband-based edge detection for a one decomposition level
    of following orthogonal transforms by using the Canny method: (a)
    Original image; (b) GHM multiwavelet; (c) SA4 multiwavelet; (d) CL
    multiwavelet; (e) Alpert multiwavelet; (f) New1 multiwavelet; (g)
    New2 multiwavelet; (h) Integer Haar multiwavelet. 
}
  \label{fig:08}
\end{figure}

The Canny method is applied to the present multiwavelet filters for one
decomposition level of a simple image, where we obtain sixteen
subbands. The results from the subband-based edge detection are shown in
Fig.~\ref{fig:08}. The contour lines of the SA1 filter show that this is
an excellent detector, better than the other six transformations.

% Figure 8

\subsubsection{Image Compression}
The measurement for comparative analysis between the original
$\boldI(x,y)$ and the noise $\boldItilde(x,y)$ images of size $N^2$ is
the mean square error (MSE),
\begin{equation*}
  MSE =\frac{1}{N^2} \sum_{(x,y)} (\boldI - \boldItilde)^2,
\end{equation*}
which in decibels on a logarithmic scale is the Peak Signal-to-Noise Ratio (PSNR)
\begin{equation*}
  PSNR = 10\log_{10}(\frac{255^2}{MSE}) 
\end{equation*}
for $x,y=1:N$. PSNR of the compression ratio (CR) is defined as the
ratio between the original and compressed image size. Results are shown
in Table~\ref{tab:05}.

The structural information of natural images is based on the human
visual system (HVS). Quality assessment of compressed images is obtained
through the structural similarity index measures (SSIM) \cite{RH-08} and
its multiscale extension, the multiscale structural similarity index
(MS-SSIM) \cite{WSB-03}, where the objective quality evaluation at each
image scale contributes to the overal scorel.

The SSIM metric combines local image structure, luminance, and contrast
into a single local quality score. In this metric, structures are
patterns of pixel intensities, especially among neighboring pixels,
after normalizing for luminance and contrast. Because the human visual
system is good at perceiving structure, the SSIM quality metric agrees
more closely with the subjective quality score. The SSIM perceived
quality by quantifying the SSIM between an image and a reference image
\cite{RH-08}:
\begin{equation*}
  SSIM({\mathbf I},\tilde {\mathbf I}) = m_K({\mathbf I},\tilde {\mathbf
    I}) v_K({\mathbf I},\tilde {\mathbf I}) = \frac{(2 \mu_x \mu_y +
    C_1)(2 \sigma_{xy} + C_2)}{(\mu_x^2  + \mu_y^2 + C_1)(\sigma_x^2 +
    \sigma_y^2 + C_2)},
\end{equation*}
where $m_K({\mathbf I}, \tilde {\mathbf I})$ is the luminance
distortion, $v_K({\mathbf I}, \tilde {\mathbf I})$ are the contrast
components, and $\mu_x$, $\mu_y$, $\sigma_x$, $\sigma_y$ and
$\sigma_{xy}$ are the local means, standard deviations, and
cross-covariance for images ${\mathbf I}(x, y)$ and
$\tilde {\mathbf I}(x, y)$ with $C_1 = 2.55$, $C_2 = 7.65$.

MS-SSIM \cite{WSB-03} expands on the SSIM index by combining luminance
information with the variance and cross-correlation components
at scale $K$. The multiple scales account for variability in the
perception of image details caused by factors such as viewing distance
from the image, distance from the scene to the sensor, and resolution of
the image acquisition sensor.

The MS-SSIM index for scale $K$ is given by \cite{WSB-03}
\begin{equation*}
  MS-SSIM = m_K({\mathbf I},\tilde {\mathbf I})^{\alpha_K} \prod_{k=1}^K
  v_k({\mathbf I},\tilde {\mathbf I})^{\beta_k} r_k({\mathbf I},\tilde
  {\mathbf I})^{\gamma_k}, 
\end{equation*}
where
\begin{equation*}
  r_K({\mathbf I}, \tilde {\mathbf I}) = \frac{\sigma_{xy}+C_3}{\sigma_x \sigma_y + C_3},
\end{equation*}
and $C_3 = C_2/2$ is the structure component at scale $K$ (where $K = 1$
is the full-resolution image). For $K=5$, the exponents are
$\beta_1= \gamma_1 = 0.0448$, $\beta_2= \gamma_2 = 0.2856$,
$\beta_3= \gamma_3 = 0.3001$, $\beta_4= \gamma_4 = 0.2363$, and
$\alpha_5 = \beta_5= \gamma_5 = 0.1333$. The best quality values for
structural information are obtained for SSIM = MS-SSIM = 1.

Image compression of the gray level images {\tt Bart Run}, {\tt
  Cameraman} and {\tt Kiel}, with sizes $128^2$, $256^2$, and $512^2$
pixels respectively, is applied using balanced GHM, CL, SA4, Alpert, and
the two new supercompact multifilters. The balancing for GHM and CL is
according to \cite{SW-98}; for SA4, Alpert, and the new supercompact
multifilters it is of the Haar type.

Results are shown in Tables \ref{tab:04}--\ref{tab:9}. All minimal MSE
and maximal PSNR, SSIM and MS-SSIM values are marked in bold.

Table \ref{tab:04} shows results for the gray level image {\tt Bart
  Run}. When CR = 4, all multiwavelet filter banks from decomposition
levels 2 through 5 achieve nearly constant SSIM and MS-SSIM. The lowest
MSE and highest values for PSNR, SSIM and MS-SSIM are obtained by the
Integer Haar multiwavelet.  The supercompact multiwavelets compared with
other multiwavelets (GHM, SA4 and CL) leads to better PSNR, SSIM and
MS-SSIM.

When CR = 8, nearly constant SSIM and MS-SSIM are achieved for all
multiwavelet filter banks from the third to the fifth level
decomposition.  The lowest MSE and highest values for PSNR are achieved
by using the Alpert multiwavelet. We need to note that better MSE and
PSNR do not lead to better SSIM and MS-SSIM.  The highest SSIM is by
using CL multiwavelet while and highest MS-SSIM values are achieved for
3 and more level decompositions from the New1 supercompact multiwavelet.

The next two CR = 16 and 32 show that at the first compression level all
the measured values for the Alpert, New1 and New2 multiwavelets are
equal. This comes from the similarity in structure of the pairs of
scaling and wavelet functions. Moreover, Alpert and New1 multiwavelet
filters achieve best PSNRs, SSIMs and MS-SSIMs. For CR =16, best MSEs
and PSNRs values are for the New1 multiwavelet at 3 and more
decomposition level, while for CR = 32 it is the Alpert multiwavelet.
With respect to SSIMs and MS-SSIMs for CR = 16, the Alpert multiwavelet
achieved the best values for 3 and more decomposition levels, while for CR=
32 the best values are from 1 to 3 decomposition levels with the New1
multiwavelet filter.

% Table 4
\begin{table}
\centering 
  \caption{Results for {\tt Bart Run}. The four lines for each multiwavelet represent MSE, PSNR, SSIM and MS-SSIM, in this order.}%
     {\label{tab:04}
       \footnotesize
       \setlength\tabcolsep{4pt}
       \renewcommand{\arraystretch}{1.0}
       \begin{tabular}{|c|c|c|c|c|c|c|c|}
         \hline
         \multirow{2}{*}{Image} &
         \multirow{2}{*}{CR} &
         Multi- &                       
         \multicolumn{5}{c|}{Decomposition level} \\
         \cline{4-8}
         & & wavelet & 1 & 2 & 3 & 4 & 5 \\
         \hline
         \multirow{57}{*}{\parbox{1.5in}{\includegraphics[width=1.5in]{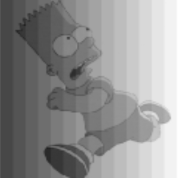}
         \\\centering {\tt Bart Run} }} & \multirow{28}{*}{4:1} &
         \multirow{4}{*}{GHM} & 
            86.167 & 3.0603 & 2.4312 & 2.3896 & 2.3896 \\
            & & & 28.78 & 43.27 & 44.27 & 44.35 & 44.35 \\
            & & & 0.9031 & 0.9830 & 0.9857 & 0.9859 & 0.9859 \\
            & & & 0.9509 & 0.9984 & 0.9986 & 0.9986 & 0.9986 \\
         \cline{3-8}
         & &  \multirow{4}{*}{SA4} &
            1.0416 & 1.0153 & 1.0305 & 1.0559 & 1.0548 \\
            & & & 47.95 & 48.07 & 48.00 & 47.89 & 47.9 \\
            & & & 0.9935 & 0.9932 & 0.993 & 0.9929 & 0.9929 \\
            & & & 0.9995 & 0.9994 & 0.9994 & 0.9994 & 0.9994 \\
          \cline{3-8}
         & &  \multirow{4}{*}{CL} &
           {\bf  0.5866} & 0.5192 & 0.4995 & 0.5065 & 0.5111 \\
            & & & {\bf 50.45} & 50.98 & 51.15 & 51.08 & 51.05 \\
            & & & 0.9967 & 0.9969 & 0.9969 & 0.9969 & 0.9969 \\
            & & & {\bf 0.9997} & 0.9997 & {\bf 0.9997} & {\bf 0.9997} & {\bf 0.9997} \\
          \cline{3-8}
         & & \multirow{4}{*}{Alpert} &
            0.6630 & 0.5090 & 0.4955 & 0.4993 & 0.5040 \\
            & & & 49.92 & 51.06 & 51.18 & 51.15 & 51.11 \\
            & & & 0.9966 & 0.9970 & 0.9971 & {\bf 0.9971} & {\bf 0.997} \\
            & & & {\bf 0.9997} & 0.9997 & 0.9997 & {\bf 0.9997} & {\bf 0.9997} \\
         \cline{3-8}
         & & \multirow{4}{*}{New1} &
            0.6096 & 0.4987 & 0.4923 & 0.4983 & 0.5021 \\
            & & & 50.28 & 51.15 & 51.21 & 51.16 & 51.12 \\
            & & & {\bf 0.9967} & 0.9972 & 0.9972 & 0.9971 & {\bf 0.9971} \\
            & & & {\bf 0.9997} & 0.9997 & 0.9997 & 0.9997 & {\bf 0.9997} \\
         \cline{3-8}
         & & \multirow{4}{*}{New2} &
            1.1408 & 0.9912 & 0.9570 & 0.9782 & 0.9921 \\
            & & & 47.56 & 48.17 & 48.32 & 48.23 & 48.17 \\
            & & & 0.9948 & 0.9949 & 0.9951 & 0.9950 & 0.9949 \\
            & & & 0.9995 & 0.9994 & 0.9994 & 0.9994 & 0.9994 \\
         \cline{3-8}
         & & \multirow{4}{*}{Integer Haar} &
            0.6234 & {\bf 0.4246} & {\bf 0.3891} & {\bf 0.3919} & {\bf 0.3984} \\
            & & & 50.18 & {\bf 51.85} & {\bf 52.23} & {\bf 52.20} & {\bf 52.13} \\
            & & & {\bf 0.9972} & {\bf 0.9977} & {\bf 0.9978} & {\bf 0.9978} & {\bf 0.9978} \\
            & & & {\bf 0.9997} & {\bf 0.9998} & {\bf 0.9998} & {\bf 0.9998} & {\bf 0.9998} \\
         \cline{2-8} \\
         \cline{2-8}
        & \multirow{28}{*}{8:1} &
         \multirow{4}{*}{GHM} & 
            5385.1 & 15.463 & 11.505 & 11.127 & 11.071 \\
            & & & 10.82 & 36.24 & 37.52 & 37.67 & 37.69 \\
            & & & 0.2538 & 0.9241 & 0.9404 & 0.9425 &  0.9428 \\
            & & & 0.6421 & 0.9912 & 0.9934 & 0.9936 & 0.9936 \\
         \cline{3-8}
         & &  \multirow{4}{*}{SA4} &
            9.9828 & 6.9655 & 6.815 & 6.9071 & 6.9234 \\
            & & & 38.14 & 39.70 & 39.80 & 39.74 & 39.73 \\
            & & & 0.9507 & 0.9633 & 0.9634 & 0.9634 & 0.9632 \\
            & & & 0.9953 & {\bf 0.9957} & 0.9955 & 0.9954 & 0.9954 \\
          \cline{3-8}
         & &  \multirow{4}{*}{CL} &
            {\bf 9.8477} & 6.6463 & 5.931 & 5.973 & 6.0143 \\
            & & & {\bf 38.20} & 39.91 & 40.40 & 40.37 & 40.34 \\
            & & & {\bf 0.9531} & {\bf 0.9668} & 0.9699 & {\bf 0.9696} & {\bf 0.9695} \\
            & & & 0.9936 & 0.9956 & 0.9957 & 0.9956 & 0.9956 \\
          \cline{3-8}
         & & \multirow{4}{*}{Alpert} &
            10.4090 & {\bf 6.3780} & {\bf 5.7271} & {\bf 5.7040} & {\bf 5.7534} \\
            & & & 37.96 & {\bf 40.08} & {\bf 40.55} & {\bf 40.57} & {\bf 40.53} \\
            & & & 0.9458 & 0.9658 & 0.9683 & 0.9682 & 0.9680 \\
            & & & 0.9955 & 0.9962 & 0.9961 & 0.9960 & {\bf 0.9960} \\
         \cline{3-8}
         & & \multirow{4}{*}{New1} &
            9.9606 & 6.4132 & 5.9374 & 5.9642 & 6.0197 \\
            & & & 38.15 & 40.06 & 40.39 & 40.38 & 40.34 \\
            & & & 0.9494 & 0.9646 & {\bf 0.9684} & 0.9682 & 0.9681 \\
            & & & {\bf 0.9958} & 0.9960 & {\bf 0.9962} &{\bf 0.9962} & {\bf 0.9960} \\
         \cline{3-8}
         & & \multirow{4}{*}{New2} &
            14.023 & 11.193 & 10.575 & 10.776 & 10.767 \\
            & & & 36.66 & 37.64 & 37.89 & 37.81 & 37.81 \\
            & & & 0.9244 & 0.9381 & 0.9460 & 0.944 & 0.9444 \\
            & & & 0.9918 & 0.9919 & 0.9932 & 0.9931 & 0.9931 \\
         \cline{3-8}
         & & \multirow{4}{*}{Integer Haar} &
            12.76 & 8.2607 & 7.4608 & 7.5038 & 7.6042 \\
            & & & 37.07 & 38.96 & 39.40 & 39.38 & 39.32 \\
            & & & 0.9411 & 0.9593 & 0.9628 & 0.9625 & 0.9620 \\
            & & & 0.9932 & 0.9950 & 0.9953 & 0.9952 & 0.9952 \\
         \hline
      \end{tabular}}{}
\end{table}
% Table 4(continued)
\begin{table}
\centering 
  \renewcommand\thetable{4(continued)}
  \caption{Results for {\tt Bart Run}. The four lines for each multiwavelet represent MSE, PSNR, SSIM and MS-SSIM, in this order.}%
     {
       \footnotesize
       \setlength\tabcolsep{4pt}
       \renewcommand{\arraystretch}{1.0}
       \begin{tabular}{|c|c|c|c|c|c|c|c|}
         \hline
         \multirow{2}{*}{Image} &
         \multirow{2}{*}{CR} &
         Multi- &                       
         \multicolumn{5}{c|}{Decomposition level} \\
         \cline{4-8}
         & & wavelet & 1 & 2 & 3 & 4 & 5 \\
         \hline
         \multirow{57}{*}{\parbox{1.5in}{\includegraphics[width=1.5in]{bart_run}
         \\\centering {\tt Bart Run} }} & \multirow{28}{*}{16:1} &
         \multirow{4}{*}{GHM} & 
            11837 & 189.7 & 27.636 & 25.712 & 25.596 \\
            & & & 7.399 & 25.35 & 33.72 & 34.03 & 34.05 \\
            & & & 0.0142 & 0.7820 & 0.8719 & 0.8798 & 0.8802 \\
            & & & 0.4809 & 0.8622 & 0.9792 & 0.9806 & 0.9806 \\
          \cline{3-8}
         & &  \multirow{4}{*}{SA4} &
            189.23 & {\bf 19.975} & 18.572 & 18.818 & 18.889 \\
            & & & 25.36 & {\bf 35.13} & 35.44 & 35.39 & 35.37 \\
            & & & 0.766 & {\bf 0.9033} & 0.9099 & 0.9089 & 0.9086 \\
            & & & 0.9008 & 0.9850 & 0.9869 & 0.9863 & 0.9862 \\
          \cline{3-8}
         & &  \multirow{4}{*}{CL} &
            240.87 & 22.234 & 18.827 & 18.763 & 18.888 \\
            & & & 24.31 & 34.66 & 35.38 & 35.40 & 35.37 \\
            & & & 0.7795 & 0.8946 & 0.9076 & 0.9076 & 0.9072 \\
            & & & 0.8490 & 0.9829 & 0.9852 & 0.9848 & 0.9846 \\
         \cline{3-8}
         & & \multirow{4}{*}{Alpert} &
            {\bf 64.561} & 20.539 & 17.289 & 17.243 & 17.264 \\
            & & & {\bf 30.03} & 35.00 & 35.75 & 35.76 & {\bf 35.76} \\
            & & & {\bf 0.7730} & 0.8943 & {\bf 0.9111} & {\bf 0.9113} & {\bf 0.9113} \\
            & & & {\bf 0.9622} & 0.9852 & {\bf 0.9881} & {\bf 0.9879} & {\bf 0.9878} \\
         \cline{3-8}
         & & \multirow{4}{*}{New1} &
            {\bf 64.561} & 20.152 & {\bf 17.055} & {\bf 17.054} & {\bf 17.251} \\
            & & & {\bf 30.03} & 35.09 & {\bf 35.81} & {\bf 35.81} & {\bf 35.76} \\
            & & & {\bf 0.7730} & 0.8969 & 0.9107 & 0.9105 & 0.9096 \\
            & & & {\bf 0.9622} & {\bf 0.9856} & 0.9878 & 0.9874 & 0.9872 \\
         \cline{3-8}
         & & \multirow{4}{*}{New2} &
            {\bf 64.561} & 29.797 & 27.838 & 28.114 & 28.25 \\
            & & & {\bf 30.03} & 33.39 & 33.68 & 33.64 & 33.62 \\
            & & & {\bf 0.7730} & 0.8478 & 0.8597 & 0.8586 & 0.8581 \\
            & & & {\bf 0.9622} & 0.9743 & 0.9777 & 0.9770 & 0.9768 \\
         \cline{3-8}
         & & \multirow{4}{*}{Integer Haar} &
            317.99 & 26.172 & 23.221 & 23.302 & 23.494 \\
            & & & 23.11 & 33.95 & 34.47 & 34.46 & 34.42 \\
            & & & 0.7559 & 0.8760 & 0.8878 & 0.8872 & 0.8862 \\
            & & & 0.8150 & 0.9791 & 0.9816 & 0.9811 & 0.9809 \\
         \cline{2-8} \\
         \cline{2-8}
        & \multirow{28}{*}{32:1} &
         \multirow{4}{*}{GHM} & 
            17368 & 4704.3 & 73.563 & 51.504 & 50.794 \\
            & & & 5.733 & 11.41 & 29.46 & 31.01 & 31.07 \\
            & & & 0.00814 & 0.2671 & 0.7643 & 0.7942 & 0.7976 \\
            & & & 0.4401 & 0.5499 & 0.9338 & 0.9531 & 0.9542 \\
         \cline{3-8}
         & &  \multirow{4}{*}{SA4} &
            5415.3 & 47.809 & 37.967 & 37.859 & 38.058 \\
            & & & 10.79 & 31.34 & 32.34 & 32.35 & 32.33 \\
            & & & 0.3788 & 0.8026 & 0.8412 & 0.8424 & 0.842 \\
            & & & 0.4941 & 0.9528 & 0.9681 & 0.9664 & 0.9663 \\
          \cline{3-8}
         & &  \multirow{4}{*}{CL} &
            5269.8 & 56.575 & 38.712 & 37.714 & 37.993 \\
            & & & 10.91 & 30.60 & 32.25 & 32.37 & 32.33 \\
            & & & 0.378 & 0.7439 & 0.8311 & 0.8342 & 0.8338 \\
            & & & 0.5236 & 0.9347 & 0.9650 & 0.9654 & 0.9652 \\
          \cline{3-8}
         & & \multirow{4}{*}{Alpert} &
            {\bf 5018} & 44.849 & 32.806 & {\bf 32.498} & {\bf 32.648} \\
            & & & {\bf 11.13} & 31.61 & {\bf 32.97} & {\bf 33.01} & {\bf 32.99} \\
            & & & {\bf 0.3735} & 0.7993 & 0.8463 & {\bf 0.8463} & {\bf 0.8462} \\
            & & & {\bf 0.6502} & 0.9596 & 0.9711 & {\bf 0.9722} & {\bf 0.9718} \\
         \cline{3-8}
         & & \multirow{4}{*}{New1} &
            {\bf 5018} & {\bf 44.08} & {\bf 32.803} & 32.601 & 32.927 \\
            & & & {\bf 11.13} & {\bf 31.69} & {\bf 32.97} & 33.00 & 32.96 \\
            & & & {\bf 0.3735} & {\bf 0.8029} & {\bf 0.8478} & 0.8469 & {\bf 0.8462} \\
            & & & {\bf 0.6502} & {\bf 0.9609} & {\bf 0.9728} & 0.9717 & 0.9713 \\
         \cline{3-8}
         & & \multirow{4}{*}{New2} &
            {\bf 5018} & 57.571 & 51.015 & 51.224 & 51.497 \\
            & & & {\bf 11.13} & 30.53 & 31.05 & 31.04 & 31.01 \\
            & & & {\bf 0.3735} & 0.7503 & 0.7703 & 0.7692 & 0.7693 \\
            & & & {\bf 0.6502} & 0.9389 & 0.9510 & 0.9500 & 0.9495 \\
         \cline{3-8}
         & & \multirow{4}{*}{Integer Haar} &
            5524.4 & 58.614 & 48.135 & 47.572 & 47.93 \\
            & & & 10.71 & 30.45 & 31.31 & 31.36 & 31.32 \\
            & & & 0.3684 & 0.7495 & 0.801 & 0.8023 & 0.8003 \\
            & & & 0.4943 & 0.9393 & 0.9586 & 0.9585 & 0.9579 \\
         \hline
      \end{tabular}}{}
\end{table}

Table \ref{tab:05} shows results for the gray level image {\tt
  Cameraman}. For CR = 4 GHM has better MS-SSIM, but for CR = 8, 16 and
32 the New1 supercompact multifilter filter for 3 and more decomposition
levels achieved better values than all the rest. It is important to note
that all supercompact multiwavelets have nearly constant values for
PSNR, SSIM and MS-SSIM for 4 and more decomposition levels.

% Table 5
\begin{table}
\centering 
  \renewcommand\thetable{5}
  \caption{Results for {\tt Cameramsan}. The four lines for each multiwavelet represent MSE, PSNR, SSIM and MS-SSIM, in this order.}%
     {\label{tab:05}
       \footnotesize
       \setlength\tabcolsep{4pt}
       \renewcommand{\arraystretch}{1.0}
       \begin{tabular}{|c|c|c|c|c|c|c|c|c|}
         \hline
         \multirow{2}{*}{Image} &
         \multirow{2}{*}{CR} &
         Multi- &                       
         \multicolumn{6}{c|}{Decomposition level} \\
         \cline{4-9}
         & & wavelet & 1 & 2 & 3 & 4 & 5 & 6\\
         \hline
         \multirow{57}{*}{\parbox{1.5in}{\includegraphics[width=1.5in]{Cameraman}
         \\\centering {\tt cameraman} }} & \multirow{28}{*}{4:1} &
         \multirow{4}{*}{GHM} & 
            65.773 & 9.814 & 7.9468 & 7.7663 & 7.7578 & 7.7579 \\
            & & & 29.95 & 38.21 & 39.13 & 39.23 & 39.23 & 39.23 \\
            & & & 0.846 & 0.9596 & 0.963 & 0.9633 & 0.9633 & 0.9634 \\
            & & & 0.9778 & {\bf 0.9945} & {\bf 0.9947} & {\bf 0.9947} & {\bf 0.9947} & {\bf 0.9947} \\
         \cline{3-9}
         & &  \multirow{4}{*}{SA4} &
            6.4772 & 5.6798 & 5.7046 & 5.7165 & 5.7208 & 5.7204 \\
            & & & 40.02 & 40.59 & 40.57 & 40.56 & 40.56 &  40.56 \\
            & & & {\bf 0.9646} & 0.9656 & 0.9659 & 0.9659 & 0.9659 & 0.9659 \\
            & & & 0.9946 & {\bf 0.9945} & 0.9945 & 0.9945 & 0.9945 & 0.9945 \\
          \cline{3-9}
         & &  \multirow{4}{*}{CL} &
            6.3258 & 5.2186 & 5.1314 & 5.129 & 5.1226 & 5.1235 \\
            & & & 40.12 & 40.96 & 41.03 & 41.03 & 41.04 & 41.04 \\
            & & & 0.9637 & 0.9655 & 0.9663 & 0.9664 & 0.9665 & 0.9665 \\
            & & & 0.9941 & 0.9939 & 0.9942 & 0.9941 & 0.9941 & 0.9941 \\
          \cline{3-9}
         & & \multirow{4}{*}{Alpert} &
            6.1808 & 5.1158 & 4.9445 & 4.9299 & 4.9375 & 4.9356 \\
            & & & 40.22 & 41.04 & 41.19 & 41.20 & 41.20  & 41.20 \\
            & & & 0.9637 & 0.9662 & 0.9672 & 0.9672 & 0.9672 & 0.9672 \\
            & & & 0.9952 & 0.9941 & 0.9946 & 0.9945 & 0.9945  & 0.9945 \\
         \cline{3-9}
         & & \multirow{4}{*}{New1} &
            {\bf 6.1103} & {\bf 5.0703} & 4.9072 & {\bf 4.8982} & {\bf 4.9033} & {\bf 4.9044} \\
            & & & {\bf 40.27} & {\bf 41.08} & {\bf 41.22} & {\bf 41.23} & {\bf 41.23} & {\bf 41.22} \\
            & & & {\bf 0.964} & {\bf 0.9664} & {\bf 0.9675} & {\bf 0.9675} & {\bf 0.9674} & {\bf 0.9674} \\
            & & & {\bf 0.9953} & 0.9942 & 0.9947 & 0.9946 & 0.9946 & 0.9946 \\
         \cline{3-9}
         & & \multirow{4}{*}{New2} &
            8.0861 & 7.0528 & 7.0942 & 7.1256 & 7.1481 & 7.1492 \\
            & & & 39.05 & 39.65 & 39.62 & 39.6 & 39.59 & 39.59 \\
            & & & 0.9584 & 0.9585 & 0.9585 & 0.9584 & 0.9583 & 0.9583 \\
            & & & 0.9944 & 0.9920 & 0.9923 & 0.9923 & 0.9923 & 0.9923 \\
         \cline{3-9}
         & & \multirow{4}{*}{Integer Haar} &
            7.045 & 5.9673 & 5.8743 & 5.8631 & 5.8739 & 5.874 \\
            & & & 39.65 & 40.37 & 40.44 & 40.45 & 40.44 & 40.44 \\
            & & & 0.9616 & 0.9628 & 0.9633 & 0.9634 & 0.9633 & 0.9633 \\
            & & & 0.9935 & 0.9933 & 0.9935 & 0.9936 & 0.9935 & 0.9935 \\
         \cline{2-9} \\
         \cline{2-9}
        & \multirow{28}{*}{8:1} &
         \multirow{4}{*}{GHM} & 
            4488 & 54.003 & 37.34 & 35.658 & 35.56 & 35.538 \\
            & & & 11.61 & 30.81 & 32.41 & 32.61 & 32.62 & 32.62 \\
            & & & 0.1478 & 0.8883 & 0.9086 & 0.9111 & 0.9113 & 0.9113 \\
            & & & 0.6785 & 0.98 & {\bf 0.9835} & 0.9833 & 0.9833 & 0.9833 \\
         \cline{3-9}
         & &  \multirow{4}{*}{SA4} &
            39.525 & 27.101 & 26.908 & 26.987 & 27.012 &  27.013 \\
            & & & 32.16 & 33.8 & 33.83 & 33.82 & 33.82 & 33.82 \\
            & & & {\bf 0.9050} & 0.9197 & 0.9188 & 0.919 & 0.919 &  0.9189 \\
            & & & 0.9826 & 0.9838 & 0.9820 & 0.9818 & 0.9818 & 0.9818 \\
          \cline{3-9}
         & &  \multirow{4}{*}{CL} &
            41.284 & 26.052 & 25.502 & 25.39 & 25.433 & 25.427 \\
            & & & 31.97 & 33.97 & 34.06 & 34.08 & 34.08 & 34.08 \\
            & & & 0.8995 & 0.9192 & 0.9203 & 0.9209 & 0.9208 & 0.9208 \\
            & & & 0.9793 & 0.9819 & 0.9809 & 0.9813 & 0.9815 & 0.9815 \\
          \cline{3-9}
         & & \multirow{4}{*}{Alpert} &
            40.38 & 25.849 & 24.791 & 24.635 & 24.693 & 24.698 \\
            & & & 32.07 & 34.01 & 34.19 & 34.22 & 34.21 & 34.20 \\
            & & & 0.8986 & 0.9193 & 0.9215 & 0.9222 & 0.9221 & 0.922 \\
            & & & 0.9859 & 0.9841 & 0.9831 & 0.9839 & 0.9838 & 0.9838 \\
         \cline{3-9}
         & & \multirow{4}{*}{New1} &
            {\bf 39.706} & {\bf 25.503} & {\bf 24.488} & {\bf 24.351} & {\bf 24.413} & {\bf 24.429} \\
            & & & {\bf 32.14} & {\bf 34.06} & {\bf 34.24} & {\bf 34.27} & {\bf 34.25} & {\bf 34.25} \\
            & & & 0.9004 & {\bf 0.9204} & {\bf 0.9226} & {\bf 0.9231} & {\bf 0.9228} & {\bf 0.9228} \\
            & & & {\bf 0.9863} & {\bf 0.9844} & {\bf 0.9836} & {\bf 0.9841} & {\bf 0.9841} & {\bf 0.9841} \\
         \cline{3-9}
         & & \multirow{4}{*}{New2} &
            52.833 & 37.008 & 36.229 & 36.612 & 36.775 & 36.796 \\
            & & & 30.90 & 32.45 & 32.54 & 32.49 & 32.48 & 32.47 \\
            & & & 0.8819 & 0.8995 & 0.8990 & 0.8976 & 0.8971 &  0.8971 \\
            & & & 0.9822 & 0.9793 & 0.9743 & 0.9753 & 0.975 & 0.9751 \\
         \cline{3-9}
         & & \multirow{4}{*}{Integer Haar} &
            48.803 & 32.369 & 31.122 & 31.10 & 31.151 & 31.15 \\
            & & & 31.25 & 33.03 & 33.20 & 33.20 & 33.20 & 33.20 \\
            & & & 0.8869 & 0.9078 & 0.9080 & 0.9077 & 0.9075 & 0.9074 \\
            & & & 0.9742 & 0.9794 & 0.9783 & 0.9787 & 0.9787 & 0.9787 \\
         \hline
      \end{tabular}}{}
\end{table}
% Table 5(continued)
\begin{table}
\centering 
  \renewcommand\thetable{5(continued)}
  \caption{Results for {\tt Cameraman}. The four lines for each multiwavelet represent MSE, PSNR, SSIM and MS-SSIM, in this order.}%
    {
       \footnotesize
       \setlength\tabcolsep{4pt}
       \renewcommand{\arraystretch}{1.0}
       \begin{tabular}{|c|c|c|c|c|c|c|c|c|}
         \hline
         \multirow{2}{*}{Image} &
         \multirow{2}{*}{CR} &
         Multi- &                       
         \multicolumn{6}{c|}{Decomposition level} \\
         \cline{4-9}
         & & wavelet & 1 & 2 & 3 & 4 & 5 & 6\\
         \hline
         \multirow{57}{*}{\parbox{1.5in}{\includegraphics[width=1.5in]{Cameraman}
         \\\centering {\tt cameraman} }} & \multirow{28}{*}{16:1} &
         \multirow{4}{*}{GHM} & 
            9398.3 & 206.16 & 97.873 & 89.405 & 88.794 & 88.729 \\
            & & & 8.4 & 24.99 & 28.22 & 28.62 & 28.65  & 28.65 \\
            & & & 0.05697 & 0.6939 & 0.8304 & 0.837 & 0.8379  & 0.8379 \\
            & & & 0.5289 & 0.9337 & 0.9586 & 0.959 & 0.959   & 0.959 \\
         \cline{3-9}
         & &  \multirow{4}{*}{SA4} &
            {\bf 180.76} & 75.635 & 71.015 & 71.511 & 71.684  & 71.656 \\
            & & & {\bf 25.56} & 29.34 & 29.62 & 29.59 & 29.58  & 29.58 \\
            & & & {\bf 0.6773} & {\bf 0.8468} & 0.8472 & 0.8458 & 0.8455  &  0.8456 \\
            & & & 0.9446 & 0.9618 & 0.9562 & 0.9564 & 0.956  & 0.9559 \\
          \cline{3-9}
         & &  \multirow{4}{*}{CL} &
            203.43 & 75.691 & 70.133 & 70.075 & 70.233 &70.217 \\
            & & & 25.05 & 29.34 & 29.67 & 29.68 & 29.67  & 29.67 \\
            & & & 0.6668 & 0.8445 & 0.8466 & 0.8486 & 0.8481 & 0.8482 \\
            & & & 0.9383 & 0.9594 & 0.9550 & 0.956 & 0.9555  & 0.9556 \\
          \cline{3-9}
         & & \multirow{4}{*}{Alpert} &
            202.99 & 75.024 & 67.809 & 67.454 & 67.673 & 67.704 \\
            & & & 25.06 & 29.38 & 29.82 & 29.84 & 29.83 & 29.82 \\
            & & & 0.665 & 0.8436 & 0.8472 & 0.8492 & 0.8488  & 0.8488 \\
            & & & 0.9601 & 0.9641 & 0.9582 & 0.962 & 0.962 & 0.9622 \\
         \cline{3-9}
         & & \multirow{4}{*}{New1} &
            199.72 & {\bf 73.971} & {\bf 66.886} & {\bf 66.615} & {\bf 66.842} & {\bf 66.858} \\
            & & & 25.13 & {\bf 29.44} & {\bf 29.88} & {\bf 29.90} & {\bf 29.88} & {\bf 29.88} \\
            & & & 0.6665 & 0.8448 & {\bf 0.8497} & {\bf 0.8516} & {\bf 0.8515} & {\bf 0.8516} \\
            & & & {\bf 0.9605} & {\bf 0.9651} & {\bf 0.9594} & {\bf 0.9631} & {\bf 0.9628 } & {\bf 0.9630} \\
         \cline{3-9}
         & & \multirow{4}{*}{New2} &
            234.83 & 105.86 & 99.378 & 100.62 & 101.19 &101.3 \\
            & & & 24.42 & 27.88 & 28.16 & 28.1 & 28.08  &  28.07 \\
            & & & 0.6541 & 0.8131 & 0.8135 & 0.81 & 0.8088  & 0.8087 \\
            & & & 0.9566 & 0.954 & 0.9460 & 0.9444 & 0.9441 & 0.9442 \\
         \cline{3-9}
         & & \multirow{4}{*}{Integer Haar} &
            229.89 & 97.676 & 87.721 & 87.61 & 87.707 & 87.699 \\
            & & & 24.52 & 28.23 & 28.7 & 28.71 & 28.7 & 28.7 \\
            & & & 0.655 & 0.8190 & 0.8212 & 0.8193 & 0.8191 & 0.8191 \\
            & & & 0.9284 & 0.9495 & 0.9467 & 0.9479 & 0.9479 & 0.9478 \\
         \cline{2-9} \\
         \cline{2-9}
        & \multirow{28}{*}{32:1} &
         \multirow{4}{*}{GHM} & 
            12589 & 3469.7 & 214.42 & 179.14 & 175.67 & 175.25 \\
            & & & 7.131 & 12.73 & 24.82 & 25.60 & 25.68 & 25.69 \\
            & & & 0.02968 & 0.1307 & 0.7325 & 0.7593 & 0.7615 & 0.7616 \\
            & & & 0.4926 & 0.5191 & 0.9122 & 0.92 & 0.9203 & 0.9201 \\
         \cline{3-9}
         & &  \multirow{4}{*}{SA4} &
            {\bf 4094.8} & {\bf 180.79} & 145.72 & 147.17 & 147.62 & 147.59 \\
            & & & {\bf 12.01} & {\bf 25.56} & 26.50 & 26.45 & 26.44 & 26.44 \\
            & & & {\bf 0.4219} & 0.7478 & 0.7673 & 0.7637 & 0.7636 & 0.7635 \\
            & & & 0.6840 & 0.9209 & 0.9179 & 0.9132 & 0.9131 & 0.9133 \\
          \cline{3-9}
         & &  \multirow{4}{*}{CL} &
            4190.9 & 188.21 & 150.33 & 149.38 & 149.54 & 149.68 \\
            & & & 11.91 & 25.38 & 26.36 & 26.39 & 26.38 & 26.38 \\
            & & & 0.4191 & 0.7432 & 0.7676 & 0.7672 & 0.767 & 0.7670 \\
            & & & 0.6926 & 0.9178 & 0.9184 & 0.9156 & 0.9154 & 0.9152 \\
          \cline{3-9}
         & & \multirow{4}{*}{Alpert} &
            4250.4 & 186.87 & 144.55 & 143.11 & 143.5 & 143.71 \\
            & & & 11.85 & 25.42 & 26.53 & 26.57 & 26.56 & 26.56 \\
            & & & 0.4214 & 0.7467 & 0.7739 & 0.7753 & 0.7735 & 0.7734 \\
            & & & {\bf 0.7224} & 0.931 & 0.9274 & 0.9283 & {\bf 0.9268} & {\bf 0.9269} \\
         \cline{3-9}
         & & \multirow{4}{*}{New1} &
            4250.4 & 183.77 & {\bf 142.56} & {\bf 140.93} & {\bf 141.35} & {\bf 141.41} \\
            & & & 11.85 & 25.49 & {\bf 26.59} & {\bf 26.64} & {\bf 26.63} & {\bf 26.63} \\
            & & & 0.4214 & {\bf 0.7503} & {\bf 0.7756} & {\bf 0.7768} & {\bf 0.7755} & {\bf 0.7752} \\
            & & & {\bf 0.7224} & {\bf 0.9319} & {\bf 0.9276} & {\bf 0.9285} & 0.9264 & 0.9263 \\
         \cline{3-9}
         & & \multirow{4}{*}{New2} &
            4250.4 &  247.83 & 209 & 208.59 & 211.29 & 211.87 \\
            & & & 11.85 & 24.19 & 24.93 & 24.94 & 24.88 & 24.87 \\
            & & & 0.4214 & 0.7057 & 0.7383 & 0.7307 & 0.7273 & 0.727 \\
            & & & {\bf 0.7224} & 0.9141 & 0.9092 & 0.8941 & 0.8962 & 0.8954 \\
         \cline{3-9}
         & & \multirow{4}{*}{Integer Haar} &
            4295.2 & 238.78 & 188.02 & 185.13 & 185.47 & 185.24 \\
            & & & 11.80 & 24.35 & 25.39 & 25.46 & 25.45 & 25.45 \\
            & & & 0.4115 & 0.7034 & 0.7408 & 0.7336 & 0.7331 & 0.733 \\
            & & & 0.6747 & 0.8994 & 0.9052 & 0.9036 & 0.9044 & 0.9046 \\
         \hline
      \end{tabular}}{}
\end{table}

Table \ref{tab:06} shows results for the gray level image {\tt
  Kiel}. For 4 and more decomposition levels and CR = 4, 8 and 16 all
multiwavelets achieved nearly constant MSEs and PSNRs. Better results
are obtained for the two supercompact Alpert and New1 multiwavelet
filters. For the three considered CRs, the New1 multiwavelet has better
values for MS-SSIM, MSE and PSNR.

% Table 6
\begin{table}
\centering 
  \renewcommand\thetable{6}
  \caption{Results for {\tt Kiel}. The four lines for each multiwavelet represent MSE, PSNR, SSIM and MS-SSIM, in this order.}%
     {\label{tab:06}
       \footnotesize
       \setlength\tabcolsep{4pt}
       \renewcommand{\arraystretch}{1.0}
       \begin{tabular}{|c|c|c|c|c|c|c|c|c|}
         \hline
         \multirow{2}{*}{Image} &
         \multirow{2}{*}{CR} &
         Multi- &                       
         \multicolumn{6}{c|}{Decomposition level} \\
         \cline{4-9}
         & & wavelet & 1 & 2 & 3 & 4 & 5 & 6\\
         \hline
         \multirow{57}{*}{\parbox{1.5in}{\includegraphics[width=1.5in]{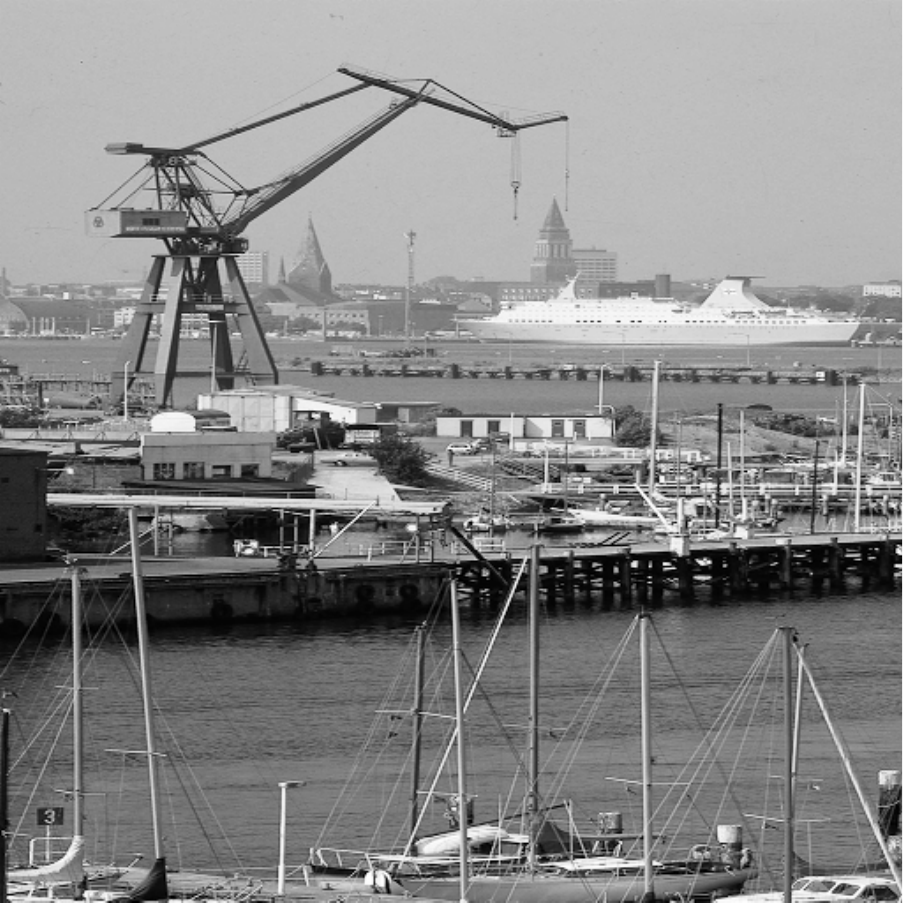}
         \\\centering {\tt Kiel} }} & \multirow{28}{*}{4:1} &
         \multirow{4}{*}{GHM} & 
            235.84 & 34.756 & 28.619 & 27.844 & 27.761 & 27.758 \\
            & & & 24.40 & 32.72 & 33.56 & 33.68 & 33.70 & 33.70 \\
            & & & 0.9432 & {\bf 0.9844} & {\bf 0.9853} & {\bf 0.9855} & {\bf 0.9855} & {\bf 0.9855} \\
            & & & 0.9688 & 0.9911 & {\bf 0.9917} & {\bf 0.9917} & {\bf 0.9916} & {\bf 0.9916} \\
         \cline{3-9}
         & &  \multirow{4}{*}{SA4} &
            24.975 & 21.198 & 20.825 & 20.849 & 20.854 & 20.859 \\
            & & & 34.16 & 34.87 & 34.94 & 34.94 & 34.94 & 34.94 \\
            & & & 0.9846 & 0.9835 & 0.983 & 0.983 & 0.983 & 0.983 \\
            & & & 0.9913 & 0.9913 & 0.9909 & 0.9907 & 0.9908 & 0.9908 \\
          \cline{3-9}
         & &  \multirow{4}{*}{CL} &
            25.139 & {\bf 20.194} & 19.694 & 19.679 & 19.70 & 19.698 \\
            & & & 34.13 & {\bf 35.08} & 35.19 & 35.19 & 35.19 & 35.19 \\
            & & & 0.9805 & 0.9826 & 0.9824 & 0.9824 & 0.9823 & 0.9823 \\
            & & & 0.9903 & 0.9911 & 0.9908 & 0.9907 & 0.9906 & 0.9907 \\
          \cline{3-9}
         & & \multirow{4}{*}{Alpert} &
            25.241 & 20.372 & 19.713 & 19.634 & 19.631 & 19.619 \\
            & & & 34.11 & 35.04 & 35.18 & 35.20 & 35.20 & 35.20 \\
            & & & 0.9844 & 0.9842 & 0.9841 & 0.9841 & 0.9842 & 0.9842 \\
            & & & 0.9926 & 0.9918 & 0.9914 & 0.9914 & 0.9914 & 0.9914 \\
         \cline{3-9}
         & & \multirow{4}{*}{New1} &
            {\bf 25.028} & 20.25 & {\bf 19.605} & {\bf 19.547} & {\bf 19.542} & {\bf 19.539} \\
            & & & {\bf 34.15} & 35.07 & {\bf 35.21} & {\bf 35.22} & {\bf 35.22} & {\bf 35.22} \\
            & & & {\bf 0.9847} & {\bf 0.9844} & 0.9842 & 0.9842 & 0.9843 & 0.9843 \\
            & & & {\bf 0.9927} & {\bf 0.9919} & 0.9914 & {\bf 0.9915} & {\bf 0.9915} & {\bf 0.9915} \\
         \cline{3-9}
         & & \multirow{4}{*}{New2} &
            29.828 & 26.691 & 26.337 & 26.378 & 26.423 & 26.422 \\
            & & & 33.38 & 33.87 & 33.93 & 33.92 & 33.91 & 33.91 \\
            & & & 0.9803 & 0.9788 & 0.9783 & 0.9781 & 0.978 & 0.978 \\
            & & & 0.9911 & 0.9893 & 0.9886 & 0.9884 & 0.9883 & 0.9883 \\
         \cline{3-9}
         & & \multirow{4}{*}{Integer Haar} &
            27.443 & 23.167 & 22.629 & 22.644 & 22.664 & 22.668 \\
            & & & 33.75 & 34.48 & 34.58 & 34.58 & 34.58 & 34.58 \\
            & & & 0.9824 & 0.9818 & 0.9817 & 0.9817 & 0.9817 & 0.9817 \\
            & & & 0.99 & 0.9902 & 0.9899 & 0.9898 & 0.9899 & 0.9899 \\
         \cline{2-9} \\
         \cline{2-9}
        & \multirow{28}{*}{8:1} &
         \multirow{4}{*}{GHM} & 
            5841.7 & 156.69 & 109.85 & 104.61 & 104.08 & 104.06 \\
            & & & 10.47 & 26.18 & 27.72 & 27.93 & 27.96 & 27.96 \\
            & & & 0.2796 & 0.9397 & 0.9482 & {\bf 0.9491} & {\bf 0.9492} & {\bf 0.9492} \\
            & & & 0.6034 & 0.9669 & 0.9722 & 0.9724 & 0.9723 & 0.9723 \\
         \cline{3-9}
         & &  \multirow{4}{*}{SA4} &
            {\bf 118.98} & 82.231 & 78.496 & 78.299 & {\bf 78.345} & 78.355 \\
            & & & {\bf 27.38} & 28.98 & 29.18 & 29.19 & 29.19 & 29.19 \\
            & & & 0.9411 & 0.9476 & 0.9475 & 0.9469 & 0.9468 & 0.9467 \\
            & & & 0.9677 & 0.9732 & 0.9722 & 0.971 & 0.9709 & 0.9708 \\
          \cline{3-9}
         & &  \multirow{4}{*}{CL} &
            123.00 & {\bf 80.573} & {\bf 76.112} & 75.721 & 75.791 & 75.785 \\
            & & & 27.23 & {\bf 29.07} & {\bf 29.32} & 29.34 & 29.33 & 29.33 \\
            & & & 0.9271 & 0.9441 & 0.9440 & 0.9435 & 0.9434 & 0.9434 \\
            & & & 0.9645 & 0.9719 & 0.9710 & 0.9703 & 0.9701 & 0.9702 \\
          \cline{3-9}
         & & \multirow{4}{*}{Alpert} &
            123.4 & 81..216 & 76.271 & 75.632 & 75.664 & 75.654 \\
            & & & 27.22 & 29.03 & 29.31 & 29.34 & 29.34 & 29.34 \\
            & & & 0.9408 & 0.9482 & 0.9486 & 0.9484 & 0.9484 & 0.9483 \\
            & & & 0.9748 & 0.9747 & 0.9734 & 0.9731 & 0.9732 & 0.9732 \\
         \cline{3-9}
         & & \multirow{4}{*}{New1} &
            122.38 & 80.805 & 75.999 & {\bf 75.447} & 75.487 & {\bf 75.451} \\
            & & & 27.25 & 29.06 & 29.32 & {\bf 29.35} & {\bf 29.35} & {\bf 29.35} \\
            & & & {\bf 0.9420} & {\bf 0.9490} & {\bf 0.9492} & {\bf 0.9491} & 0.9489 & 0.9488 \\
            & & & {\bf 0.9751} & {\bf 0.9751} & {\bf 0.9738} & {\bf 0.9736} & {\bf 0.9735} & {\bf 0.9734} \\
         \cline{3-9}
         & & \multirow{4}{*}{New2} &
            137.67 & 102.07 & 98.629 & 98.492 & 98.688 & 98.733 \\
            & & & 26.74 & 28.04 & 28.19 & 28.20 & 28.19 & 28.19 \\
            & & & 0.9344 & 0.9322 & 0.931 & 0.9307 & 0.9305 & 0.9303 \\
            & & & 0.9724 & 0.9672 & 0.9652 & 0.9645 & 0.9643 & 0.9643 \\
         \cline{3-9}
         & & \multirow{4}{*}{Integer Haar} &
            132.02 & 90.962 & 86.343 & 85.976 & 86.114 & 86.134 \\
            & & & 26.92 & 28.54 & 28.77 & 28.79 & 28.78 & 28.78 \\
            & & & 0.9351 & 0.9408 & 0.9416 & 0.941 & 0.9408 & 0.9408 \\
            & & & 0.964 & 0.9693 & 0.9687 & 0.968 & 0.9679 & 0.9679 \\
         \hline
      \end{tabular}}{}
\end{table}
% Table 6(continued)
\begin{table}
\centering 
  \renewcommand\thetable{6(continued)}
  \caption{Results for {\tt Kiel}. The four lines for each multiwavelet represent MSE, PSNR, SSIM and MS-SSIM, in this order.}%
    {
       \footnotesize
       \setlength\tabcolsep{4pt}
       \renewcommand{\arraystretch}{1.0}
       \begin{tabular}{|c|c|c|c|c|c|c|c|c|}
         \hline
         \multirow{2}{*}{Image} &
         \multirow{2}{*}{CR} &
         Multi- &                       
         \multicolumn{6}{c|}{Decomposition level} \\
         \cline{4-9}
         & & wavelet & 1 & 2 & 3 & 4 & 5 & 6\\
         \hline
         \multirow{28}{*}{\parbox{1.5in}{\includegraphics[width=1.5in]{kiel}
         \\\centering {\tt Kiel} }} & \multirow{28}{*}{16:1} &
         \multirow{4}{*}{GHM} & 
            11418 & 549 & 259.13 & 238.48 & 236.06 & 235.88 \\
            & & & 7.555 & 20.74 & 24.00 & 24.36 & 24.40 & 24.40 \\
            & & & 0.1151 & 0.8160 & 0.8871 & 0.8902 & 0.8908 & 0.8909 \\
            & & & 0.4868 & 0.8957 & 0.9384 & 0.9393 & 0.9392 & 0.9391 \\
         \cline{3-9}
         & &  \multirow{4}{*}{SA4} &
            {\bf 536.64} & 202.31 & 182.95 & 181.18 & 181.18 & 181.26 \\
            & & & {\bf 20.83} & 25.07 & 25.51 & 25.55 & 25.55 & 25.55 \\
            & & & {\bf 0.8016} & 0.8867 & 0.8895 & 0.8881 & 0.8875 & 0.8876 \\
            & & & 0.8826 & 0.9405 & 0.9393 & 0.9365 & 0.9358 & 0.9358 \\
          \cline{3-9}
         & &  \multirow{4}{*}{CL} &
            548.81 & 201.50 & 180.29 & 178.04 & 177.97 & 177.99 \\
            & & & 20.74 & {\bf 25.09} & 25.57 & 25.63 & 25.63 & 25.63 \\
            & & & 0.7703 & 0.8840 & 0.8857 & 0.8855 & 0.8850 & 0.8849 \\
            & & & 0.8838 & 0.9395 & 0.9381 & 0.9371 & 0.9365 & 0.9365 \\
          \cline{3-9}
         & & \multirow{4}{*}{Alpert} &
            560.92 & 201.99 & {\bf 180.07} & {\bf 176.93} & {\bf 176.94} & {\bf 176.88} \\
            & & & 20.64 & 25.08 & {\bf 25.58} & {\bf 25.65} & {\bf 25.65} & {\bf 25.65} \\
            & & & 0.7958 & {\bf 0.8878} & 0.8930 & 0.8932 & {\bf 0.8929} & {\bf 0.8929} \\
            & & & 0.9193 & {\bf 0.9451} & {\bf 0.9446} & 0.9434 & {\bf 0.9431} & {\bf 0.9433} \\
         \cline{3-9}
         & & \multirow{4}{*}{New1} &
            558.71 & {\bf 201.49} & 180.21 & 177.25 & 177.28 & 177.19 \\
            & & & 20.66 & {\bf 25.09} & 25.57 & 25.64 & 25.64 & {\bf 25.65} \\
            & & & 0.7982 & 0.8875 & {\bf 0.8931} & {\bf 0.8934} & 0.8927 & 0.8928 \\
            & & & {\bf 0.9201} & {\bf 0.9451} & 0.9445 & {\bf 0.9436} & 0.9428 & 0.9429 \\
         \cline{3-9}
         & & \multirow{4}{*}{New2} &
            571.14 & 242.55 & 225.26 & 223.77 & 224.12 & 224.32 \\
            & & & 20.56 & 24.28 & 24.60 & 24.63 & 24.63 & 24.62 \\
            & & & 0.7904 & 0.8645 & 0.8655 & 0.8651 & 0.8636 & 0.8634 \\
            & & & 0.917 & 0.9334 & 0.9298 & 0.9285 & 0.927 & 0.9268 \\
         \cline{3-9}
         & & \multirow{4}{*}{Integer Haar} &
            578.87 & 223.72 & 201.15 & 198.84 & 199.08 & 199.13 \\
            & & & 20.5 & 24.63 & 25.10 & 25.15 & 25.14 & 25.14 \\
            & & & 0.7879 & 0.8767 & 0.8810 & 0.8794 & 0.8787 & 0.8785 \\
            & & & 0.8696 & 0.9341 & 0.9342 & 0.9322 & 0.9315 & 0.9316 \\
         \hline
      \end{tabular}}{}
\end{table}

Table \ref{tab:07} shows results for the color image {\tt House}. The
PSNR values of all multiwavelet filters for CR = 4 and 8 for 4 and more
decomposition levels and for CR = 16 and 32 for 5 and 6 decomposition
levels are nearly constant, For all CRs, the best PSNR, SSIM and MS-SSIM
with 3 and more decomposition levels were achieved again by the New1
multiwavelet filter. Moreover, with CR = 4, its MS-SSIM values equal
those the of the SA4 multiwavelet filter, which is longer with more
vanishing moments.

% Table 7
\begin{table}
\centering 
  \renewcommand\thetable{7}
  \caption{Results for {\tt House}. The four lines for each multiwavelet represent MSE, PSNR, SSIM and MS-SSIM, in this order.}%
     {\label{tab:07}
       \footnotesize
       \setlength\tabcolsep{4pt}
       \renewcommand{\arraystretch}{1.0}
       \begin{tabular}{|c|c|c|c|c|c|c|c|c|}
         \hline
         \multirow{2}{*}{Image} &
         \multirow{2}{*}{CR} &
         Multi- &                       
         \multicolumn{6}{c|}{Decomposition level} \\
         \cline{4-9}
         & & wavelet & 1 & 2 & 3 & 4 & 5 & 6\\
         \hline
         \multirow{57}{*}{\parbox{1.5in}{\includegraphics[width=1.5in]{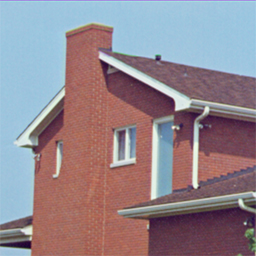}
         \\\centering {\tt House} }} & \multirow{28}{*}{4:1} &
         \multirow{4}{*}{GHM} & 
            62.454 & 6.664 & 5.9222 & 5.8497 & 5.8459 & 5.8459 \\
            & & & 30.43 & 39.95 & 40.46 & 40.51 & 40.51 & 40.51 \\
            & & & 0.8683 & 0.9547 & 0.9593 & 0.9598 & 0.9599 & 0.9599 \\
            & & & 0.9746 & 0.9949 & 0.9952 & 0.9952 & 0.9952 & 0.9952 \\
         \cline{3-9}
         & &  \multirow{4}{*}{SA4} &
            {\bf 4.9972} & 4.3791 & 4.3717 & 4.3892 & 4.3903 & 4.3899 \\
            & & & {\bf 41.22} & 41.79 & 41.80 & 41.78 & 41.78 & 41.78 \\
            & & & {\bf 0.9643} & {\bf 0.9688} & 0.9689 & 0.9688 & 0.9688 & 0.9688 \\
            & & & 0.9953 & {\bf 0.9957} & {\bf 0.9956} & {\bf 0.9956} & {\bf 0.9957} & {\bf 0.9957} \\
          \cline{3-9}
         & &  \multirow{4}{*}{CL} &
            5.0617 & 4.3374 & 4.2743 & 4.2803 & 4.2847 & 4.2846 \\
            & & & 41.16 & 41.82 & 41.89 & 41.88 & 41.88 & 41.88 \\
            & & & 0.9636 & 0.9686 & 0.9691 & 0.9691 & 0.9690 & 0.9690 \\
            & & & 0.9946 & 0.9952 & 0.9951 & 0.9951 & 0.9951 & 0.9951 \\
          \cline{3-9}
         & & \multirow{4}{*}{Alpert} &
            5.1499 & 4.3863 & 4.291 & 4.2871 & 4.2909 & 4.2912 \\
            & & & 41.07 & 41.76 & 41.85 & 41.86 & 41.85 & 41.85 \\
            & & & 0.9630 & 0.9682 & 0.9689 & 0.9690 & 0.9689 & 0.9689 \\
            & & & {\bf 0.9957} & {\bf 0.9956} & {\bf 0.9956} & {\bf 0.9956} & {\bf 0.9956} & {\bf 0.9956} \\
         \cline{3-9}
         & & \multirow{4}{*}{New1} &
            5.0737 & {\bf 4.3199} & {\bf 4.229} & {\bf 4.2339} & {\bf 4.2352} & {\bf 4.2352} \\
            & & & 41.14 & {\bf 41.83} & {\bf 41.92} & {\bf 41.92} & {\bf 41.92} & {\bf 41.92} \\
            & & & 0.9636 & {\bf 0.9687} & {\bf 0.9695} & {\bf 0.9694} & {\bf 0.9694} & {\bf 0.9694} \\
            & & & {\bf 0.9958} & {\bf 0.9957} & {\bf 0.9957} & {\bf 0.9957} & {\bf 0.9957} & {\bf 0.9957} \\
         \cline{3-9}
         & & \multirow{4}{*}{New2} &
            6.4048 & 5.848 & 5.8072 & 5.8293 & 5.8379 & 5.8389 \\
            & & & 40.09 & 40.48 & 40.51 & 40.49 & 40.48 & 40.48 \\
            & & & 0.9543 & 0.9580 & 0.9584 & 0.9582 & 0.9582 & 0.9581 \\
            & & & 0.9945 & 0.9940 & 0.9939 & 0.9939 & 0.9939 & 0.9939 \\
         \cline{3-9}
         & & \multirow{4}{*}{Integer Haar} &
            6.0433 & 5.3726 & 5.3174 & 5.3297 & 5.3319 & 5.3317 \\
            & & & 40.35 & 40.86 & 40.90 & 40.89 & 40.89 & 40.89 \\
            & & & 0.957 & 0.9615 & 0.9619 & 0.9618 & 0.9617 & 0.9617 \\
            & & & 0.9939 & 0.9941 & 0.9942 & 0.9942 & 0.9942 & 0.9942 \\
         \cline{2-9} \\
         \cline{2-9}
        & \multirow{28}{*}{8:1} &
         \multirow{4}{*}{GHM} & 
            5510.5 & 21.225 & 16.415 & 15.993 & 15.972 & 15.972 \\
            & & & 10.93 & 34.90 & 36.01 & 36.12 & 36.13 & 36.13 \\
            & & & 0.2712 & 0.8740 & 0.8956 & 0.8981 & 0.8982 & 0.8982 \\
            & & & 0.657 & 0.9824 & 0.9853 & 0.9856 & 0.9856 & 0.9856 \\
         \cline{3-9}
         & &  \multirow{4}{*}{SA4} &
            {\bf 16.127} & 12.194 & 12.057 & 12.15 & 12.167 & 12.169 \\
            & & & {\bf 36.10} & 37.33 & 37.38 & 37.35 & 37.34 & 37.34 \\
            & & & {\bf 0.8964} & {\bf 0.9200} & 0.9213 & 0.9207 & 0.9206 & 0.9206 \\
            & & & 0.9843 & 0.9874 & 0.9872 & 0.9871 & 0.9872 & 0.9872 \\
          \cline{3-9}
         & &  \multirow{4}{*}{CL} &
            16.409 & 12.007 & 11.671 & 11.686 & 11.709 & 11.707 \\
            & & & 36.02 & 37.39 & 37.51 & 37.51 & 37.50 & 37.50 \\
            & & & 0.8921 & 0.9188 & 0.9213 & 0.9212 & 0.9211 & 0.9211 \\
            & & & 0.9826 & 0.9865 & 0.9866 & 0.9865 & 0.9865 & 0.9865 \\
          \cline{3-9}
         & & \multirow{4}{*}{Alpert} &
            16.549 & 12.091 & 11.612 & 11.615 & 11.63 & 11.634 \\
            & & & 35.97 & 37.34 & 37.52 & 37.52 & 37.51 & 37.51 \\
            & & & 0.8920 & 0.9188 & 0.9223 & 0.9222 & 0.9221 & 0.9220 \\
            & & & 0.9866 & 0.9876 & 0.9878 & {\bf 0.9878} & 0.9878 & 0.9878 \\
         \cline{3-9}
         & & \multirow{4}{*}{New1} &
            16.37 & {\bf 11.96} & {\bf 11.525} & {\bf 11.534} & {\bf 11.537} & {\bf 11.541} \\
            & & & 36.02 & {\bf 37.39} & {\bf 37.56} & {\bf 37.55} & {\bf 37.55} & {\bf 37.55} \\
            & & & 0.8931 & 0.9197 & {\bf 0.9229} & {\bf 0.9228} & {\bf 0.9227} & {\bf 0.9227} \\
            & & & {\bf 0.9868} & {\bf 0.9879} & {\bf 0.9880} & {\bf 0.9879} & {\bf 0.9879} & {\bf 0.9879} \\
         \cline{3-9}
         & & \multirow{4}{*}{New2} &
            20.271 & 16..374 & 16.168 & 16.27 & 16.315 & 16.323 \\
            & & & 35.08 & 36.00 & 36.06 & 36.03 & 36.02 & 36.02 \\
            & & & 0.8770 & 0.8955 & 0.8970 & 0.8965 & 0.8963 & 0.8962 \\
            & & & 0.9843 & 0.9828 & 0.9826 & 0.9826 & 0.9825 & 0.9825 \\
         \cline{3-9}
         & & \multirow{4}{*}{Integer Haar} &
            19.548 & 14.883 & 14.521 & 14.609 & 14.609 & 14.613 \\
            & & & 35.24 & 36.42 & 36.53 & 36.51 & 36.50 & 36.50 \\
            & & & 0.8790 & 0.9033 & 0.9058 & 0.9051 & 0.9051 & 0.9051 \\
            & & & 0.9812 & 0.9837 & 0.9838 & 0.9837 & 0.9837 & 0.9837 \\
         \hline
      \end{tabular}}{}
\end{table}
% Table 7(continued)
\begin{table}
\centering 
  \renewcommand\thetable{7(continued)}
  \caption{Results for {\tt House}. The four lines for each multiwavelet represent MSE, PSNR, SSIM and MS-SSIM, in this order.}%
    {
       \footnotesize
       \setlength\tabcolsep{4pt}
       \renewcommand{\arraystretch}{1.0}
       \begin{tabular}{|c|c|c|c|c|c|c|c|c|}
         \hline
         \multirow{2}{*}{Image} &
         \multirow{2}{*}{CR} &
         Multi- &                       
         \multicolumn{6}{c|}{Decomposition level} \\
         \cline{4-9}
         & & wavelet & 1 & 2 & 3 & 4 & 5 & 6\\
         \hline
         \multirow{57}{*}{\parbox{1.5in}{\includegraphics[width=1.5in]{house}
         \\\centering {\tt House} }} & \multirow{28}{*}{16:1} &
         \multirow{4}{*}{GHM} & 
            11054 & 170.5 & 38.224 & 35.033 & 34.845 & 34.84 \\
            & & & 7.738 & 26.07 & 32.38 & 32.74 & 32.77 & 32.77 \\
            & & & 0.0391 & 0.7398 & 0.8271 & 0.8326 & 0.833 & 0.833 \\
            & & & 0.5900 & 0.9162 & 0.9691 & 0.9697 & 0.9696 & 0.9696 \\
         \cline{3-9}
         & &  \multirow{4}{*}{SA4} &
            198.44 & 26.90 & 25.683 & 26.03 & 26.116 & 26.126 \\
            & & & {\bf 25.37} & 33.88 & 34.08 & 34.02 & 34.01 & 34.01 \\
            & & & {\bf 0.7350} & 0.8481 & 0.8541 & 0.8529 & 0.8527 & 0.8525 \\
            & & & 0.9111 & 0.9727 & 0.9728 & 0.9721 & 0.9722 & 0.9722 \\
          \cline{3-9}
         & &  \multirow{4}{*}{CL} &
            190.88 & 26.63 & 24.528 & 24.533 & 24.663 & 24.662 \\
            & & & 25.51 & 33.91 & 34.27 & 34.27 & 34.24 & 34.24 \\
            & & & 0.7426 & 0.8448 & 0.8527 & 0.8529 & 0.8524 & 0.8524 \\
            & & & 0.9088 & 0.9712 & 0.9718 & 0.9715 & 0.9715 & 0.9715 \\
          \cline{3-9}
         & & \multirow{4}{*}{Alpert} &
            196.03 & 26.219 & 23.883 & 23.806 & 23.877 & 23.887 \\
            & & & 25.38 & 33.97 & 34.37 & 34.39 & 34.37 & 34.37 \\
            & & & 0.7221 & 0.8478 & 0.8573 & 0.8575 & 0.8571 & 0.8571 \\
            & & & 0.9476 & 0.9735 & 0.9742 & 0.9741 & 0.9741 & 0.9741 \\
         \cline{3-9}
         & & \multirow{4}{*}{New1} &
            {\bf 197.17} & {\bf 25.827} & {\bf 23.695} & {\bf 23.707} & {\bf 23.763} & {\bf 23.775} \\
            & & & {\bf 25.37} & {\bf 34.03} & {\bf 34.41} & {\bf 34.41} & {\bf 34.40} & {\bf 34.39} \\
            & & & 0.7222 & {\bf 0.8502} & {\bf 0.8591} & {\bf 0.8590} & {\bf 0.8588} & {\bf 0.8587} \\
            & & & 0.9477 & {\bf 0.9741} & {\bf 0.9747} & {\bf 0.9747} & {\bf 0.9748} & {\bf 0.9748} \\
         \cline{3-9}
         & & \multirow{4}{*}{New2} &
            201.1 & 37.967 & 35.901 & 36.254 & 36.519 & 36.56 \\
            & & & 25.29 & 32.36 & 32.60 & 32.56 & 32.53 & 32.53 \\
            & & & 0.7208 & 0.8214 & 0.8253 & 0.8244 & 0.8236 & 0.8234 \\
            & & & 0.9482 & 0.9656 & 0.9636 & 0.9634 & 0.9633 & 0.9633 \\
         \cline{3-9}
         & & \multirow{4}{*}{Integer Haar} &
            236.19 & 33.76 & 31.283 & 31.52 & 31.558 & 31.56 \\
            & & & 24.64 & 32.87 & 33.20 & 33.16 & 33.16 & 33.16 \\
            & & & 0.7222 & 0.8282 & 0.8336 & 0.8326 & 0.8323 & 0.8322 \\
            & & & 0.8881 & 0.9664 & 0.967 & 0.9666 & 0.9666 & 0.9666 \\
         \cline{2-9} \\
         \cline{2-9}
        & \multirow{28}{*}{32:1} &
         \multirow{4}{*}{GHM} & 
            15440 & 4641.1 & 102.12 & 80.223 & 78.789 & 78.700 \\
            & & & 6.261 & 11.60 & 28.26 & 29.27 & 29.34 & 29.35 \\
            & & & 0.02841 & 0.2525 & 0.7605 & 0.7766 & 0.7779 & 0.7779 \\
            & & & 0.5604 & 0.5453 & 0.9369 & 0.9433 & 0.9433 & 0.9434 \\
         \cline{3-9}
         & &  \multirow{4}{*}{SA4} &
            {\bf 5465.5} & {\bf 64.412} & 53.237 & 53.817 & 54.285 & 54.317 \\
            & & & {\bf 10.97} & {\bf 30.15} & 30.96 & 30.92 & 30.88 & 30.88 \\
            & & & {\bf 0.3528} & {\bf 0.7860} & 0.7956 & 0.7940 & 0.7935 & 0.7935 \\
            & & & 0.6279 & 0.9507 & 0.9518 & 0.9485 & 0.9481 & 0.9482 \\
          \cline{3-9}
         & &  \multirow{4}{*}{CL} &
            5586.6 & 73.51 & 53.266 & 52.708 & 53.512 & 53.516 \\
            & & & 10.88 & 29.58 & 30.94 & 30.99 & 30.93 & 30.93 \\
            & & & 0.3526 & 0.7792 & 0.7958 & 0.7961 & 0.7948 & 0.7949 \\
            & & & {\bf 0.6347} & 0.9485 & 0.9510 & 0.9509 & 0.9500 & 0.9500 \\
          \cline{3-9}
         & & \multirow{4}{*}{Alpert} &
            5710.4 & 66.371 & 48.518 & 47.634 & 47.921 & 48.02 \\
            & & & 10.79 & 30.00 & 31.33 & 31.41 & 31.38 & 31.37 \\
            & & & 0.3419 & 0.7828 & 0.8017 & 0.8025 & 0.802 & 0.8017 \\
            & & & 0.6331 & 0.9543 & 0.9561 & 0.9551 & 0.9550 & 0.9550 \\
         \cline{3-9}
         & & \multirow{4}{*}{New1} &
            5710.4 & 64.631 & {\bf 47.651} & {\bf 47.051} & {\bf 47.361} & {\bf 47.434} \\
            & & & 10.79 & 30.12 & {\bf 31.41} & {\bf 31.46} & {\bf 31.44} & {\bf 31.43} \\
            & & & 0.3419 & 0.7853 & {\bf 0.8037} & {\bf 0.8041} & {\bf 0.8033} & {\bf 0.8032} \\
            & & & 0.6331 & {\bf 0.9551} & {\bf 0.9564} & {\bf 0.9555} & {\bf 0.9553} & {\bf 9551}\\
         \cline{3-9}
         & & \multirow{4}{*}{New2} &
            5710.4 & 103.05 & 80.823 & 81.065 & 82.416 & 82.71 \\
            & & & 10.79 & 28.12 & 29.14 & 29.13 & 29.06 & 29.04 \\
            & & & 0.3419 & 0.7521 & 0.7643 & 0.7627 & 0.7605 & 0.7599 \\
            & & & 0.6331 & 0.9425 & 0.9397 & 0.9372 & 0.9358 & 0.9355 \\
         \cline{3-9}
         & & \multirow{4}{*}{Integer Haar} &
            5738.7 & 90.77 & 68.696 & 68.174 & 68.329 & 68.398 \\
            & & & 10.76 & 28.65 & 29.83 & 29.86 & 29.85 & 29.85 \\
            & & & 0.3423 & 0.7594 & 0.7731 & 0.7706 & 0.7700 & 0.7698 \\
            & & & 0.6235 & 0.9421 & 0.9453 & 0.9431 & 0.9426 & 0.9427 \\
         \hline
      \end{tabular}}{}
\end{table}

Table \ref{tab:08} shows results for the color image {\tt Cara}. The
best measured values for CR = 4 of all decomposition levels are achieved
by the New1 multiwavelet. The SSIM and MS-SSIM values of GHM, SA4 and
Alpert multiwavelets are equal for 2 and more levels. We need to note
that for CR = 4 and 8 at 3 and more decomposition levels all
multifilters have nearly constant SSIM and MS-SSIM values. For CR = 8,
best measured values are achieved again by the New1 multiwavelet, except
for 1 and 2 decomposition levels, and the MS-SSIM values are equal with
GHM and Alpert for 3 and more levels. For CR = 16, for 1 and 2
decomposition levels the SA4 multifilter has the best SSIM values, for
the rest the best measured values are achieved by the New1
multiwavelet. For CR = 32, the New1 multiwavelet achieves all best
measured values, except MSE and PSNR at 1, 2 and 3 decomposition
levels. The MS-SSIM values of the supercompact multiwavelet filters are
nearly constant for more than 4 decomposition levels.

% Table 8
\begin{table}
\centering 
  \renewcommand\thetable{8}
  \caption{Results for {\tt Cara}. The four lines for each multiwavelet represent MSE, PSNR, SSIM and MS-SSIM, in this order.}%
     {\label{tab:08}
       \footnotesize
       \setlength\tabcolsep{4pt}
       \renewcommand{\arraystretch}{1.0}
       \begin{tabular}{|c|c|c|c|c|c|c|c|c|}
         \hline
         \multirow{2}{*}{Image} &
         \multirow{2}{*}{CR} &
         Multi- &                       
         \multicolumn{6}{c|}{Decomposition level} \\
         \cline{4-9}
         & & wavelet & 1 & 2 & 3 & 4 & 5 & 6\\
         \hline
         \multirow{57}{*}{\parbox{1.5in}{\includegraphics[width=1.5in]{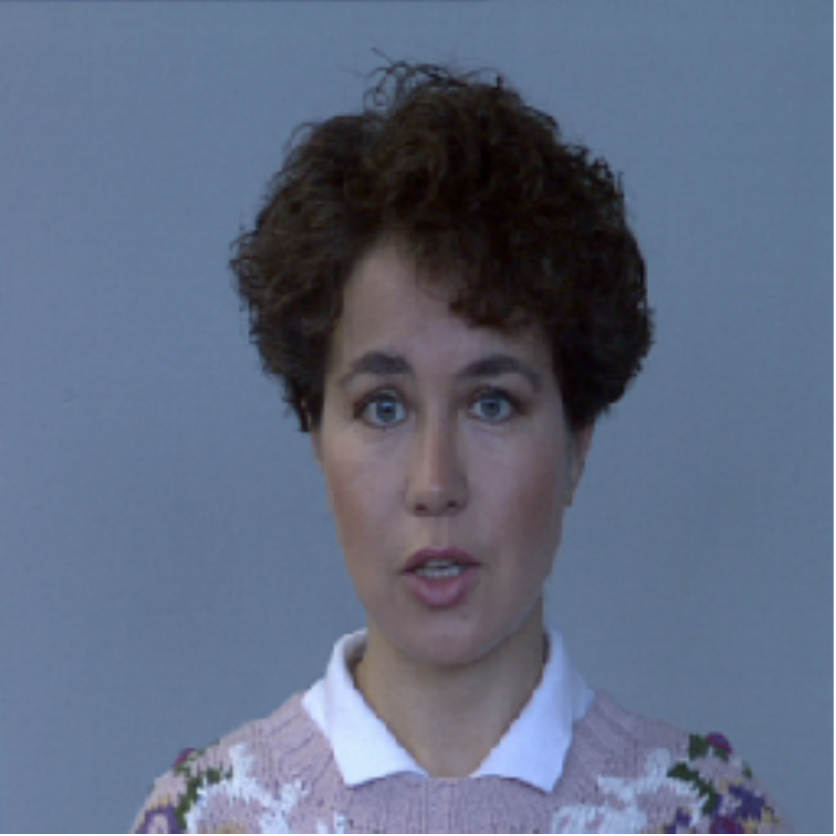}
         \\\centering {\tt Cara} }} & \multirow{28}{*}{4:1} &
         \multirow{4}{*}{GHM} & 
            14.669 & 1.5443 & 1.3911 & 1.3727 & 1.372 & 1.3719 \\
            & & & 36.47 & 46.27 & 46.72 & 46.78 & 46.78 & 46.78 \\
            & & & 0.9471 & 0.9829 & 0.9841 & 0.9843 & 0.9843 & 0.9843 \\
            & & & 0.9927 & 0.998 & {\bf 0.9981} & {\bf 0.9981} & {\bf 0.9981} & {\bf 0.9981} \\
         \cline{3-9}
         & &  \multirow{4}{*}{SA4} &
            1.2479 & 1.1258 & 1.1206 & 1.1208 & 1.1209 & 1.121 \\
            & & & 47.19 & 47.64 & 47.66 & 47.66 & 47.66 & 47.65 \\
            & & & 0.9847 & 0.9860 & 0.9861 & 0.9860 & 0.9860 & 0.9860 \\
            & & & 0.9979 & {\bf 0.9981} & {\bf 0.9981} & {\bf 0.9981} & {\bf 0.9981} & {\bf 0.9981} \\
          \cline{3-9}
         & &  \multirow{4}{*}{CL} &
            1.2791 & 1.1236 & 1.1074 & 1.1065 & 1.1052 & 1.1052 \\
            & & & 47.08 & 47.65 & 47.71 & 47.71 & 47.72 & 47.72 \\
            & & & 0.9843 & 0.9858 & 0.9860 & 0.9860 & 0.9860 & 0.9860 \\
            & & & 0.9976 & 0.9978 & 0.9978 & 0.9978 & 0.9978 & 0.9978 \\
          \cline{3-9}
         & & \multirow{4}{*}{Alpert} &
            1.2582 & 1.1111 & 1.0896 & 1.0852 & 1.0847 & 1.0846 \\
            & & & 47.16 & 47.70 & 47.78 & 47.80 & 47.80 & 47.80 \\
            & & & 0.9847 & 0.9861 & 0.9864 & 0.9864 & 0.9864 & 0.9864 \\
            & & & {\bf 0.9980} & 0.9980 & 0.9980 & 0.9980 & 0.9980 & 0.9980 \\
         \cline{3-9}
         & & \multirow{4}{*}{New1} &
            {\bf 1.2397} & {\bf 1.0961} & {\bf 1.0759} & {\bf 1.072} & {\bf 1.0717} & {\bf 1.0716} \\
            & & & {\bf 47.22} & {\bf 47.76} & {\bf 47.84} & {\bf 47.85} & {\bf 47.85} & {\bf 47.85} \\
            & & & {\bf 0.9849} & {\bf 0.9863} & {\bf 0.9866} & {\bf 0.9866} & {\bf 0.9866} & {\bf 0.9866} \\
            & & & {\bf 0.9980} & {\bf 0.9981} & {\bf 0.9981} & {\bf 0.9981} & {\bf 0.9981} & {\bf 0.9981} \\
         \cline{3-9}
         & & \multirow{4}{*}{New2} &
            1.5073 & 1.3779 & 1.3716 & 1.3728 & 1.3728 & 1.3733 \\
            & & & 46.37 & 46.76 & 46.79 & 46.78 & 46.78 & 46.78 \\
            & & & 0.982 & 0.9832 & 0.9832 & 0.9832 & 0.9832 & 0.9832 \\
            & & & 0.9975 & 0.9973 & 0.9973 & 0.9973 & 0.9973 & 0.9973 \\
         \cline{3-9}
         & & \multirow{4}{*}{Integer Haar} &
            1.4994 & 1.332 & 1.3149 & 1.3131 & 1.3131 & 1.3133 \\
            & & & 46.39 & 46.91 & 46.96 & 46.97 & 46.97 & 46.97 \\
            & & & 0.9821 & 0.9836 & 0.9837 & 0.9837 & 0.9837 & 0.9837 \\
            & & & 0.9972 & 0.9974 & 0.9974 & 0.9974 & 0.9974 & 0.9974 \\
         \cline{2-9} \\
         \cline{2-9}
        & \multirow{28}{*}{8:1} &
         \multirow{4}{*}{GHM} & 
            4659.8 & 5.9153 & 4.0861 & 3.9051 & 3.8908 & 3.8907 \\
            & & & 11.56 & 40.44 & 42.05 & 42.24 & 42.26 & 42.26 \\
            & & & 0.05861 & 0.9611 & 0.968 & 0.9688 & 0.9689 & 0.9689 \\
            & & & 0.5297 & 0.9945 & {\bf 0.9952} & {\bf 0.9952} & {\bf 0.9952} & {\bf 0.9952} \\
         \cline{3-9}
         & &  \multirow{4}{*}{SA4} &
            4.7746 & 3.2036 & 3.1461 & 3.1508 & 3.1576 & 3.1579 \\
            & & & {\bf 41.37} & 43.10 & 43.18 & 43.17 & 43.17 & 43.16 \\
            & & & {\bf 0.9638} & 0.9708 & 0.9713 & 0.9713 & 0.9712 & 0.9712 \\
            & & & 0.9946 & {\bf 0.9953} & 0.9951 & 0.9951 & 0.9951 & 0.9951 \\
          \cline{3-9}
         & &  \multirow{4}{*}{CL} &
            5.0704 & 3.2022 & 3.0691 & 3.0491 & 3.0534 & 3.0538 \\
            & & & 41.11 & 43.10 & 43.29 & 43.32 & 43.31 & 43.31 \\
            & & & 0.9619 & 0.9702 & 0.9711 & 0.9712 & 0.9711 & 0.9711 \\
            & & & 0.9940 & 0.9949 & 0.9948 & 0.9949 & 0.9949 & 0.9949 \\
          \cline{3-9}
         & & \multirow{4}{*}{Alpert} &
            4.8966 & 3.1377 & 2.9816 & 2.9554 & 2.9505 & 2.9518 \\
            & & & 41.26 & 43.20 & 43.42 & 43.46 & 43.46 & 43.46 \\
            & & & 0.9628 & 0.9710 & 0.9718 & 0.9719 & 0.9719 & 0.9719 \\
            & & & 0.9951 & 0.9950 & 0.9951 & {\bf 0.9952} & {\bf 0.9952} & {\bf 0.9952} \\
         \cline{3-9}
         & & \multirow{4}{*}{New1} &
            {\bf 4.7765} & {\bf 3.0599} & {\bf 2.9094} & {\bf 2.8828} & {\bf 2.8794} & {\bf 2.8819} \\
            & & & {\bf 41.37} & {\bf 43.30} & {\bf 43.52} & {\bf 43.56} & {\bf 43.57} & {\bf 43.56} \\
            & & & 0.9634 & {\bf 0.9715} & {\bf 0.9723} & {\bf 0.9724} & {\bf 0.9725} & {\bf 0.9724} \\
            & & & {\bf 0.9952} & 0.9951 & {\bf 0.9952} & {\bf 0.9952} & {\bf 0.9952} & {\bf 0.9952} \\
         \cline{3-9}
         & & \multirow{4}{*}{New2} &
            6.6138 & 4.7232 & 4.5804 & 4.6141 & 4.619 & 4.624 \\
            & & & 39.96 & 41.42 & 41.56 & 41.53 & 41.52 & 41.52 \\
            & & & 0.9538 & 0.9616 & 0.9623 & 0.9619 & 0.9618 & 0.9618 \\
            & & & 0.9935 & 0.9934 & 0.9930 & 0.9930 & 0.9930 & 0.9930 \\
         \cline{3-9}
         & & \multirow{4}{*}{Integer Haar} &
            6.7467 & 4.3785 & 4.1929 & 4.1946 & 4.1969 & 4.1973 \\
            & & & 39.87 & 41.75 & 41.94 & 41.94 & 41.93 & 41.93 \\
            & & & 0.9542 & 0.9637 & 0.9647 & 0.9644 & 0.9644 & 0.9644 \\
            & & & 0.9925 & 0.9938 & 0.9938 & 0.9940 & 0.9940 & 0.9940 \\
         \hline
      \end{tabular}}{}
\end{table}
% Table 8(continued)
\begin{table}
  \centering 
  \renewcommand\thetable{8(continued)}
  \caption{Results for {\tt Cara}. The four lines for each multiwavelet represent MSE, PSNR, SSIM and MS-SSIM, in this order.}%
  {
    \footnotesize
    \setlength\tabcolsep{4pt}
    \renewcommand{\arraystretch}{1.0}
    \begin{tabular}{|c|c|c|c|c|c|c|c|c|}
      \hline
      \multirow{2}{*}{Image} &
      \multirow{2}{*}{CR} &
      Multi- &                       
      \multicolumn{6}{c|}{Decomposition level} \\
      \cline{4-9}
      & & wavelet & 1 & 2 & 3 & 4 & 5 & 6\\
      \hline
      \multirow{57}{*}{\parbox{1.5in}{\includegraphics[width=1.5in]{cara}
      \\\centering {\tt Cara} }} & \multirow{28}{*}{16:1} &
      \multirow{4}{*}{GHM} & 
         8248.7 & 38.541 & 11.345 & 10.24 & 10.138 & 10.135 \\
         & & & 9.065 & 32.29 & 37.62 & 38.07 & 38.11 & 38.12 \\
         & & & 0.03618 & 0.8883 & 0.9421 & 0.9453 & 0.9457 & 0.9457 \\
         & & & 0.503 & 0.9728 & 0.9894 & 0.9899 & 0.9899 & 0.9899 \\
      \cline{3-9}
      & &  \multirow{4}{*}{SA4} &
         44.06 & 9.0035 & 8.2593 & 8.3198 & 8.3657 & 8.3696 \\
         & & & 31.71 & 38.62 & 39.00 & 38.97 & 38.95 & 38.94 \\
         & & & {\bf 0.8648} & {\bf 0.9475} & 0.9496 & 0.9495 & 0.9492 & 0.9492 \\
         & & & 0.9668 & 0.9902 & 0.9895 & 0.9895 & 0.9895 & 0.9895 \\
       \cline{3-9}
      & &  \multirow{4}{*}{CL} &
         47.142 & 9.0648 & 8.0847 & 8.0233 & 8.0603 & 8.0547 \\
         & & & 31.42 & 38.59 & 39.09 & 39.12 & 39.10 & 39.10 \\
         & & & 0.8598 & 0.9462 & 0.9494 & 0.9496 & 0.9495 & 0.9495 \\
         & & & 0.9691 & 0.9896 & 0.9894 & 0.9896 & 0.9897 & 0.9897 \\
       \cline{3-9}
      & & \multirow{4}{*}{Alpert} &
         43.916 & 8.9286 & 7.7402 & 7.6756 & 7.6529 & 7.6628 \\
         & & & 31.73 & 38.66 & 39.28 & 39.32 & 39.33 & 39.33 \\
         & & & 0.8621 & 0.9463 & 0.9503 & 0.9504 & 0.9505 & 0.9504 \\
         & & & 0.9751 & 0.9905 & 0.9904 & 0.9909 & 0.9909 & 0.9909 \\
      \cline{3-9}
      & & \multirow{4}{*}{New1} &
         {\bf 43.581} & {\bf 8.7188} & {\bf 7.5929} & {\bf 7.5075} & {\bf 7.4977} & {\bf 7.5047} \\
         & & & {\bf 31.77} & {\bf 38.76} & {\bf 39.36} & {\bf 39.41} & {\bf 39.42} & {\bf 39.41} \\
         & & & 0.8625 & {\bf 0.9474} & {\bf 0.9512} & {\bf 0.9514} & {\bf 0.9514} & {\bf 0.9514} \\
         & & & {\bf 0.9753} & {\bf 0.9907} & {\bf 0.9906} & {\bf 0.9910} & {\bf 0.9910} & {\bf 0.9910} \\
      \cline{3-9}
      & & \multirow{4}{*}{New2} &
         48.679 & 14.159 & 13.083 & 13.242 & 13.257 & 13.293 \\
         & & & 31.30 & 36.67 & 37.01 & 36.96 & 36.96 & 36.94 \\
         & & & 0.8571 & 0.9265 & 0.9291 & 0.9283 & 0.9282 & 0.9280 \\
         & & & 0.9739 & 0.9852 & 0.9849 & 0.9846 & 0.9846 & 0.9846 \\
      \cline{3-9}
      & & \multirow{4}{*}{Integer Haar} &
         54.35 & 13.133 & 11.83 & 11.86 & 11.861 & 11.881 \\
         & & & 30.81 & 36.99 & 37.44 & 37.43 & 37.43 & 37.43 \\
         & & & 0.8485 & 0.9311 & 0.9344 & 0.9338 & 0.9336 & 0.9336 \\
         & & & 0.9602 & 0.9862 & 0.9864 & 0.9864 & 0.9865 & 0.9865 \\
      \cline{2-9} \\
      \cline{2-9}
     & \multirow{28}{*}{32:1} &
      \multirow{4}{*}{GHM} & 
         10501 & 3488 & 27.742 & 22.497 & 22.009 & 21.981 \\
         & & & 8.051 & 12.82 & 33.75 & 34.66 & 34.76 & 34.76 \\
         & & & 0.01585 & 0.06779 & 0.9044 & 0.9153 & 0.9165 & 0.9165 \\
         & & & 0.521 & 0.3933 & 0.9764 & 0.9796 & 0.9796 & 0.9796 \\
      \cline{3-9}
      & &  \multirow{4}{*}{SA4} &
         {\bf 4664.8} & {\bf 22.996} & 18.123 & 18.228 & 18.414 & 18.425 \\
         & & & {\bf 11.57} & {\bf 34.56} & 35.60 & 35.58 & 35.54 & 35.53 \\
         & & & 0.3833 & {\bf 0.9111} & {\bf 0.9221} & 0.9216 & 0.9210 & 0.9210 \\
         & & & 0.5896 & 0.9780 & 0.9799 & 0.9783 & 0.9783 & 0.9784 \\
       \cline{3-9}
      & &  \multirow{4}{*}{CL} &
         4721.1 & 24.749 & 18.57 & 18.076 & 18.272 & 18.254 \\
         & & & 11.52 & 34.25 & 35.49 & 35.61 & 35.56 & 35.57 \\
         & & & 0.3841 & 0.9051 & 0.9191 & 0.9203 & 0.9197 & 0.9198 \\
         & & & 0.5860 & 0.9765 & 0.9800 & 0.9798 & 0.9798 & 0.9798 \\
       \cline{3-9}
      & & \multirow{4}{*}{Alpert} &
         4761 & 23.861 & 17.324 & 17.001 & 16.96 & 17.015 \\
         & & & 11.49 & 34.41 & 35.80 & 35.88 & 35.89 & 35.88 \\
         & & & 0.3867 & 0.905 & 0.9204 & 0.921 & 0.9211 & 0.9208 \\
         & & & {\bf 0.6056} & 0.9789 & 0.9823 & 0.9822 & 0.9826 & 0.9826 \\
      \cline{3-9}
      & & \multirow{4}{*}{New1} &
         4761.00 & 23.364 & {\bf 17.139} & {\bf 16.803} & {\bf 16.787} & {\bf 16.845} \\
         & & & 11.49 & 34.50 & {\bf 35.84} & {\bf 35.93} & {\bf 35.93} & {\bf 35.92} \\
         & & & {\bf 0.3867} & 0.9070 & 0.9218 & {\bf 0.9228} & {\bf 0.9228} & {\bf 0.9226} \\
         & & & {\bf 0.6056} & {\bf 0.9795} & {\bf 0.9826} & {\bf 0.9827} & {\bf 0.9831} & {\bf 0.9831} \\
      \cline{3-9}
      & & \multirow{4}{*}{New2} &
         4761 & 35.318 & 29.309 & 29.573 & 29.675 & 29.859 \\
         & & & 11.49 & 32.73 & 33.53 & 33.50 & 33.48 & 33.45 \\
         & & & 0.3867 & 0.8801 & 0.8888 & 0.8871 & 0.8865 & 0.886 \\
         & & & {\bf 0.6056} & 0.9710 & 0.9719 & 0.9696 & 0.9694 & 0.9694 \\
      \cline{3-9}
      & & \multirow{4}{*}{Integer Haar} &
         4770.8 & 33.309 & 26.202 & 26.049 & 26.042 & 26.146 \\
         & & & 11.48 & 32.97 & 34.01 & 34.03 & 34.03 & 34.02 \\
         & & & 0.3792 & 0.8852 & 0.8964 & 0.8953 & 0.8950 & 0.8946 \\
         & & & 0.5835 & 0.9703 & 0.9740 & 0.9729 & 0.9730 & 0.9731 \\
      \hline
    \end{tabular}
  }
  {}
\end{table}
  
Table \ref{tab:9} shows results for the color image {\tt Voit}. For CR =
4 and 8, all multiwavelet filters at 4 and more decomposition levels
have nearly constants measured values. With a minimal difference from
the SA4 multiwavelet, all best measured values are achieved again by the
New1 multiwavelet filter. For CR = 4 and 8, the New1 multiwavelet filter
has the best MSE, PSNR and SSIM for 3 and more decomposition levels. The
best MS-SSIM, is achieved by the SA4 multiwavelet, except at the 2nd
level. For CR = 32, from 1 to 3 decomposition level the best SSIM is
with the SA4 multiwavelet filter, while the best MS-SSIM values are for
CL, New1 and SA4 multiwavelets, respectively. For 4 and more
decomposition level, the New1 multiwavelet has the best SSIM and MS–SSIM
values, while its PSNR values are equal with Alpert multiwavelet.

% Table 9
\begin{table}
\centering 
  \renewcommand\thetable{9}
  \caption{Results for {\tt Voit}. The four lines for each multiwavelet represent MSE, PSNR, SSIM and MS-SSIM, in this order.}%
     {\label{tab:9}
       \footnotesize
       \setlength\tabcolsep{4pt}
       \renewcommand{\arraystretch}{1.0}
       \begin{tabular}{|c|c|c|c|c|c|c|c|c|}
         \hline
         \multirow{2}{*}{Image} &
         \multirow{2}{*}{CR} &
         Multi- &                       
         \multicolumn{6}{c|}{Decomposition level} \\
         \cline{4-9}
         & & wavelet & 1 & 2 & 3 & 4 & 5 & 6\\
         \hline
         \multirow{57}{*}{\parbox{1.5in}{\includegraphics[width=1.5in]{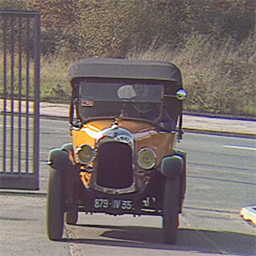}
         \\\centering {\tt Voit} }} & \multirow{28}{*}{4:1} &
         \multirow{4}{*}{GHM} & 
            192.89 & 34.812 & 31.389 & 31.084 & 31.061 & 31.056 \\
            & & & 25.29 & 32.73 & 33.18 & 33.22 & 33.22 & 33.22 \\
            & & & 0.7299 & 0.8896 & 0.898 & 0.8988 & 0.8989 & 0.8989 \\
            & & & 0.9547 & 0.9867 & 0.9876 & 0.9877 & 0.9877 & 0.9877 \\
         \cline{3-9}
         & &  \multirow{4}{*}{SA4} &
            27.374 & 24.717 & 24.37 & 24.369 & 24.377 & 24.378 \\
            & & & 33.77 & 34.22 & 34.28 & 34.28 & 34.27 & 34.27 \\
            & & & {\bf 0.9055} & 0.9136 & 0.9147 & 0.9148 & 0.9148 & 0.9148 \\
            & & & 0.9874 & 0.9881 & {\bf 0.9882} & {\bf 0.9882} & {\bf 0.9882} & {\bf 0.9882} \\
          \cline{3-9}
         & &  \multirow{4}{*}{CL} &
            27.639 & 24.311 & 23.967 & 23.917 & 23.932 & 23.931 \\
            & & & 33.73 & 34.29 & 34.35 & 34.36 & 34.35 & 34.35 \\
            & & & 0.9032 & 0.9134 & 0.9147 & 0.9149 & 0.9148 & 0.9148 \\
            & & & 0.9854 & 0.9865 & 0.9865 & 0.9866 & 0.9866 & 0.9866 \\
          \cline{3-9}
         & & \multirow{4}{*}{Alpert} &
            27.554 & 24.036 & 23.544 & 23.432 & 23.424 & 23.429 \\
            & & & 33.74 & 34.33 & 34.42 & 34.44 & 34.45 & 34.44 \\
            & & & 0.9036 & 0.9136 & 0.915 & 0.9155 & 0.9155 & 0.9154 \\
            & & & 0.9888 & 0.9881 & 0.9879 & 0.9879 & 0.9878 & 0.9879 \\
         \cline{3-9}
         & & \multirow{4}{*}{New1} &
            {\bf 27.205} & {\bf 23.891} & {\bf 23.427} & {\bf 23.347} & {\bf 23.338} & {\bf 23.342} \\
            & & & {\bf 33.80} & {\bf 34.36} & {\bf 34.45} & {\bf 34.46} & {\bf 34.46} & {\bf 34.46} \\
            & & & 0.9047 & {\bf 0.9144} & {\bf 0.9159} & {\bf 0.9162} & {\bf 0.9163} & {\bf 0.9163} \\
            & & & {\bf 0.9891} & {\bf 0.9883} & {\bf 0.9882} & 0.9881 & 0.9881 & 0.9881 \\
         \cline{3-9}
         & & \multirow{4}{*}{New2} &
            30.78 & 28.92 & 29.171 & 29.252 & 29.260 & 29.267 \\
            & & & 33.26 & 33.54 & 33.50 & 33.49 & 33.49 & 33.49 \\
            & & & 0.896 & 0.9014 & 0.9011 & 0.9009 & 0.9009 & 0.9009 \\
            & & & 0.9874 & 0.9853 & 0.9845 & 0.9845 & 0.9846 & 0.9846 \\
         \cline{3-9}
         & & \multirow{4}{*}{Integer Haar} &
            29.85 & 26.893 & 26.702 & 26.65 & 26.652 & 26.655 \\
            & & & 33.39 & 33.85 & 33.88 & 33.89 & 33.89 & 33.89 \\
            & & & 0.8971 & 0.9053 & 0.906 & 0.9061 & 0.9062 & 0.9061 \\
            & & & 0.9847 & 0.9856 & 0.9856 & 0.9856 & 0.9855 & 0.9855 \\
         \cline{2-9} \\
         \cline{2-9}
        & \multirow{28}{*}{8:1} &
         \multirow{4}{*}{GHM} & 
            3941.9 & 105.44 & 82.277 & 80.29 & 80.138 & 80.112 \\
            & & & 12.20 & 27.93 & 29.01 & 29.11 & 29.12 & 29.12 \\
            & & & 0.2293 & 0.7473 & 0.7846 & 0.789 & 0.7894 & 0.7894 \\
            & & & 0.6774 & 0.9559 & 0.9632 & 0.9637 & 0.9638 & 0.9638 \\
         \cline{3-9}
         & &  \multirow{4}{*}{SA4} &
            {\bf 82.81} & 64.558 & 62.664 & 62.671 & 62.667 & 62.688 \\
            & & & {\bf 28.97} & 30.06 & 30.18 & 30.18 & 30.18 & 30.18 \\
            & & & {\bf 0.7800} & {\bf 0.8197} & 0.8242 & 0.8242 & 0.8242 & 0.8241 \\
            & & & {\bf 0.9581} & 0.9668 & {\bf 0.9669} & {\bf 0.9669} & {\bf 0.9669} & {\bf 0.9669} \\
          \cline{3-9}
         & &  \multirow{4}{*}{CL} &
            85.127 & 64.228 & 62.134 & 61.891 & 61.934 & 61.933 \\
            & & & 28.86 & 30.08 & 30.22 & 30.24 & 30.24 & 30.24 \\
            & & & 0.7739 & 0.8167 & 0.8216 & 0.8223 & 0.8222 & 0.8222 \\
            & & & 0.9543 & 0.9630 & 0.9636 & 0.9636 & 0.9636 & 0.9636 \\
          \cline{3-9}
         & & \multirow{4}{*}{Alpert} &
            85.654 & 63.808 & 61.202 & 60.714 & 60.674 & 60.696 \\
            & & & 28.83 & 30.10 & 30.29 & 30.32 & 30.32 & 30.32 \\
            & & & 0.7712 & 0.8172 & 0.8228 & 0.8239 & 0.824 & 0.8239 \\
            & & & 0.9665 & 0.9665 & 0.9661 & 0.966 & 0.966 & 0.966 \\
         \cline{3-9}
         & & \multirow{4}{*}{New1} &
            84.412 & {\bf 63.328} & {\bf 60.746} & {\bf 60.379} & {\bf 60.354} & {\bf 60.379} \\
            & & & 28.89 & {\bf 30.14} & {\bf 30.32} & {\bf 30.35} & {\bf 30.35} & {\bf 30.35} \\
            & & & 0.774 & 0.8191 & {\bf 0.8247} & {\bf 0.8256} & {\bf 0.8257} & {\bf 0.8256} \\
            & & & 0.9672 & {\bf 0.9672} & 0.9666 & 0.9667 & 0.9667 & 0.9667 \\
         \cline{3-9}
         & & \multirow{4}{*}{New2} &
            93.636 & 75.327 & 74.947 & 75.224 & 75.275 & 75.314 \\
            & & & 28.44 & 29.39 & 29.41 & 29.40 & 29.39 & 29.39 \\
            & & & 0.757 & 0.7939 & 0.7959 & 0.7953 & 0.7952 & 0.7951 \\
            & & & 0.9626 & 0.9588 & 0.957 & 0.9567 & 0.9567 & 0.9567 \\
         \cline{3-9}
         & & \multirow{4}{*}{Integer Haar} &
            90.456 & 69.434 & 67.948 & 67.661 & 67.633 & 67.637 \\
            & & & 28.59 & 29.74 & 29.83 & 29.85 & 29.85 & 29.85 \\
            & & & 0.7632 & 0.804 & 0.8067 & 0.8072 & 0.8073 & 0.8073 \\
            & & & 0.9514 & 0.9604 & 0.9603 & 0.9605 & 0.9605 & 0.9605 \\
         \hline
      \end{tabular}}
  \end{table}
  
% Table 9(continued)
\begin{table}
\centering 
  \renewcommand\thetable{9(continued)}
  \caption{Results for {\tt Voit}. The four lines for each multiwavelet represent MSE, PSNR, SSIM and MS-SSIM, in this order.}%
    {
       \footnotesize
       \setlength\tabcolsep{4pt}
       \renewcommand{\arraystretch}{1.0}
       \begin{tabular}{|c|c|c|c|c|c|c|c|c|}
         \hline
         \multirow{2}{*}{Image} &
         \multirow{2}{*}{CR} &
         Multi- &                       
         \multicolumn{6}{c|}{Decomposition level} \\
         \cline{4-9}
         & & wavelet & 1 & 2 & 3 & 4 & 5 & 6\\
         \hline
         \multirow{57}{*}{\parbox{1.5in}{\includegraphics[width=1.5in]{voit}
         \\\centering {\tt Voit} }} & \multirow{28}{*}{16:1} &
         \multirow{4}{*}{GHM} & 
            7554.6 & 375.82 & 164.29 & 154.67 & 153.91 & 153.84 \\
            & & & 9.367 & 22.39 & 26.00 & 26.27 & 26.29 & 26.29 \\
            & & & 0.0826 & 0.5001 & 0.6575 & 0.6707 & 0.6718 & 0.6719 \\
            & & & 0.6229 & 0.8494 & 0.9211 & 0.9228 & 0.9232 & 0.9232 \\
         \cline{3-9}
         & &  \multirow{4}{*}{SA4} &
            {\bf 395.03} & 129.48 & 121.36 & 121.23 & 121.23 & 121.27 \\
            & & & {\bf 22.17} & 27.04 & 27.32 & 27.32 & 27.32 & 27.32 \\
            & & & {\bf 0.5099} & {\bf 0.6940} & 0.7097 & 0.7098 & 0.7099 & 0.7098 \\
            & & & 0.8262 & 0.9278 & 0.9293 & 0.9295 & 0.9295 & 0.9295 \\
          \cline{3-9}
         & &  \multirow{4}{*}{CL} &
            405.90 & 130.05 & 120.54 & 119.55 & 119.59 & 119.62 \\
            & & & 22.06 & 27.02 & 27.35 & 27.38 & 27.38 & 27.38 \\
            & & & 0.5011 & 0.6914 & 0.7073 & 0.7092 & 0.7095 & 0.7094 \\
            & & & 0.8423 & 0.9208 & 0.9236 & 0.9235 & 0.9233 & 0.9233 \\
          \cline{3-9}
         & & \multirow{4}{*}{Alpert} &
            415.65 & 128.04 & 118.22 & 116.63 & 116.52 & 116.58 \\
            & & & 21.97 & 27.08 & 27.43 & 27.49 & 27.49 & 27.49 \\
            & & & 0.4804 & 0.6921 & 0.7084 & 0.7112 & 0.7115 & 0.7113 \\
            & & & 0.8864 & 0.9269 & 0.9293 & 0.9300 & 0.9299 & 0.9299 \\
         \cline{3-9}
         & & \multirow{4}{*}{New1} &
            414.61 & {\bf 127.11} & {\bf 117.48} & {\bf 116.14} & {\bf 116.05} & {\bf 116.12} \\
            & & & 21.98 & {\bf 27.12} & {\bf 27.46} & {\bf 27.51} & {\bf 27.51} & {\bf 27.51} \\
            & & & 0.4808 & {\bf 0.6939} & {\bf 0.7108} & {\bf 0.7132} & {\bf 0.7134} & {\bf 0.7133} \\
            & & & 0.8861 & {\bf 0.9280} & {\bf 0.9305} & {\bf 0.9308} & {\bf 0.9307} & {\bf 0.9306} \\
         \cline{3-9}
         & & \multirow{4}{*}{New2} &
            417.16 & 149.21 & 143.56 & 144.00 & 144.33 & 144.43 \\
            & & & 21.95 & 26.42 & 26.59 & 26.58 & 26.57 & 26.57 \\
            & & & 0.4808 & 0.6636 & 0.6741 & 0.6736 & 0.673 & 0.6729 \\
            & & & {\bf 0.8871} & 0.9144 & 0.9121 & 0.9112 & 0.9106 & 0.9107 \\
         \cline{3-9}
         & & \multirow{4}{*}{Integer Haar} &
            422.81 & 139.29 & 131.03 & 130.01 & 129.68 & 129.72 \\
            & & & 21.88 & 26.72 & 26.98 & 27.02 & 27.03 & 27.03 \\
            & & & 0.4923 & 0.6754 & 0.6887 & 0.6903 & 0.6907 & 0.6908 \\
            & & & 0.8197 & 0.9178 & 0.9189 & 0.9201 & 0.9199 & 0.9199 \\
         \cline{2-9} \\
         \cline{2-9}
        & \multirow{28}{*}{32:1} &
         \multirow{4}{*}{GHM} & 
            9930.8 & 3519.9 & 283.21 & 248.98 & 246.14 & 245.9 \\
            & & & 8.178 & 12.69 & 23.64 & 24.20 & 24.25 & 24.25 \\
            & & & 0.0484 & 0.1632 & 0.5327 & 0.5633 & 0.5664 & 0.5667 \\
            & & & 0.5877 & 0.5275 & 0.8622 & 0.8701 & 0.871 & 0.8711 \\
         \cline{3-9}
         & &  \multirow{4}{*}{SA4} &
            {\bf 3805.5} & {\bf 236.61} & 201.48 & 199.9 & 199.67 & 199.86 \\
            & & & {\bf 12.35} & {\bf 24.42} & 25.12 & 25.15 & 25.16 & 25.15 \\
            & & & {\bf 0.2487} & {\bf 0.5628} & {\bf 0.6022} & 0.6038 & 0.6043 & 0.6042 \\
            & & & 0.6103 & 0.8725 & {\bf 0.8811} & 0.8805 & 0.8804 & 0.8803 \\
          \cline{3-9}
         & &  \multirow{4}{*}{CL} &
            3939.5 & 244.59 & 201.47 & 197.79 & 197.64 & 197.75 \\
            & & & 12.20 & 24.27 & 25.11 & 25.20 & 25.20 & 25.20 \\
            & & & 0.24 & 0.5619 & 0.6009 & 0.6052 & 0.6054 & 0.6053 \\
            & & & {\bf 0.6241} & 0.8595 & 0.8706 & 0.8716 & 0.8713 & 0.8713 \\
          \cline{3-9}
         & & \multirow{4}{*}{Alpert} &
            4044.6 & 236.81 & 197.71 & {\bf 192.21} & {\bf 191.70} & {\bf 191.90} \\
            & & & 12.08 & {\bf 24.41} & 25.20 & {\bf 25.32} & {\bf 25.33} & {\bf 25.33} \\
            & & & 0.2242 & 0.5604 & 0.5999 & 0.6061 & 0.6069 & 0.6067 \\
            & & & 0.6053 & 0.8754 & 0.8799 & 0.8817 & 0.8817 & 0.8816 \\
         \cline{3-9}
         & & \multirow{4}{*}{New1} &
            4044.6 & 236.46 & {\bf 197.54} & 192.33 & 191.93 & 192.16 \\
            & & & 12.08 & 24.42 & {\bf 25.20} & {\bf 25.32} & {\bf 25.33} & 25.32 \\
            & & & 0.2242 & 0.5611 & 0.6008 & {\bf 0.6067} & {\bf 0.6075} & {\bf 0.6073} \\
            & & & 0.6053 & {\bf 0.8760} & 0.8805 & {\bf 0.8820} & {\bf 0.8819} & {\bf 0.8818} \\
         \cline{3-9}
         & & \multirow{4}{*}{New2} &
            4044.6 & 273.36 & 238.9 & 238.69 & 239.17 & 239.67 \\
            & & & 12.08 & 23.79 & 24.38 & 24.39 & 24.38 & 24.37 \\
            & & & 0.2242 & 0.5342 & 0.5636 & 0.5634 & 0.563 & 0.5628 \\
            & & & 0.6053 & 0.8639 & 0.8544 & 0.8527 & 0.8520 & 0.8522 \\
         \cline{3-9}
         & & \multirow{4}{*}{Integer Haar} &
            4029.1 & 264.94 & 222.82 & 218.29 & 216.56 & 216.82 \\
            & & & 12.10 & 23.93 & 24.68 & 24.77 & 24.8 & 24.8 \\
            & & & 0.2329 & 0.5406 & 0.5734 & 0.5753 & 0.5771 & 0.5768 \\
            & & & 0.6029 & 0.8535 & 0.8621 & 0.8637 & 0.8631 & 0.863 \\
         \hline
      \end{tabular}}
\end{table}

Table \ref{tab:10} shows results for the color image {\tt Texture 1}.
For CR = 2 for decomposition levels bigger than 1, the New1 multifilter
has max PSNR, SSIM and MS-SSIM values.  For CR = 4, the SA4 multifilter
has max PSNR and SSIM values for all decomposition levels, which are
slightly higher (0.03dB and 0.020 respectively) than New1, while New1
has max MS-SSIM slightly higher (0.20dB, 0.020dB and 0.03dB,
respectively) than New2.  For CR = 8, SA4 has max PSNR and SSIM values
for all decomposition level, which are slightly higher (0.03dB, 0.04dB
and 0.012dB, respectively) than New1.

% Table 10
\begin{table}
\centering 
  \renewcommand\thetable{10}
  \caption{Results for {\tt Texture 1}. The four lines for each multiwavelet represent MSE, PSNR, SSIM and MS-SSIM, in this order.}%
     {\label{tab:10}
       \footnotesize
       \setlength\tabcolsep{4pt}
       \renewcommand{\arraystretch}{1.0}
       \begin{tabular}{|c|c|c|c|c|c|c|c|c|}
         \hline
         \multirow{2}{*}{Image} &
         \multirow{2}{*}{CR} &  Multi- &                       
         \multicolumn{6}{c|}{Decomposition level} \\
         \cline{4-9}
         & & wavelet & 1 & 2 & 3 & 4 & 5 & 6\\
         \hline
         \multirow{57}{*}{\parbox{1.5in}{\includegraphics[width=1.5in]{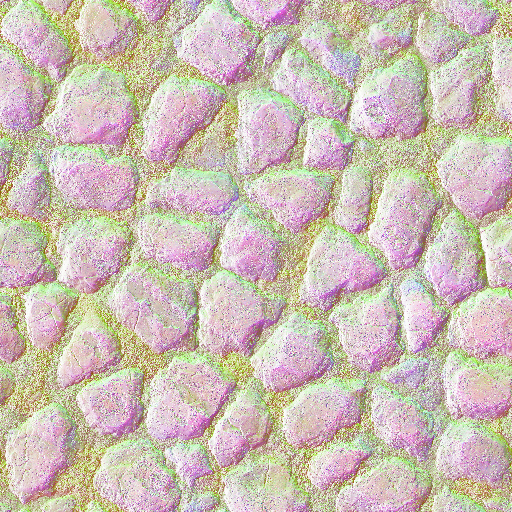}
         \\\centering {\tt Texture 1} }} & \multirow{28}{*}{2:1} &
         \multirow{4}{*}{GHM} & 
         79.713 & 53.973 & 51.129 & 50.784 & 50.761 & 50.754 \\
         & & & 29.31 & 30.99 & 31.22 & 31.25 & 31.25 & 31.25 \\
         & & & 0.9875 & 0.9898 & 0.9900 & 0.9900 & 0.9900 & 0.9900 \\
         & & & 0.9919 & 0.9939 & 0.9941 & 0.9941 & 0.9941 & 0.9941 \\
         \cline{3-9}
         & &  \multirow{4}{*}{SA4} &
         46.476 & 43.135 & 42.608 & 42.554 & 42.556 & 42.554 \\
         & & & 31.63 & 31.95 & 32.00 & 32.01 & 32.01 & 32.01 \\
         & & & 0.9854 & 0.9847 & {\bf 0.9848} & {\bf 0.9848} & {\bf 0.9848} & {\bf 0.9848} \\
         & & & 0.9922 & 0.9924 & 0.9924 & 0.9924 & 0.9924 & 0.9924 \\
          \cline{3-9}
         & &  \multirow{4}{*}{CL} &
         46.341 & 43.049	& 42.432 &	42.344	& 42.334 &	42.33 \\
         & & & 31.65& {\bf 31.97} & {\bf 32.03 } & {\bf 32.04}	& {\bf 32.04}	& {\bf 32.04}\\
         & & & 0.9788 & 0.98	& 0.98 & 0.98 &	0.98 &	0.98 \\
         & & & 0.9905 & 0.9909 &	0.9908 & 0.9908	& 0.9908 & 0.9908\\
          \cline{3-9}
         & & \multirow{4}{*}{Alpert} &
         213.6  & 274.26 & 312.27	& 361.87 & 437.23 & 485.49 \\
         & & & 24.97  & 23.88 & 23.31 & 22.64 & 21.79 & 21.31 \\
         & & & 0.9768  & 0.8952	& 0.8301 & 0.7752 & 0.7414 & 0.7329 \\
         & & & 0.9829  & 0.9545	& 0.9156 & 0.8599 & 0.7942 & 0.7681 \\
         \cline{3-9}
         & & \multirow{4}{*}{New1} &
		 46.024	&  {\bf 42.953}  & {\bf 42.32} & {\bf 42.242}&	{\bf 42.223} & {\bf 42.227}\\
		 & & & 31.67 & {\bf 31.97} & {\bf 32.03} &	{\bf 32.04} & {\bf 32.04} &	{\bf 32.04}\\
		 & & & {\bf 0.9853} & {\bf 0.9848} & {\bf 0.9848} & 0.9847 & 0.9847 & 0.9847\\
 & & & {\bf 0.9938} & {\bf 0.9928} & {\bf 0.9926} & {\bf 0.9926} & {\bf 0.9926} &{\bf 0.9926}\\
         \cline{3-9}
         & & \multirow{4}{*}{New2} &
         46.177	& 44.069 & 44.158 &	44.243 & 44.24 & 44.238\\
		& & & 31.66 & 31.85 & 31.84 &	31.83 &	31.83 &	31.83\\
		& & & {\bf 0.9854} & 0.9843 &	0.9844 & 0.9845 & 0.9845 & 0.9845\\
		& & & 0.9922 & 0.9920 &	0.9920 & 0.9920	& 0.9920 & 0.9920 \\
         \cline{3-9}
         & & \multirow{4}{*}{Integer Haar} &
         {\bf 45.865}	& 45.222 & 45.112 & 45.123 & 45.112 & 45.108\\
		 & & & {\bf 31.69} &  31.74 & 31.75 & 31.75 & 31.75 & 31.75\\
		 & & & 0.9844 & 0.9825 & 0.9825	& 0.9825 & 0.9825 & 0.9825\\
		 & & & 0.9936 & 0.9918 & 0.9916	& 0.9916 & 0.9916 & 0.9916\\
         \cline{2-9} \\

         \cline{2-9}
        & \multirow{28}{*}{4:1} &
         \multirow{4}{*}{GHM} & 
          432.58 & 230.44 & 212.9 & 210.92 & 210.76 & 210.73\\
		 & & & 21.94 & 24.66 & 25 & 25.04 & 25.05 & 25.05\\
		 & & & 0.9051 & 0.9466 & 0.9488	& 0.9493 & 0.9493 & 0.9493\\
		 & & & 0.9442 & 0.9702 & 0.9717 & 0.9718 & 0.9718 & 0.9718\\
         \cline{3-9}
         & &  \multirow{4}{*}{SA4} &
          {\bf 204.65} & {\bf 184.38} & {\bf 181.18}& {\bf 180.82} &{\bf  180.81} &{\bf 180.78}\\
			& & & {\bf 25.17} & {\bf 25.62} & {\bf 25.7} & {\bf 25.71}	& {\bf 25.71} & {\bf 25.71}\\
			& & & {\bf 0.9301} & {\bf 0.9288} & {\bf 0.9296} & {\bf 0.9296} & {\bf 0.9296} & {\bf 0.9296}\\
			& & & 0.9626 & 0.9648 & {\bf 0.9650} & {\bf 0.9649} & {\bf 0.9649}	&{\bf 0.9649}\\
          \cline{3-9}
         & &  \multirow{4}{*}{CL} &
         206.16	 & 185.71 & 182.11 & 181.55 & 181.53 & 181.52\\
		& & & 25.14 & 25.59	& 25.68 & 25.69 & 25.69 & 25.69\\
		& & & 0.9049 & 0.9114 & 0.9120 & 0.9122	& 0.9122 & 0.9122\\
		& & & 0.9568 & 0.9586 & 0.9586 & 0.9586 & 0.9586 & 0.9586\\
          \cline{3-9}
         & & \multirow{4}{*}{Alpert} &
        326.72 & 366.72	& 401.02 & 450.07 & 525.38 & 573.66\\
		& & & 23.13 & 22.62 & 22.22 & 21.70 & 21.00 & 20.60\\
		& & & 0.9435 & 0.8529 & 0.7824 & 0.725 & 0.6897 & 0.6807\\
		& & & 0.9658 & 0.9328 &	0.8902 & 0.833 & 0.7671 & 0.7410\\
         \cline{3-9}
         & & \multirow{4}{*}{New1} &
         205.68	& 185.71 & 182.16 &	181.61 & 181.53 & 181.53\\
	  & & & 25.15	& 25.59 & 25.67	& 25.69 & 25.69 &	25.69\\
	  & & & 0.9299 & 0.9272 & 0.9277 & 0.9276 & 0.9277 & 0.9277\\
      & & & {\bf 0.9705} & {\bf 0.9657} & {\bf 0.9649} & {\bf 0.9648} & {\bf 0.9648} & {\bf 0.9648}\\
         \cline{3-9}
         & & \multirow{4}{*}{New2} &
            206.37 & 190.66	& 190.61 & 191.07 & 191.08 & 191.07\\
	  & & & 25.14 & 25.47 & 25.47 & 25.46 & 25.46 & 25.46\\
	  & & & 0.9296 & 0.9243	& 0.9246 & 0.9247 & 0.9248 & 0.9248\\
	  & & & 0.9622 & 0.9617	& 0.9612 & 0.9613 & 0.9613 & 0.9613\\
         \cline{3-9}
         & & \multirow{4}{*}{Integer Haar} &
          207.13& 189.59& 187.42& 187.32 & 187.29 & 187.27\\
	  & & & 25.12& 25.50 & 25.55 & 25.55 & 25.55 & 25.55\\
	  & & & 0.9245& 0.9201&	0.9203 & 0.9205 & 0.9205 & 0.9205\\
	  & & & 0.9690 & 0.9625 & 0.9614 & 0.9614 & 0.9614 & 0.9614\\
         \hline
      \end{tabular}}
\end{table}

Table \ref{tab:11} shows results for the color image {\tt Texture
  4}. Since the image allows superb values at higher compression levels,
the beginning value is CR = 64.

For CR=64, the max PSNR value at the first decomposition level is
achieved by the SA4 multifilter, slightly higher (by 0.04dB) than
New1. At the second decomposition level, New1 is better. For
decompositin levels 3 to 7, Integer Haar has the max PSNR levels,
significantly higher (by 0.64dB -- 1.36dB) than New1.
        
For SSIM and MS-SSIM, the maximal values at the first 3 decomposition
levels are for New1 and New2, and at levels 4 to 7 the Integer Haar
multifilter has negligibly higher levels (by 0.003dB).
        
For CR = 128, max PSNR at the first decomposition level is for SA4,
higher by 0.02dB than New1. The New1 multifilter has superior values for
PSNR, SSIM and MS-SSIM at levels 2 to 7 (6.5dB better than CL and SA4,
12dB better than GHM and 6.5dB better than New2).
        
For CR = 256, SA4 has the maximal PSNR values at the first and second
decomposition levels, better by 0.03dB and 0.05dB respectively) than
New1. New1 has max SSIM and MS-SSIM values. From levels 3 to 7 all
maximal values are for New1, bigger by around 5.0dB than SA4 and CL, and
around 8.0dB better than GHM.
 	
For CR = 512, SA4 achieves the maximal PSNR values at the first and
second decomposition levels, negligibly higher (by 0.03dB and 0.04dB,
respectively) than New1. At the second level, New1 have max SSIM and
MS-SSIM values. From levels 3 to 7 New1 has all maximal values, bigger
by around 5.0dB than SA4 and CL, and around 6.0dB better than GHM.

The multifilter New1 applied to the image {\tt Texture 4}, which has a
strongly combed histogram, achieves better results for all metrics, with
minimal differences for all compression levels.

% Table 11
\begin{table}
\centering 
  \renewcommand\thetable{11}
  \caption{Results for {\tt Texture 4}. The four lines for each multiwavelet represent MSE, PSNR, SSIM and MS-SSIM, in this order.}%
     {\label{tab:11}
       \footnotesize
       \setlength\tabcolsep{4pt}
       \renewcommand{\arraystretch}{1.0}
       \begin{tabular}{|c|c|c|c|c|c|c|c|c|c|}
         \hline
         \multirow{2}{*}{Image} &
         \multirow{2}{*}{CR} &  Multi- &                       
         \multicolumn{7}{c|}{Decomposition level} \\
         \cline{4-10}
         & & wavelet & 1 & 2 & 3 & 4 & 5 & 6 & 7\\
         \hline
         \multirow{57}{*}{\parbox{1.5in}{\includegraphics[width=1.5in]{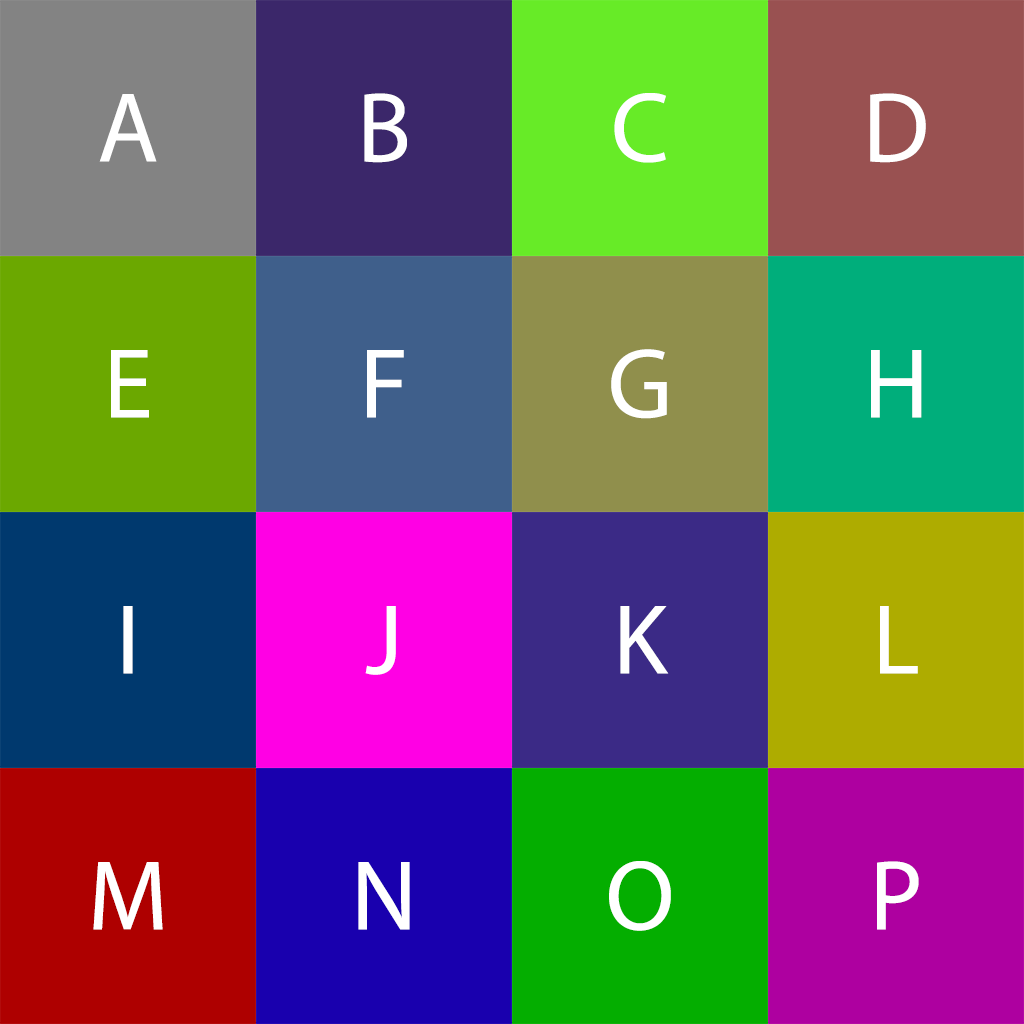}
         \\\centering {\tt Texture 4} }} & \multirow{28}{*}{64:1} &
         \multirow{4}{*}{GHM} & 
          1.10E+04 & 4985.0 & 58.32 & 15.81 & 12.58 & 12.00 & 12.00\\
        & & & 7.54 & 11.16 & 30.52 & 36.19 & 37.20 & 37.41 & 37.41\\
        & & & 0.2919 & 0.3080 & 0.9815 & 0.9976 & 0.9957 & 0.9953 & 0.9952\\
        & & & 0.7652 & 0.6724 & 0.9716 & 0.9921 & 0.9926 & 0.9928 & 0.9928\\
         \cline{3-10}
         & &  \multirow{4}{*}{SA4} &
         {\bf 5995.0} & 28.03 & 4.139 & 3.3107 & 3.5373 & 3.77 & 3.745\\
        & & & {\bf 10.36} & 33.72 & 42.01 & 42.99 & 42.70 & 42.42 & 42.45\\
        & & & 0.4384 & 0.9844 & 0.9983 & 0.9963 & 0.9945 & 0.9945 & 0.9944\\
        & & & 0.8289 & 0.9828 & 0.9962 & 0.9962 & 0.9954 & 0.9951 & 0.9951\\
          \cline{3-10}
         & &  \multirow{4}{*}{CL} &
		  6035 & 37.53 & 1.913 & 1.191 & 1.193 & 1.222 & 1.251\\
        & & & 10.33 & 32.48 & 45.40 & 47.50 & 47.49 & 47.39 & 47.29\\
        & & & 0.4443 & 0.9838 & 0.9995 & 0.9995 & 0.9986 & 0.9985 & 0.9985\\
        & & & 0.8326 & 0.9789 & 0.9989 & 0.9991 & 0.9989 & 0.9988 & 0.9988\\
          \cline{3-10}
         & & \multirow{4}{*}{Alpert} &
		 6039 & 19.85 & 37.37 & 86.73 & 119.8 & 141 & 179.2\\
        & & & 10.32 & 35.16 & 32.41 & 28.75 & 27.35 & 26.64 & 25.60\\
        & & & 0.4362 & 0.9875 & 0.9937 & 0.9758 & 0.9497 & 0.9135 & 0.8859\\
        & & & 0.8236 & 0.9975 & 0.9936 & 0.9813 & 0.9672 & 0.9625 & 0.9658\\
        \cline{3-10}
         & & \multirow{4}{*}{New1} &
		 6039 & {\bf 7.116} & 0.5232 & 0.4045 & 0.4921 & 0.4427 & 0.4368\\
        & & & 10.32 & {\bf 39.72} & 50.95 & 52.07 & 51.22 & 51.68 & 51.74\\
        & & & 0.4362 & {\bf 0.9882} & {\bf 0.9998} & 0.9998 & 0.9998 & 0.9998 & 0.9998\\
        & & & 0.8236 & {\bf 0.9983} & {\bf 0.9996} & 0.9996 & 0.9995 & 0.9996 & 0.9996\\
         \cline{3-10}
         & & \multirow{4}{*}{New2} &
          6064 & 40.49 & 1.642 & 0.9366 & 0.9678 & 1.135 & 1.108\\
        & & & 10.31 & 32.14 & 46.09 & 48.52 & 48.39 & 47.70 & 47.80\\
        & & & {\bf 0.4421} & 0.9830 & 0.9996 & 0.9997 & 0.9996 & 0.9995	& 0.9995\\
        & & & {\bf 0.8311} & 0.9790 & 0.9992 & 0.9995 & 0.9995 & 0.9994 & 0.9994\\
         \cline{3-10}
         & & \multirow{4}{*}{Integer Haar} &
		 6039 & 8.795 & 0.4513 & 0.3109 & 0.2893 & 0.3237 & 0.3197\\
        & & & 10.32 & 38.80 & 51.59 & 53.21 & 53.54 & 53.05 & 53.10\\
        & & & 0.4362 & 0.9880 & {\bf 0.9998} & {\bf 0.9999} & {\bf 0.9999} & {\bf 0.9999} &{\bf  0.9999}\\
        & & & 0.8236 & 0.9981 & {\bf 0.9996} & {\bf 0.9998} &{\bf  0.9998} &{\bf  0.9997}& {\bf 0.9997}\\
         \cline{2-10} \\
         \cline{2-10}
        & \multirow{28}{*}{128:1} &
         \multirow{4}{*}{GHM} & 
        1.30E+04 & 8189 & 1352 & 57.81 & 44.90 & 42.65 & 42.63\\
        & & & 6.99 & 9.00 & 16.90 & 30.53 & 31.63 & 31.85 & 31.85\\
        & & & 0.2381 & 0.2386 & 0.5278 & 0.9768 & 0.9841 & 0.9831 & 0.9826\\
        & & & 0.7586 & 0.6887 & 0.7073 & 0.9683 & 0.9734 & 0.9742 & 0.9741\\
         \cline{3-10}
         & &  \multirow{4}{*}{SA4} &
         {\bf 9496} & 1607 & 29.78 & 20.07 & 20.75 & 22.44 & 22.18\\
        & & & {\bf 8.36} & 16.18 & 33.41 & 35.13 & 34.99 & 34.64 & 34.69\\
        & & & 0.3120 & 0.6811 & 0.9748 & 0.9777 & 0.9757 & 0.9749 & 0.9741\\
        & & & 0.7853 & 0.8825 & 0.9739 & 0.9779 & 0.9757 & 0.9740 & 0.9742\\
          \cline{3-10}
         & &  \multirow{4}{*}{CL} &
         9540 & 1710 & 28.20 & 16.94 & 15.61 & 15.62 & 15.91\\
        & & & 8.34 & 15.88 & 33.66 & 35.87 & 36.23 & 36.22 & 36.14\\
        & & & {\bf 0.3168} & 0.6488 & 0.9839 & 0.9934 & 0.9899 & 0.9865 & 0.9864\\
        & & & {\bf 0.7950} & 0.8553 & 0.9820 & 0.9871 & 0.9860 & 0.9856 & 0.9854\\
          \cline{3-10}
         & & \multirow{4}{*}{Alpert} &
        9551 & {\bf 1554} & 41.02 & 88.57 & 121.6 & 143.1 & 181.2\\
        & & & 8.34 & {\bf 16.32 }& 32.01 & 28.66 & 27.29 & 26.58 & 25.55\\
        & & & 0.3114 & {\bf 0.6895} & 0.9930 & 0.9751 & 0.9490 & 0.9125 & 0.8850\\
        & & & 0.7847 & 0.9100 & 0.9925 & 0.9805 & 0.9664 & 0.9614 & 0.9648\\
	     \cline{3-10}
         & & \multirow{4}{*}{New1} &
         9551 & {\bf 1553} & {\bf 5.55} & {\bf 2.956} & {\bf 2.869} & {\bf 3.475} & {\bf 3.353}\\
        & & & 8.34 & {\bf 16.32} & {\bf 40.69} & {\bf 43.43} & {\bf 43.56} & {\bf 42.73} & {\bf 42.89}\\
        & & & 0.3114 & {\bf 0.6895} & {\bf 0.9987} & {\bf 0.9987} & {\bf 0.9985} & {\bf 0.9979} & {\bf 0.9979}\\
        & & & 0.7847 & {\bf 0.9102} & {\bf 0.9979} & {\bf 0.9981} & {\bf 0.9980} & {\bf 0.9974} & {\bf 0.9974}\\
         \cline{3-10}
         & & \multirow{4}{*}{New2} &
         9575 & 1707 & 37.31 & 20.02 & 20.08 & 23.18 & 	22.46\\
        & & & 8.33 & 15.91 & 32.45 & 35.18 & 35.17 & 34.54 & 34.68\\
        & & & 0.3130 & 0.6827 & 0.9795 & 0.9903 & 0.9869 & 0.9843 & 0.9826\\
        & & & 0.7922 & 0.8808 & 0.9774 & 0.9854 & 0.9840 & 0.9813 & 0.9816\\
         \cline{3-10}
         & & \multirow{4}{*}{Integer Haar} &
         9551 & 1555 & 6.976 & 3.422 & 3.308 & 3.863 & 3.732\\
        & & & 8.34 & 16.31 & 39.70 & 42.80 & 42.95 & 42.28 & 42.43\\
        & & & 0.3114 & 0.6893 & 0.9984 & 0.9983 & 0.9982 & 0.9969 & 0.9969\\
        & & & 0.7847 & 0.9096 & 0.9975 & 0.9979 & 0.9978 & 0.9971 & 0.9971\\
       \hline
      \end{tabular}}
\end{table}
% Table 11(continued)
\begin{table}
\centering 
  \renewcommand\thetable{11(continued)}
  \caption{Results for {\tt Texture 4}. The four lines for each multiwavelet represent MSE, PSNR, SSIM and MS-SSIM, in this order.}%
     {
       \footnotesize
       \setlength\tabcolsep{4pt}
       \renewcommand{\arraystretch}{1.0}
       \begin{tabular}{|c|c|c|c|c|c|c|c|c|c|}
         \hline
         \multirow{2}{*}{Image} &
         \multirow{2}{*}{CR} &  Multi- &                       
         \multicolumn{7}{c|}{Decomposition level} \\
         \cline{4-10}
         & & wavelet & 1 & 2 & 3 & 4 & 5 & 6 & 7\\
         \hline
         \multirow{57}{*}{\parbox{1.5in}{\includegraphics[width=1.5in]{Flipbook_test}
         \\\centering {\tt Texture 4} }} & \multirow{28}{*}{256:1} &
         \multirow{4}{*}{GHM} & 
          1.41E+04 & 1.10E+04 & 4983 & 188.4 & 94.64 & 86.37 & 86.37\\
        & & & 6.66 & 7.71 & 11.16 & 25.41 & 28.38 & 28.78 & 28.78\\
        & & & 0.1936 & 0.2070 & 0.2806 & 0.8938 & 0.9424 & 0.9505 & 0.9468\\
        & & & 0.6961 & 0.7198 & 0.5966 & 0.8823 & 0.9365 & 0.9428 & 0.9425\\
         \cline{3-10}
         & &  \multirow{4}{*}{SA4} &
         {\bf 1.18E+04} & {\bf 6094} & 129.1 & 62.03 & 64.08 & 72.09 & 70.10\\
        & & & {\bf 7.42} & {\bf 10.29} & 27.06 & 30.22 & 30.08 & 29.57 & 29.69\\
        & & & {\bf 0.2406} & 0.4089 & 0.9504 & 0.9417 & 0.9329 & 0.9268 & 0.9241\\
        & & & {\bf 0.7621} & 0.7943 & 0.9373 & 0.9405 & 0.9323 & 0.9253 & 0.9266\\
          \cline{3-10}
         & &  \multirow{4}{*}{CL} &
		  1.18E+04 & 6175 & 155.1 & 61.75 & 52.84 & 52.23 & 53.44\\
        & & & 7.41 & 10.23 & 26.26 & 30.24 & 30.92 & 30.97 & 30.87\\
        & & & 0.2422 & 0.4048 & 0.9314 & 0.9692 & 0.9685 & 0.9637 & 0.9617\\
        & & & 0.7600 & 0.7877 & 0.9183 & 0.9614 & 0.9617 & 0.9616 & 0.9606\\
          \cline{3-10}
         & & \multirow{4}{*}{Alpert} &
		 1.19E+04 & 6164 & 78.59 & 98.25 & 130.5 & 153.9 & 191.1\\
        & & & 7.39 & 10.24 & 29.18 & 28.21 & 26.98 & 26.26 & 25.32\\
        & & & 0.2370 & 0.4230 & 0.9756 & 0.9721 & 0.9458 & 0.9085 & 0.8807\\
        & & & 0.7431 & 0.8101 & 0.9858 & 0.9776 & 0.9632 & 0.9568 & 0.9604\\
        \cline{3-10}
         & & \multirow{4}{*}{New1} &
		1.19E+04 & 6164 & {\bf 53.05} & {\bf 15.89} & {\bf 15.05} &{\bf 18.74} & {\bf 17.72}\\
        & & & 7.39 & 10.24 & {\bf 30.90} & {\bf 36.12} & {\bf 36.36} & {\bf 35.41} & {\bf 35.65}\\
        & & & 0.2370 & {\bf 0.4230} & {\bf 0.9788} & {\bf 0.9929} & {\bf 0.9902} &{\bf  0.9839} & {\bf 0.9839}\\
        & & & 0.7431 & {\bf 0.8101} & {\bf 0.9883} & {\bf 0.9926} & {\bf 0.9909} & {\bf 0.9864} & {\bf 0.9870}\\
         \cline{3-10}
         & & \multirow{4}{*}{New2} &
          1.19E+04 & 6223 & 208.1 & 81.83 & 79.52 & 90.38 & 87.29\\
        & & & 7.39 & 10.19 & 24.98 & 29.02 & 29.14 & 28.59 & 28.74\\
        & & & 0.2414 & 0.4131 & 0.9390 & 0.9520 & 0.9408 & 0.9184 & 0.9202\\
        & & & 0.7546 & 0.7978 & 0.9253 & 0.9537 & 0.9466 & 0.9329 & 0.9362\\
         \cline{3-10}
         & & \multirow{4}{*}{Integer Haar} &
		 1.19E+04 & 6164 & 67.16 & 19.00 & 17.91 & 20.98 & 19.78\\
        & & & 7.39 & 10.24 & 29.87 & 35.35 & 35.60 & 34.92 & 35.17\\
        & & & 0.2370 & 0.4230 & 0.9767 & 0.9916 & 0.9892 & 0.9808 & 0.9810\\
        & & & 0.7431 & 0.8101 & 0.9863 & 0.9919 & 0.9905 & 0.9855 & 0.9862\\
         \cline{2-10} \\
         \cline{2-10}
       & \multirow{28}{*}{512:1} &
         \multirow{4}{*}{GHM} & 
          1.48E+04 & 1.27E+04 & 8213 & 1631 & 199.0 & 159.7 & 158.3\\
       & & & 6.44 & 7.11 & 8.99 & 16.06 & 25.15 & 26.11 & 26.15\\
       & & & 0.1744 & 0.1956 & 0.2181 & 0.4446 & 0.8571 & 0.8906 & 0.8844\\
       & & & 0.6838 & 0.7316 & 0.6184 & 0.6104 & 0.8629 & 0.8875 & 0.8874\\
         \cline{3-10}
         & &  \multirow{4}{*}{SA4} &
         {\bf 1.36E+04} & {\bf 9585} & 1751 & 130.7 & 134.2 & 155.9 & 149.6\\
       & & & {\bf 6.81} & {\bf 8.32} & 15.78 & 26.98 & 26.87 & 26.21 & 26.39\\
       & & & {\bf 0.2092} & 0.2908 & 0.6348 & 0.9128 & 0.8702 & 0.8515 & 0.8526\\
       & & & {\bf 0.7466} & 0.7514 & 0.8420 & 0.9097 & 0.8847 & 0.8666 & 0.8703\\
          \cline{3-10}
         & &  \multirow{4}{*}{CL} &
         1.36E+04 & 9684 & 2083 & 144.5 & 108 & 105.2 & 108.2\\
       & & & 6.79 & 8.28 & 15.00 & 26.55 & 27.81 & 27.92 & 27.80\\
       & & & 0.2094 & 0.2820 & 0.5581 & 0.9317 & 0.9426 & 0.9382 & 0.9333\\
       & & & 0.7419 & 0.7520 & 0.7533 & 0.9184 & 0.9365 & 0.9373 & 0.9352\\
          \cline{3-10}
         & & \multirow{4}{*}{Alpert} &
        1.37E+04 & 9680 & 1747 & 124.6 & 152.5 & 179.8 & 214.2\\
       & & & 6.78 & 8.28 & 15.79 & 27.18 & 26.3 & 25.59 & 24.83\\
       & & & 0.2026 & 0.2956 & 0.6476 & 0.9558 & 0.9376 & 0.8982 & 0.8714\\
       & & & 0.7207 & 0.7580 & 0.8690 & 0.9721 & 0.9559 & 0.9460 & 0.9511\\
	     \cline{3-10}
         & & \multirow{4}{*}{New1} &
         1.37E+04 & 9680 & {\bf 1745} & {\bf 49.30} & {\bf 44.09} & {\bf 56.05} & {\bf 51.34}\\
       & & & 6.78 & 8.28 & {\bf 15.80} & {\bf 31.21} & {\bf 31.69} & {\bf 30.65} & {\bf 31.03}\\
       & & & 0.2026 & {\bf 0.2956} & {\bf 0.6485} & {\bf 0.9734} & {\bf 0.9696} & {\bf 0.9436} & {\bf 0.9448}\\
       & & & 0.7207 & {\bf 0.7580} & {\bf 0.8704} & {\bf 0.9841} & {\bf 0.9755} & {\bf 0.9588} & {\bf 0.9614}\\
         \cline{3-10}
         & & \multirow{4}{*}{New2} &
         1.37E+04 & 9728 & 1905 & 193.4 & 178.1 & 199.3 & 193.2\\
       & & & 6.78 & 8.26 & 15.41 & 25.28 & 25.64 & 25.15 & 25.28\\
       & & & 0.2056 & 0.2953 & 0.6384 & 0.9318 & 0.8936 & 0.8371 & 0.8451\\
       & & & 0.7302 & 0.7685 & 0.8447 & 0.9242 & 0.9051 & 0.8780 & 0.8819\\
         \cline{3-10}
         & & \multirow{4}{*}{Integer Haar} &
         1.37E+04 & 9680 & 1750 & 60.02 & 53.95 & 62.12 & 56.73\\
       & & & 6.78 & 8.28 & 15.78 & 30.35 & 30.82 & 30.21 & 30.60\\
       & & & 0.2026 & 0.2956 & 0.6470 & 0.9713 & 0.9662 & 0.9386 & 0.9414\\
       & & & 0.7207 & 0.7580 & 0.8678 & 0.9827 & 0.9730 & 0.9589 & 0.9619\\
       \hline
      \end{tabular}}
\end{table}
  
Table \ref{tab:12} shows results for the grayscale image {\tt Brick
  wall}. For CR=16, at the first decomposition level SA4 has max PSNR
values, while at levels 3 to 7 CL is best, negligibly higher (by 0.01dB)
than New1. The max SSIM is for the GHM multifilter, while the SA4
multifilter has max MS-SSIM, negligible higher (by 0.004dB) than New1.
For CR=32 and 64, SA4 has the max PSNR, SSIM and MS-SSIM values at
decomposition levels 1 and 2, whuile for levels 3 to 7 the SA4
multifilter New1 has all max values.

% Table 12
\begin{table}
\centering 
  \renewcommand\thetable{12}
  \caption{Results for {\tt Brick wall}. The four lines for each multiwavelet represent MSE, PSNR, SSIM and MS-SSIM, in this order.}%
     {\label{tab:12}
       \footnotesize
       \setlength\tabcolsep{4pt}
       \renewcommand{\arraystretch}{1.0}
       \begin{tabular}{|c|c|c|c|c|c|c|c|c|c|}
         \hline
         \multirow{2}{*}{Image} &
         \multirow{2}{*}{CR} &  Multi- &                       
         \multicolumn{7}{c|}{Decomposition level} \\
         \cline{4-10}
         & & wavelet & 1 & 2 & 3 & 4 & 5 & 6 & 7\\
         \hline
         \multirow{57}{*}{\parbox{1.5in}{\includegraphics[width=1.5in]{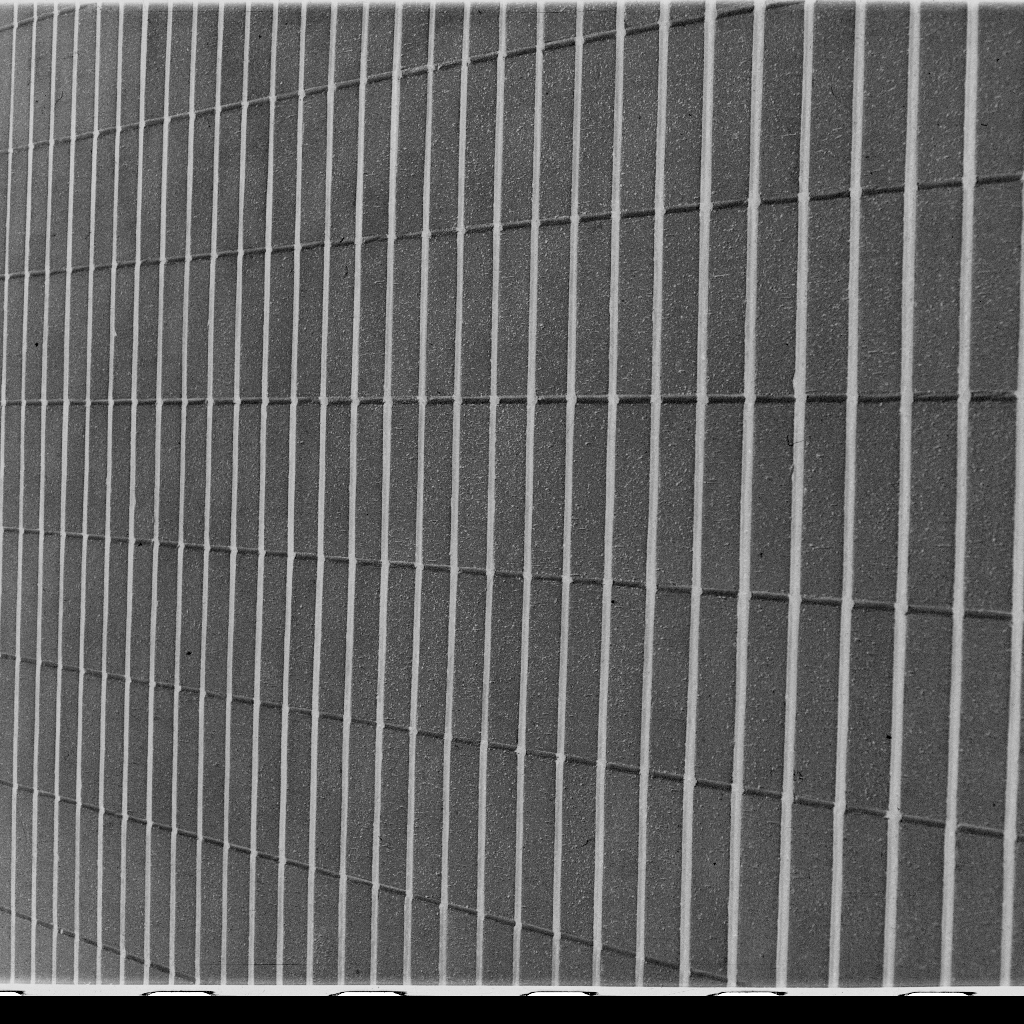}
         \\\centering {\tt Brick wall} }} & \multirow{28}{*}{16:1} &
         \multirow{4}{*}{GHM} & 
          5617 & 278.9 & 141.7 & 135 & 133.8 & 133.6 & 133.6\\
        & & & 10.64 & 23.68 & 26.62 & 26.83 & 26.87 & 26.87 & 26.87\\
        & & & 0.5177 & {\bf 0.9870} & {\bf 0.9860} & {\bf 0.9859} & {\bf 0.9859} & {\bf 0.9859} & {\bf 0.9859}\\
        & & & 0.7018 & 0.9006 & 0.9383 & 0.9414 & 0.9420 & 0.9421 & 0.9421\\
         \cline{3-10}
         & &  \multirow{4}{*}{SA4} &
         {\bf 288.7} & 111.8 & 106.7 & 105.8 & 105.5 & 105.5 & 105.4\\
        & & & {\bf 23.53} & 27.65 & 27.85 & 27.89 & 27.90 & 27.90 & 27.90\\
        & & & {\bf 0.9648} & 0.9854 & 0.9850 & 0.9852 & 0.9851 & 0.9851 & 0.9851\\
        & & & 0.8836 & 0.9463 & {\bf 0.9493} & {\bf 0.9499} & {\bf 0.9501} & {\bf 0.9501} & {\bf 0.9501}\\
          \cline{3-10}
         & &  \multirow{4}{*}{CL} &
		 319.4 & {\bf 109.1} &{\bf 103.4} & {\bf 102.5} & {\bf 102.2} & {\bf 102.2} & {\bf 102.2}\\
        & & & 23.09 & {\bf 27.75} & {\bf 27.99}& {\bf 28.02} & {\bf 28.04}& {\bf 28.04} & {\bf 28.04}\\
        & & & 0.9231 & 0.9856 & 0.9843 & 0.9842 & 0.9841 & 0.9841 & 0.9841\\
        & & & 0.8869 & 0.9432 & 0.9463 & 0.9467 & 0.9469 & 0.9469 & 0.9469\\
          \cline{3-10}
         & & \multirow{4}{*}{Alpert} &
		 298.8 & 167.6 & 252.1 & 366.5 & 459.3 & 495.9 & 511.3\\
        & & & 23.38 & 25.89 & 24.12 & 22.49 & 21.51 & 21.18 & 21.04\\
        & & & 0.9594 & 0.9782 & 0.9285 & 0.8381 & 0.8100 & 0.8058 & 0.8046\\
        & & & 0.9162 & 0.9340 & 0.8850 & 0.8149 & 0.7861 & 0.7817 & 0.7801\\
        \cline{3-10}
         & & \multirow{4}{*}{New1} &
		 296.1 & {\bf 109.1} & 103.5 & 102.7 & 102.5 & 102.5 & 102.5\\
        & & & 23.42 & {\bf 27.75} & 27.98 & 28.01 & 28.02 & 28.03 & 28.03\\
        & & & 0.9593 & 0.9848 & 0.9844 & 0.9846 & 0.9845 & 0.9845 & 0.9845\\
        & & & {\bf 0.9165} & {\bf 0.9467} & 0.9491 & 0.9496 & 0.9497 & 0.9497 & 0.9497\\
         \cline{3-10}
         & & \multirow{4}{*}{New2} &
          332 & 112.2 & 107.1 & 106.8 & 106.8 & 106.8 & 106.8\\
        & & & 22.92 & 27.63 & 27.83 & 27.84 & 27.85 & 27.85 & 27.85\\
        & & & 0.9533 & 0.9835 & 0.9829 & 0.9830 & 0.9830 & 0.9830 & 0.9830\\
        & & & 0.8750 & 0.9431 & 0.9463 & 0.9468 & 0.9469 & 0.9469 & 0.9469\\
         \cline{3-10}
         & & \multirow{4}{*}{Integer Haar} &
	     307.2 & 112.4 & 106.2 & 105.3 & 105.2 & 105.3 & 105.2\\
        & & & 23.26 & 27.62 & 27.87 & 27.91 & 27.91 & 27.91 & 27.91\\
        & & & 0.9581 & 0.9831 & 0.9830 & 0.9832 & 0.9832 & 0.9832 & 0.9832\\
        & & & 0.9138 & 0.9422 & 0.9449 & 0.9454 & 0.9456 & 0.9454 & 0.9456\\
         \cline{2-10} \\
         \cline{2-10}
        & \multirow{28}{*}{32:1} &
         \multirow{4}{*}{GHM} & 
         7665 & 2719 & 232.3 & 200 & 195.4 & 194.5 & 194.4\\
        & & & 9.29 & 13.79 & 24.47 & 25.12 & 25.22 & 25.24 & 25.24\\
        & & & 0.3534 & 0.5730 & 0.9697 & 0.9713 & 0.9716 & 0.9716 & 0.9717\\
        & & & 0.7400 & 0.6433 & 0.8972 & 0.9075 & 0.9092 & 0.9096 & 0.9097\\
         \cline{3-10}
         & &  \multirow{4}{*}{SA4} &
        {\bf 3205} & {\bf 171.8} & 149.9 & 146.6 & 145.8 & 145.7 & 145.7\\
        & & & {\bf 13.07} & {\bf 25.78} & 26.37 & 26.47 & 26.49 & 26.50 & 26.50\\
        & & & {\bf 0.5909} & {\bf 0.9726} & 0.9735 & 0.9742 & 0.9742 & 0.9742 & 0.9742\\
        & & & {\bf 0.6891} & 0.9117 & 0.9214 & 0.9234 & 0.9238 & 0.9239 & 0.9239\\
          \cline{3-10}
         & &  \multirow{4}{*}{CL} &
        3327 & 175.2 & 145.4 & 142.5 & 141.6 & 141.5 & 141.5\\
        & & & 12.91 & 25.7 & 26.5 & 26.59 & 26.62 & 26.62 & 26.62\\
        & & & 0.6012 & 0.9751 & 0.9733 & 0.9733 & 0.9734 & 0.9735 & 0.9735\\
        & & & 0.6929 & 0.9109 & 0.9191 & 0.9204 & 0.9209 & 0.9210 & 0.9210\\
          \cline{3-10}
         & & \multirow{4}{*}{Alpert} &
       3344 & 217.7 & 282.6 & 394.9 & 487 & 523.5 & 538.9\\
        & & & 12.89 & 24.75 & 23.62 & 22.17 & 21.26 & 20.94 & 20.82\\
        & & & 0.5870 & 0.9664 & 0.9193 & 0.8290 & 0.7997 & 0.7953 & 0.7941\\
        & & & 0.6540 & 0.9086 & 0.8656 & 0.7967 & 0.7673 & 0.7629 & 0.7612\\
	     \cline{3-10}
         & & \multirow{4}{*}{New1} &
         3344 & 173 & {\bf 145.3} & {\bf 142.5} & {\bf 141.7} & {\bf 141.6} & {\bf 141.5}\\
        & & & 12.89 & 25.75 & {\bf 26.51} & {\bf 26.59} & {\bf 26.62} & {\bf 26.62} & {\bf 26.62}\\
        & & & 0.587 & 0.9722 & {\bf 0.9735} & {\bf 0.9738} & {\bf 0.9738} & {\bf 0.9738} & {\bf 0.9738}\\
        & & & 0.654 & {\bf 0.9166} & {\bf 0.9215} & {\bf 0.9228} & {\bf 0.9231}& {\bf 0.9231} & {\bf 0.9232}\\
         \cline{3-10}
         & & \multirow{4}{*}{New2} &
         3451 & 191.2 & 155 & 152.1 & 151.7 & 151.7 & 151.6\\
        & & & 12.75 & 25.32 & 26.23 & 26.31 & 26.32 & 26.32 & 26.32\\
        & & & 0.5871 & 0.968 & 0.9699 & 0.9703 & 0.9702 & 0.9703 & 0.9703\\
        & & & 0.6808 & 0.9093 & 0.9163 & 0.9175 & 0.9177 & 0.9177 & 0.9177\\
         \cline{3-10}
         & & \multirow{4}{*}{Integer Haar} &
         3344 & 192.1 & 152.4 & 148.5 & 147.9 & 147.8 & 147.8\\
       & & & 12.89 & 25.3 & 26.3 & 26.41 & 26.43 & 26.43 & 26.43\\
       & & & 0.5871 & 0.9686 & 0.9707 & 0.9711 & 0.9709 & 0.9710 & 0.9710\\
       & & & 0.6539 & 0.9115 & 0.9156 & 0.9170 & 0.9171 & 0.9171 & 0.9171\\
       \hline
      \end{tabular}}
\end{table}
% Table 12(continued)
\begin{table}
\centering 
  \renewcommand\thetable{13(continued)}
  \caption{Results for {\tt Brick wall}. The four lines for each multiwavelet represent MSE, PSNR, SSIM and MS-SSIM, in this order.}%
    {
       \footnotesize
       \setlength\tabcolsep{4pt}
       \renewcommand{\arraystretch}{1.0}
       \begin{tabular}{|c|c|c|c|c|c|c|c|c|c|}
         \hline
         \multirow{2}{*}{Image} &
         \multirow{2}{*}{CR} &
         Multi- &                       
         \multicolumn{7}{c|}{Decomposition level} \\
         \cline{4-10}
         & & wavelet & 1 & 2 & 3 & 4 & 5 & 6 & 7\\
         \hline
         \multirow{27}{*}{\parbox{1.5in}{\includegraphics[width=1.5in]{1.4.09}
         \\\centering {\tt Brick wall} }} & \multirow{28}{*}{64:1} &
         \multirow{4}{*}{GHM} & 
          9198 & 5278 & 576.6 & 308.7 & 288.3 & 284.4 & 283.8\\
       & & & 8.494 & 10.91 & 20.52 & 23.24 & 23.53 & 23.59 & 23.60\\
       & & & 0.1961 & 0.4234 & 0.8255 & 0.9446 & 0.9481 & 0.9493 & 0.9495\\
       & & & 0.6883 & 0.6511 & 0.7417 & 0.8630 & 0.8697 & 0.8715 & 0.8718\\
         \cline{3-10}
         & &  \multirow{4}{*}{SA4} &
         {\bf 5513.0} & {\bf 624.0} & 216.5 & 201.3 & 198 & 197.5 & 197.4\\
       & & & {\bf 10.72} & {\bf 20.18} & 24.78 & 25.09 & 25.16 & 25.18 & 25.18\\
       & & & {\bf 0.4275} & {\bf 0.7833} & 0.9552 & 0.9591 & 0.9595 & 0.9597 & 0.9597\\
       & & & {\bf 0.6928} & 0.7199 & 0.8904 & 0.8968 & 0.8981 & 0.8984 & 0.8984\\
         \cline{3-10}
         & &  \multirow{4}{*}{CL} &
          5624 & 649.3 & 218.5 & 203 & 199.2 & 198.6 & 198.5\\
       & & & 10.63 & 20.01 & 24.74 & 25.06 & 25.14 & 25.15 & 25.15\\
       & & & 0.4359 & 0.7845 & 0.9614 & 0.9608 & 0.9609 & 0.9611 & 0.9611\\
       & & & 0.7105 & 0.7042 & 0.8912 & 0.8948 & 0.8958 & 0.8960 & 0.8961\\
          \cline{3-10}
         & & \multirow{4}{*}{Alpert} &
         5644 & 650.1 & 337.1 & 439.2 & 528.9 & 564.9 & 580.2\\
       & & & 10.62 & 20.00 & 22.85 & 21.70 & 20.90 & 20.61 & 20.50\\
       & & & 0.4280 & 0.7709 & 0.9047 & 0.8160 & 0.7858 & 0.7813 & 0.7801\\
       & & & 0.6696 & 0.7310 & 0.8469 & 0.7807 & 0.7510 & 0.7465 & 0.7449\\
         \cline{3-10}
         & & \multirow{4}{*}{New1} &
       5644 & 641.2 & {\bf 215.7} & {\bf 201.1} & {\bf 197.6} & {\bf 197.0} & {\bf 196.9}\\
       & & & 10.62 & 20.06 & {\bf 24.79} & {\bf 25.10} & {\bf 25.17} & {\bf 25.19} & {\bf 25.19}\\
       & & & 0.4280 & 0.7674 & {\bf 0.9596} & {\bf 0.9608} & {\bf 0.9608} & {\bf 0.9609} & {\bf 0.9609}\\
       & & & 0.6696 & {\bf 0.7299} & {\bf 0.8987} & {\bf 0.9012} & {\bf 0.9017} & {\bf 0.9019} & {\bf 0.9019}\\
         \cline{3-10}
         & & \multirow{4}{*}{New2} &
        5779 & 692 & 279 & 242.1 & 235.6 & 234.9 & 234.7\\
       & & & 10.51 & 19.73 & 23.67 & 24.29 & 24.41 & 24.42 & 24.43\\
       & & & 0.4263 & 0.7541 & 0.9380 & 0.9473 & 0.9487 & 0.9487 & 0.9488\\
       & & & 0.6910 & 0.6892 & 0.8691 & 0.8841 & 0.8867 & 0.8871 & 0.8871\\
        \cline{3-10}
         & & \multirow{4}{*}{Integer Haar} &
        5644 & 647 & 244.6 & 222.2 & 218.3 & 217.9 & 217.8\\
       & & & 10.62 & 20.02 & 24.25 & 24.66 & 24.74 & 24.75 & 24.75\\
       & & & 0.4280 & 0.7617 & 0.9530 & 0.9557 & 0.9557 & 0.9558 & 0.9558\\
       & & & 0.6695 & 0.7253 & 0.8910 & 0.8943 & 0.8946 & 0.8946 & 0.8947\\
        \hline
      \end{tabular}}
\end{table}

Table \ref{tab:13} shows results for the grayscale image {\tt Texture
  5}.  For CR = 4, 8 and 16, the New1 multifilter has max PSNR, SSIM and
MS-SSIM values at all decomposition levels.  For CR=32, the SA4
multifilter has maximal values for the first and second decomposition
levels, while New1 has max values for the third to seventh decomposition
levels.

% Table 13
\begin{table}
\centering 
  \renewcommand\thetable{13}
  \caption{Results for {\tt Texture 5}. The four lines for each multiwavelet represent MSE, PSNR, SSIM and MS-SSIM, in this order.}%
     {\label{tab:13}
       \footnotesize
       \setlength\tabcolsep{4pt}
       \renewcommand{\arraystretch}{1.0}
       \begin{tabular}{|c|c|c|c|c|c|c|c|c|}
         \hline
         \multirow{2}{*}{Image} &
         \multirow{2}{*}{CR} &  Multi- &                       
         \multicolumn{6}{c|}{Decomposition level} \\
         \cline{4-9}
         & & wavelet & 1 & 2 & 3 & 4 & 5 & 6 \\
         \hline
         \multirow{57}{*}{\parbox{1.5in}{\includegraphics[width=1.5in]{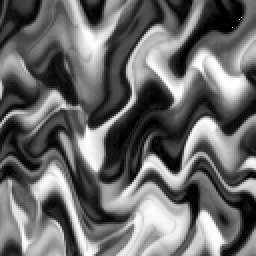}
         \\\centering {\tt Texture 5} }} & \multirow{28}{*}{4:1} &
         \multirow{4}{*}{GHM} & 
          67.96 & 16.95 & 16.25 & 16.22 & 16.21 & 16.21\\
        & & & 29.81 & 35.84 & 36.02 & 36.03 & 36.03 & 36.03\\
        & & & 0.9055 & 0.9605 & 0.9616 & 0.9617 & 0.9617 & 0.9617\\
        & & & 0.9922 & 0.9967 & 0.9968 & 0.9968 & 0.9968 & 0.9968\\
         \cline{3-9}
         & &  \multirow{4}{*}{SA4} &
         11.63 & 7.95 & 7.82 & 7.82 & 7.81 & 7.81\\
        & & & 37.48 & 39.13 & 39.20 & 39.20 & 39.20 & 39.20\\
        & & & 0.9693 & 0.9788 & 0.9793 & 0.9793 & 0.9793 & 0.9793\\
        & & & 0.9972 & 0.9981 & 0.9981 & 0.9981 & 0.9981 & 0.9981\\
          \cline{3-9}
         & &  \multirow{4}{*}{CL} &
		21.75 & 13.21 & 12.89 & 12.89 & 12.89 & 12.89\\
        & & & 34.76 & 36.92 & 37.03 & 37.03 & 37.03 & 37.03\\
        & & & 0.9462 & 0.9661 & 0.9669 & 0.9669 & 0.9669 & 0.9669\\
        & & & 0.9949 & 0.9967 & 0.9968 & 0.9968 & 0.9968 & 0.9968\\
          \cline{3-9}
         & & \multirow{4}{*}{Alpert} &
	     29.08 & 63.98 & 213.98 & 575.5 & 1043 & 1412.5\\
        & & & 33.50 & 30.07 & 24.83 & 20.53 & 17.95 & 16.63\\
        & & & 0.9633 & 0.9298 & 0.8160 & 0.7121 & 0.6869 & 0.6741\\
        & & & 0.9965 & 0.9905 & 0.9495 & 0.8567 & 0.7826 & 0.7496\\
        \cline{3-9}
         & & \multirow{4}{*}{New1} &
	   {\bf 5.71} & {\bf 3.73} & {\bf 3.65} &{\bf  3.65} & {\bf 3.65} &{\bf 3.65}\\
        & & & {\bf 40.57} & {\bf 42.41} & {\bf 42.51 }& {\bf 42.51} & {\bf 42.51} & {\bf 42.51}\\
        & & & {\bf 0.9835} & {\bf 0.9898} & {\bf 0.9900} & {\bf 0.9900} & {\bf 0.9900} & {\bf 0.9900}\\
        & & & {\bf 0.9985} & {\bf 0.9991} &{\bf  0.9991} & {\bf 0.9991} &{\bf  0.9991} & {\bf 0.9991}\\
         \cline{3-9}
         & & \multirow{4}{*}{New2} &
          8.28 & 8.67 & 8.91 & 8.92 & 8.92 & 8.92\\
        & & & 38.95 & 38.75 & 38.63 & 38.63 & 38.63 & 38.63\\
        & & & 0.9766 & 0.9758 & 0.9755 & 0.9755 & 0.9755 & 0.9755\\
        & & & 0.9978 & 0.9977 & 0.9977 & 0.9977 & 0.9977 & 0.9977\\
         \cline{3-9}
         & & \multirow{4}{*}{Integer Haar} &
		7.81 & 7.9 & 7.91 & 7.89 & 7.89 & 7.89\\
        & & & 39.20 & 39.15 & 39.15 & 39.16 & 39.16 & 39.16\\
        & & & 0.9777 & 0.9780 & 0.9780 & 0.9780 & 0.9780 & 0.9780\\
        & & & 0.9976 & 0.9977 & 0.9977 & 0.9977 & 0.9977 & 0.9977\\
         \cline{2-9} \\
         \cline{2-9}
       & \multirow{28}{*}{8:1} &
         \multirow{4}{*}{GHM} & 
         1569.8 & 59.01 & 51.79 & 51.48 & 51.46 & 51.462\\
        & & & 16.17 & 30.42 & 30.99 & 31.01 & 31.02 & 31.02\\
        & & & 0.4348 & 0.9100 & 0.9141 & 0.9146 & 0.9146 & 0.9146\\
        & & & 0.8711 & 0.99100 & 0.9913 & 0.9914 & 0.9914 & 0.9914\\
         \cline{3-9}
         & &  \multirow{4}{*}{SA4} & 
        54.59 & 29.02 & 27.94 & 27.90 & 27.87 & 27.86\\
        & & & 30.76 & 33.50 & 33.67 & 33.67 & 33.68 & 33.68\\
        & & & 0.8931 & 0.9410 & 0.9435 & 0.9435 & 0.9435 & 0.9435\\
        & & & 0.9860 & 0.9936 & 0.9939 & 0.9939 & 0.9939 & 0.9939\\
          \cline{3-9}
         & &  \multirow{4}{*}{CL} &
         95.17 & 46.01 & 43.40 & 43.32 & 43.32 & 43.32\\
        & & & 28.35 & 31.50 & 31.76 & 31.76 & 31.76 & 31.76\\
        & & & 0.8366 & 0.9143 & 0.9183 & 0.9182 & 0.9182 & 0.9182\\
        & & & 0.9816 & 0.9900 & 0.9905 & 0.9905 & 0.9905 & 0.9905\\
          \cline{3-9}
         & & \multirow{4}{*}{Alpert} &
        61.65 & 80.89 & 229.31 & 590.77 & 1058.3 & 1427.7\\
        & & & 30.23 & 29.05 & 24.53 & 20.42 & 17.88 & 16.58\\
        & & & 0.9070 & 0.9002 & 0.7934 & 0.6947 & 0.6709 & 0.6588\\
        & & & 0.9882 & 0.9865 & 0.9448 & 0.8515 & 0.7771 & 0.7441\\
	     \cline{3-9}
         & & \multirow{4}{*}{New1} &
         {\bf 40.33} & {\bf 20.62} & {\bf 19.31} &{\bf  19.27} &{\bf 19.27} & {\bf 19.27}\\
        & & & {\bf 32.07} & {\bf 34.99} & {\bf 35.27} & {\bf 35.28} & {\bf 35.28} & {\bf 35.28}\\
        & & & {\bf 0.9136} & {\bf 0.9537} & {\bf 0.9555} & {\bf 0.9556} & {\bf 0.9556} & {\bf 0.9556}\\
        & & & {\bf 0.9889} & {\bf 0.9950} & {\bf 0.9952} & {\bf 0.9952} &{\bf 0.9952} & {\bf 0.9952}\\
         \cline{3-9}
         & & \multirow{4}{*}{New2} &
         44.37 & 44.23 & 47.84 & 47.99 & 47.968 & 47.968\\
        & & & 31.66 & 31.67 & 31.33 & 31.32 & 31.32 & 31.32\\
        & & & 0.9074 & 0.9107 & 0.9051 & 0.9051 & 0.9052 & 0.9052\\
        & & & 0.9864 & 0.9867 & 0.9857 & 0.9857 & 0.9857 & 0.9857\\
         \cline{3-9}
         & & \multirow{4}{*}{Integer Haar} &
         49.65 & 47.34 & 46.84 & 46.88 & 46.90& 46.889\\
        & & & 31.17 & 31.38 & 31.42 & 31.42 & 31.42 & 31.42\\
        & & & 0.9002 & 0.9003 & 0.9012 & 0.9011 & 0.9010 & 0.9011\\
        & & & 0.9856 & 0.9862 & 0.9868 & 0.9868 & 0.9868 & 0.9868\\
       \hline
      \end{tabular}}{}
\end{table}
% Table 13(continued)
\begin{table}
\centering 
  \renewcommand\thetable{13(continued)}
  \caption{Results for {\tt Texture 5}. The four lines for each multiwavelet represent MSE, PSNR, SSIM and MS-SSIM, in this order.}%
     {
       \footnotesize
       \setlength\tabcolsep{4pt}
       \renewcommand{\arraystretch}{1.0}
       \begin{tabular}{|c|c|c|c|c|c|c|c|c|}
         \hline
         \multirow{2}{*}{Image} &
         \multirow{2}{*}{CR} &  Multi- &                       
         \multicolumn{6}{c|}{Decomposition level} \\
         \cline{4-9}
         & & wavelet & 1 & 2 & 3 & 4 & 5 & 6 \\
         \hline
         \multirow{57}{*}{\parbox{1.5in}{\includegraphics[width=1.5in]{Texture5}
         \\\centering {\tt Texture 5} }} & \multirow{28}{*}{16:1} &
         \multirow{4}{*}{GHM} & 
          5720 & 170.5 & 118.2 & 116.5 & 116.4 & 116.4\\
        & & & 10.56 & 25.81 & 27.41 & 27.47 & 27.47 & 27.47\\
        & & & 0.1313 & 0.8197 & 0.8578 & 0.8594 & 0.8596 & 0.8596\\
        & & & 0.6833 & 0.9729 & 0.9810 & 0.9812 & 0.9813 & 0.9813\\
         \cline{3-9}
         & &  \multirow{4}{*}{SA4} &
       221.13 & 80.21 & 71.77 & 71.32 & 71.22 & 71.2\\
        & & & 24.68 & 29.09 & 29.57 & 29.60 & 29.61 & 29.61\\
        & & & 0.7388 & 0.8777 & 0.8902 & 0.8908 & 0.8910 & 0.8910\\
        & & & 0.9544 & 0.9819 & 0.9845 & 0.9845 & 0.9845 & 0.9845\\
          \cline{3-9}
         & &  \multirow{4}{*}{CL} &
		341.3 & 116 & 102.3 & 101.7 & 101.6 & 101.6\\
        & & & 22.80 & 27.49 & 28.03 & 28.06 & 28.06 & 28.06\\
        & & & 0.6459 & 0.8391 & 0.8558 & 0.8560 & 0.8563 & 0.8564\\
        & & & 0.9508 & 0.9751 & 0.9786 & 0.9787 & 0.9787 & 0.9787\\
          \cline{3-9}
         & & \multirow{4}{*}{Alpert} &
	    208.5 & 129.4 & 267.8 & 629.1 & 1097 & 1466\\
        & & & 24.94 & 27.01 & 23.85 & 20.14 & 17.73 & 16.47\\
        & & & 0.7655 & 0.8446 & 0.7545 & 0.6628 & 0.6411 & 0.6299\\
        & & & 0.9610 & 0.9752 & 0.9347 & 0.8403 & 0.7653 & 0.7323\\
        \cline{3-9}
         & & \multirow{4}{*}{New1} &
	     {\bf 195.64} & {\bf 72.66} & {\bf 61.17} &{\bf  60.82} &{\bf  60.85} & {\bf 60.84}\\
        & & & {\bf 25.22} & {\bf 29.52} & {\bf 30.27} & {\bf 30.29} & {\bf 30.29} & {\bf 30.29}\\
        & & & {\bf 0.7651} & {\bf 0.8808} & {\bf 0.8967} & {\bf 0.8971} & {\bf 0.8972} & {\bf 0.8971}\\
        & & & {\bf 0.9609 }& {\bf 0.9818} & {\bf 0.9849} & {\bf 0.9850} & {\bf 0.9850} & {\bf 0.9850}\\
         \cline{3-9}
         & & \multirow{4}{*}{New2} &
        215.9 & 149.6 & 166.1 & 168.9 & 168.7 & 168.7\\
        & & & 24.79 & 26.38 & 25.93 & 25.85 & 25.86 & 25.86\\
        & & & 0.7518 & 0.7897 & 0.7724 & 0.7691 & 0.7692 & 0.7691\\
        & & & 0.9521 & 0.9556 & 0.9499 & 0.9493 & 0.9494 & 0.9495\\
         \cline{3-9}
         & & \multirow{4}{*}{Integer Haar} &
		214.67 & 149.7 & 148.24 & 148.5 & 148.8 & 148.6\\
        & & & 24.81 & 26.38 & 26.42 & 26.41 & 26.41 & 26.41\\
        & & & 0.7568 & 0.7772 & 0.7764 & 0.7760 & 0.7756 & 0.7757\\
        & & & 0.9577 & 0.9588 & 0.9612 & 0.9613 & 0.9613 & 0.9613\\
         \cline{2-9} \\
         \cline{2-9}
         & \multirow{28}{*}{32:1} &
         \multirow{4}{*}{GHM} & 
        9408 & 1443 & 217.3 & 206.2 & 205.4 & 205.4\\
        & & & 8.4 & 16.54 & 24.76 & 24.99 & 25.01 & 25.01\\
        & & & 0.0538 & 0.4411 & 0.7761 & 0.7836 & 0.7843 & 0.7843\\
        & & & 0.5223 & 0.8358 & 0.9554 & 0.9579 & 0.9582 & 0.9582\\
         \cline{3-9}
         & &  \multirow{4}{*}{SA4} & 
        {\bf 1699} & {\bf 205.3} & 154.4 & 152.7 & 152.3 & 152.1\\
        & & & {\bf 15.83} & {\bf 25.01} & 26.24 & 26.29 & 26.30 & 26.31\\
        & & & 0.4132 & {\bf 0.7512} & 0.8132 & 0.8157 & 0.8161 & 0.8162\\
        & & & {\bf 0.8201} & {\bf 0.9469} & 0.9634 & 0.9644 & 0.9645 & 0.9646\\
          \cline{3-9}
         & &  \multirow{4}{*}{CL} &
         1932 & 264.5 & 207 & 203.7 & 203.5 & 203.4\\
        & & & 15.27 & 23.91 & 24.97 & 25.04 & 25.05 & 25.05\\
        & & & 0.3508 & 0.7098 & 0.7720 & 0.7757 & 0.7761 & 0.7760\\
        & & & 0.8272 & 0.9350 & 0.9545	& 0.9556 & 0.9557 & 0.9557\\
          \cline{3-9}
         & & \multirow{4}{*}{Alpert} &
        1732 & 260.1 & 341.5 & 700.4 & 1168 & 1538\\
        & & & 15.74 & 23.98 & 22.8 & 19.68 & 17.46 & 16.26\\
        & & & 0.4198 & 0.7234 & 0.7021 & 0.6192 & 0.5989 & 0.5881\\
        & & & 0.8061 & 0.9380 & 0.9161 & 0.8205 & 0.7446 & 0.7112\\
	     \cline{3-9}
         & & \multirow{4}{*}{New1} &
         1732 & 215.8 & {\bf 150.5} & {\bf 148.9} & {\bf 148.7} & {\bf 148.7}\\
        & & & 15.74 & 24.79 & {\bf 26.36} & {\bf 26.4} & {\bf 26.41} & {\bf 26.41}\\
        & & & {\bf 0.4198} & 0.7344 & {\bf 0.8105} & {\bf 0.8126} &{\bf  0.8127} & {\bf 0.8127}\\
        & & & 0.8061 & 0.9387 & {\bf 0.9616} & {\bf 0.9624} & {\bf 0.9623} & {\bf 0.9623}\\
         \cline{3-9}
         & & \multirow{4}{*}{New2} &
         1823 & 371.5 & 398.3 & 411.7 & 412.2 & 412.1\\
        & & & 15.52 & 22.43 & 22.13 & 21.98 & 21.98 & 21.98\\
        & & & 0.4163 & 0.6328 & 0.6186 & 0.6038 & 0.6031 & 0.6031\\
        & & & 0.8126 & 0.8987 & 0.8844 & 0.8824 & 0.8822 & 0.8823\\
         \cline{3-9}
         & & \multirow{4}{*}{Integer Haar} &
        1732 & 347.1 & 327.9 & 329.9 & 331.3 & 331\\
        & & & 15.74 & 22.73 & 22.97 & 22.95 & 22.93 & 22.93\\
        & & & 0.4196 & 0.6340 & 0.6382 & 0.6331 & 0.6319 & 0.6323\\
        & & & 0.8061 & 0.9101 & 0.9176 & 0.9171 & 0.9169 & 0.9170\\
       \hline
      \end{tabular}}{}
\end{table}

Table \ref{tab:14} shows results for the color image {\tt Subwater}.
For CR= 8 all maximal values are obtained for SA4, slightly higher (by
0.17dB and 0.004dB, respectively) than New1; the values of MS-SSIM are
equal.  For CR=16, max PSNR, SSIM ans MS-SSIM values are obtained by
New1, except at the second level, where the values are negligible lower
(by 0.06dB, 0.0007dB and 0.0001dB, respectively) than SA4. Note that SA4
and New1 multifilters have constant MS-SSIM from the third to the
seventh decomposition level.  For CR=32, all maximal values are obtained
for New1, significantly higher than the others (by around 1.00dB for CL,
around 0.70dB) for SA4 and even by 3.00dB for GHM).

% Table 14
\begin{table}
\centering 
  \renewcommand\thetable{14}
  \caption{Results for {\tt Subwater}. The four lines for each multiwavelet represent MSE, PSNR, SSIM and MS-SSIM, in this order.}%
     {\label{tab:14}
       \footnotesize
       \setlength\tabcolsep{4pt}
       \renewcommand{\arraystretch}{1.0}
       \begin{tabular}{|c|c|c|c|c|c|c|c|}
         \hline
         \multirow{2}{*}{Image} &
         \multirow{2}{*}{CR} &  Multi- &                       
         \multicolumn{5}{c|}{Decomposition level} \\
         \cline{4-8}
         & & wavelet & 1 & 2 & 3 & 4 & 5 \\
         \hline
         \multirow{57}{*}{\parbox{1.5in}{\includegraphics[width=1.5in]{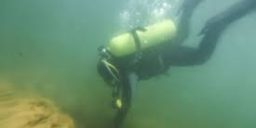}
         \\\centering {\tt Subwater} }} & \multirow{28}{*}{8:1} &
         \multirow{4}{*}{GHM} & 
          4140 & 0.9046 & 0.703 & 0.6931 & 0.6917\\
        & & & 6.91 & 31.22 & 31.95 & 32.00 & 32.00\\
        & & & 0.0569 & 0.6589 & 0.6601 & 0.6601 & 0.6602\\
        & & & 0.3204 & 0.6656 & 0.6657 & 0.6657 & 0.6657\\
         \cline{3-8}
         & &  \multirow{4}{*}{SA4} &
         {\bf 0.8359} & {\bf 0.4812} & {\bf 0.4567} & {\bf 0.4561} & {\bf 0.4569}\\
        & & & {\bf 31.45} & {\bf 33.05} & {\bf 33.19} & {\bf 33.20} & {\bf 33.19}\\
        & & & {\bf 0.6581} & {\bf 0.6617} & {\bf 0.6619} & {\bf 0.6619} & {\bf 0.6619}\\
        & & & {\bf 0.6651} & {\bf 0.6659} & {\bf 0.6659} & {\bf 0.6659} & {\bf 0.6659}\\
          \cline{3-8}
         & &  \multirow{4}{*}{CL} &
		1.004 & 0.5448 & 0.5036 & 0.5002 & 0.499\\
        & & & 30.91 & 32.68 & 32.91 & 32.93 & 32.94\\
        & & & 0.6566 & 0.6611	& 0.6614 & 0.6615 & 0.6615\\
        & & & 0.6652 & 0.6657 & 0.6658 & 0.6658 & 0.6658\\
          \cline{3-8}
         & & \multirow{4}{*}{Alpert} &
	    1.443 & 2.786 & 7.515 & 15.13 & 39.26\\
        & & & 29.86 & 27.95 & 25.11 & 23.11 & 20.45\\
        & & & 0.6554 & 0.6518 & 0.6399 & 0.6309 & 0.6255\\
        & & & {\bf 0.6651} & 0.6641 & 0.6561 & 0.6420 & 0.6150\\
        \cline{3-8}
         & & \multirow{4}{*}{New1} &
	     0.9442 & 0.5117 & 0.4716 & 0.4686 & 0.4687\\
        & & & 31.10 & 32.87 & 33.11 & 33.13 & 33.13\\
        & & & 0.6569 & 0.6613 & 0.6616 & 0.6617 & 0.6617\\
        & & & 0.6650 & {\bf 0.6659} & {\bf 0.6659} & {\bf 0.6659} & {\bf 0.6659}\\
         \cline{3-8}
         & & \multirow{4}{*}{New2} &
        1.114 & 0.8457 & 0.844 & 0.8431 & 0.8438\\
        & & & 30.61 & 31.41 & 31.41 & 31.42 & 31.41\\
        & & & 0.6557 & 0.6585 & 0.6585 & 0.6585 & 0.6585\\
        & & & 0.6646 & 0.6651 & 0.6651 & 0.6651 & 0.6651\\
         \cline{3-8}
         & & \multirow{4}{*}{Integer Haar} &
		1.385 & 0.9685 & 0.9087 & 0.9023 & 0.9031\\
        & & & 29.98 & 31.01 & 31.20 & 31.22 & 31.22\\
        & & & 0.6532 & 0.6572 & 0.6577 & 0.6577 & 0.6577\\
        & & & 0.6644 & 0.6650 & 0.6651 & 0.6651 & 0.6651\\
         \cline{2-8} \\
         \cline{2-8}
        & \multirow{28}{*}{16:1} &
         \multirow{4}{*}{GHM} & 
        7027 & 29.38 & 2.214 & 2.063 & 2.053\\
        & & & 5.35 & 21.28 & 28.63 & 28.83 & 28.84\\
        & & & 0.0103 & 0.6278 & 0.6512 & 0.6520 & 0.6521\\
        & & & 0.3318 & 0.6234 & 0.6634 & 0.6635 & 0.6635\\
         \cline{3-8}
         & &  \multirow{4}{*}{SA4} & 
        43.59 & {\bf 1.647} & 1.372 & 1.355 & 1.36\\
        & & & 20.07 & {\bf 29.48} & 30.01 & 30.04 & 30.03\\
        & & & {\bf 0.6121} & {\bf 0.6532} & {\bf 0.6556} & {\bf 0.6557} & {\bf 0.6557}\\
        & & & 0.6073 & {\bf 0.6636} & 0.6642 & 0.6642 & 0.6642\\
          \cline{3-8}
         & &  \multirow{4}{*}{CL} &
         37.81 & 1.914 & 1.523 & 1.482 & 1.474\\
        & & & 20.50 & 29.05 & 29.70 & 29.78 & 29.80\\
        & & & 0.6109 & 0.6513 & 0.6543 & 0.6545 & 0.6546\\
        & & & 0.6218 & 0.6631 & 0.6639 & 0.6639 & 0.6639\\
          \cline{3-8}
         & & \multirow{4}{*}{Alpert} &
        {\bf 13.6} & 3.72 & 8.119 & 15.7 & 39.82\\
        & & & {\bf 23.38} & 27.11 & 24.88 & 23.00 & 20.40\\
        & & & 0.6079 & 0.6451 & 0.6355 & 0.6268 & 0.6214\\
        & & & 0.6555 & 0.6625 & 0.6549 & 0.6407 & 0.6137\\
	     \cline{3-8}
         & & \multirow{4}{*}{New1} &
         {\bf 13.6} & 1.681 & {\bf 1.322 }& {\bf 1.284} & {\bf 1.282}\\
        & & & {\bf 23.38} & 29.42 & {\bf 30.11} & {\bf 30.20} & {\bf 30.20}\\
        & & & 0.6079 & 0.6525 & 0.6552 & 0.6555 & 0.6555\\
        & & & {\bf 0.6555} & 0.6635 & {\bf 0.6643} & {\bf 0.6643} & {\bf 0.6643}\\
         \cline{3-8}
         & & \multirow{4}{*}{New2} &
         47.96 & 3.24 & 3.304 & 3.273 & 3.29\\
        & & & 19.80 & 27.52 & 27.46 & 27.48 & 27.47\\
        & & & 0.6038 & 0.6406 & 0.6405 & 0.6405 & 0.6403\\
        & & & 0.6047 & 0.6599 & 0.6597 & 0.6598 & 0.6598\\
         \cline{3-8}
         & & \multirow{4}{*}{Integer Haar} &
        13.6 & 2.953 & 2.553 & 2.516 & 2.519\\
        & & & 23.38 & 27.78 & 28.20 & 28.24 & 28.24\\
        & & & 0.60790 & 0.6430 & 0.6460 & 0.6463 & 0.6462\\
        & & & 0.6555 & 0.6608 & 0.6622 & 0.6623 & 0.6622\\
       \hline
      \end{tabular}}{}
\end{table}
% Table 14(continued)
\begin{table}
\centering 
  \renewcommand\thetable{14(continued)}
  \caption{Results for {\tt Subwater}. The four lines for each multiwavelet represent MSE, PSNR, SSIM and MS-SSIM, in this order.}%
     {
       \footnotesize
       \setlength\tabcolsep{4pt}
       \renewcommand{\arraystretch}{1.0}
       \begin{tabular}{|c|c|c|c|c|c|c|c|}
         \hline
         \multirow{2}{*}{Image} &
         \multirow{2}{*}{CR} &  Multi- &                       
         \multicolumn{5}{c|}{Decomposition level} \\
         \cline{4-8}
         & & wavelet & 1 & 2 & 3 & 4 & 5 \\
         \hline
         \multirow{27}{*}{\parbox{1.5in}{\includegraphics[width=1.5in]{subwater1}
         \\\centering {\tt Subwater} }} & \multirow{28}{*}{32:1} &
         \multirow{4}{*}{GHM} & 
         8976 & 3250 & 9.484 & 6.745 & 6.603\\
        & & & 4.63 & 7.62 & 24.46 & 25.42 & 25.48\\
        & & & 0.0065 & 0.0652 & 0.6322 & 0.6362 & 0.6366\\
        & & & 0.3774 & 0.2303 & 0.6523 & 0.6553 & 0.6555\\
         \cline{3-8}
         & &  \multirow{4}{*}{SA4} &
        3944 & 6.575 & 4.264 & 4.018 & 4.036\\
        & & & 7.09 & 25.48 & 26.72 & 26.89 & 26.88\\
        & & & 0.2598 & 0.6276 & 0.6409 & 0.6423 & 0.6422\\
        & & & 0.3466 & 0.6531 & 0.6578 & 0.6583 & 0.6583\\
          \cline{3-8}
         & &  \multirow{4}{*}{CL} &
		3977.4 & 7.37 & 4.771 & 4.271 & 4.222\\
        & & & 7.08 & 25.14 & 26.39 & 26.71 & 26.74\\
        & & & 0.2638 & 0.6221 & 0.6382 & 0.6407 & 0.6409\\
        & & & 0.3683 & 0.6523 & 0.6572 & 0.6585 & 0.6586\\
          \cline{3-8}
         & & \multirow{4}{*}{Alpert} &
	    {\bf 3955.2} & 6.846 & 9.572 & 16.943 & 41.073\\
        & & & {\bf 7.1} & 25.35 & 24.4 & 22.77 & 20.31\\
        & & & {\bf 0.2626} & 0.6251 & 0.6275 & 0.6203 & 0.6150\\
        & & & {\bf 0.3969} & 0.6543 & 0.6523 & 0.6382 & 0.6112\\
        \cline{3-8}
         & & \multirow{4}{*}{New1} &
	    {\bf 3955} & {\bf 5.416} & {\bf 3.4} & {\bf 3.155} &{\bf 3.137}\\
        & & & {\bf 7.10} & {\bf 26.02} & {\bf 27.37} & {\bf 27.59} & {\bf 27.60}\\
        & & & {\bf 0.2626} & {\bf 0.6282} & {\bf 0.6430} & {\bf 0.6444} & {\bf 0.6446}\\
        & & & {\bf 0.3968} & {\bf 0.6542} & {\bf 0.6603} & {\bf 0.6606} & {\bf 0.6607}\\
         \cline{3-8}
         & & \multirow{4}{*}{New2} &
        4046 & 11.35 & 10.11 & 9.914 & 10.04\\
        & & & 7.02 & 23.88 & 24.22 & 24.27 & 24.23\\
        & & & 0.2552 & 0.6038 & 0.6070 & 0.6066 & 0.6059\\
        & & & 0.3440 & 0.6463 & 0.6452 & 0.646 & 0.6458\\
         \cline{3-8}
         & & \multirow{4}{*}{Integer Haar} &
		3955.2 & 7.689 & 5.9762 & 5.6887 & 5.6662\\
        & & & 7.10 & 25.01 & 25.74 & 25.88 & 25.89\\
        & & & 0.2626 & 0.6164 & 0.6275 & 0.6289 & 0.6290\\
        & & & 0.3968 & 0.6507 & 0.6557 & 0.6562 & 0.6563\\
        \hline
      \end{tabular}}{}
\end{table}

\subsubsection{Signal denoising}
Multiwavelet denoising is a non-linear process, which depends heavily on the choice of
a thresholding parameter. For a better comparison of the new supercompact and 
orthogonal multifilters New1 and New2 with orthogonal GHM, SA4, CL, Alpert and Integer Haar 
multifilters, noise reduction in 1D and 2D is considered.

\paragraph{1D denoising} 

\begin{figure}
\centering 
  \includegraphics[width=6in]{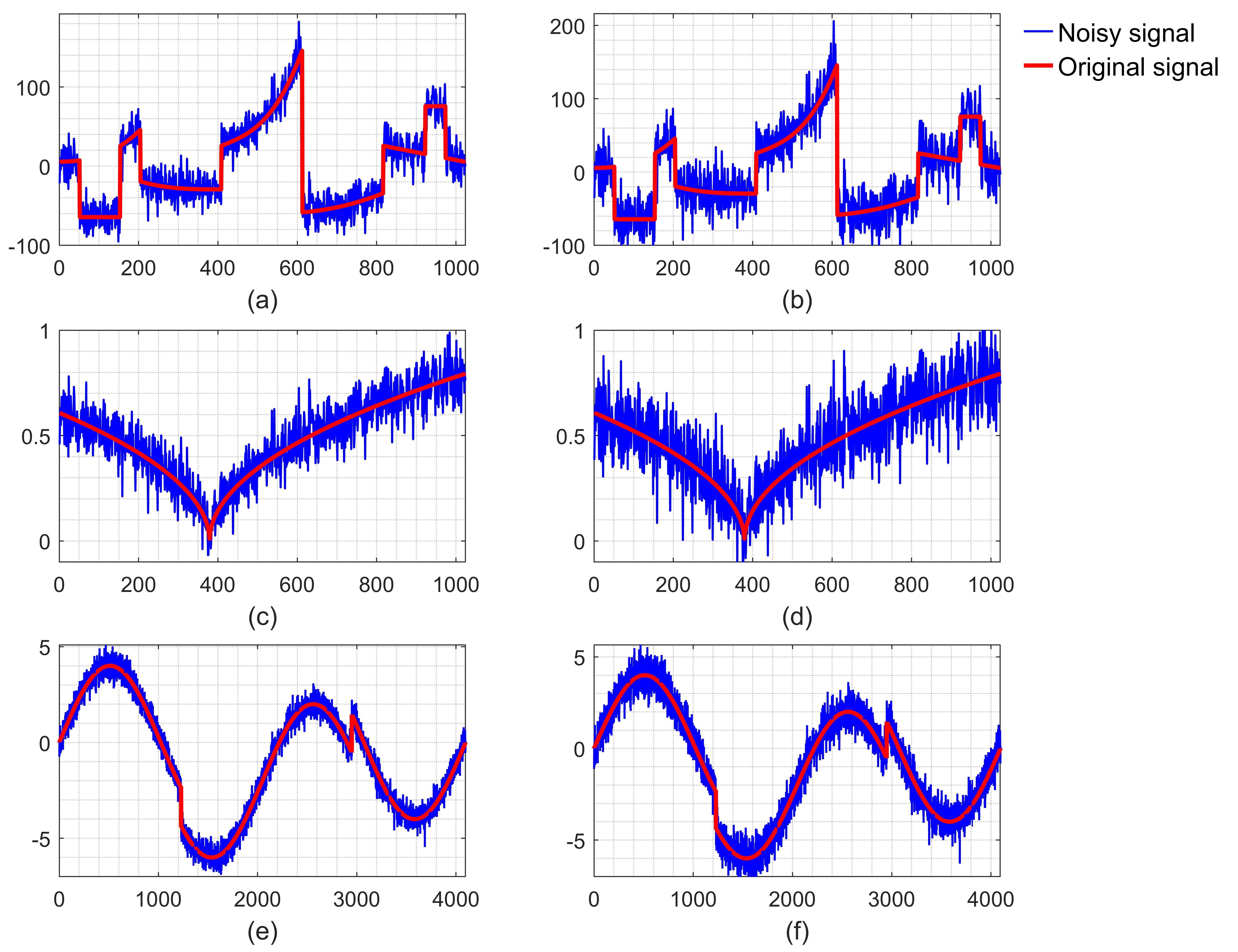}
  \caption{Original and Noisy 1D signals with added AWGN with variance
    $\sigma$ = 10, 15; 
(a) {\tt Piece-Polynomial} with variance $\sigma$ = 10; (b) {\tt Piece-Polynomial} with variance $\sigma$ = 15; 
(c) {\tt Cusp} with variance $\sigma$ = 10; (d) {\tt Cusp} with variance $\sigma$ = 15; 
(e) {\tt HeaviSine} with variance $\sigma$ = 10; (f) {\tt HeaviSine} with variance $\sigma$ = 15.}
  \label{fig:09}
\end{figure}

\begin{figure}
\centering 
  \includegraphics[height=0.2in]{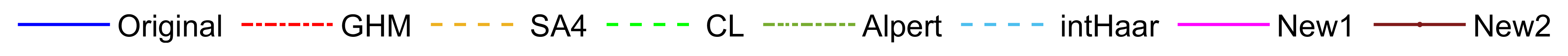}\\
 \includegraphics[height=1.95in]{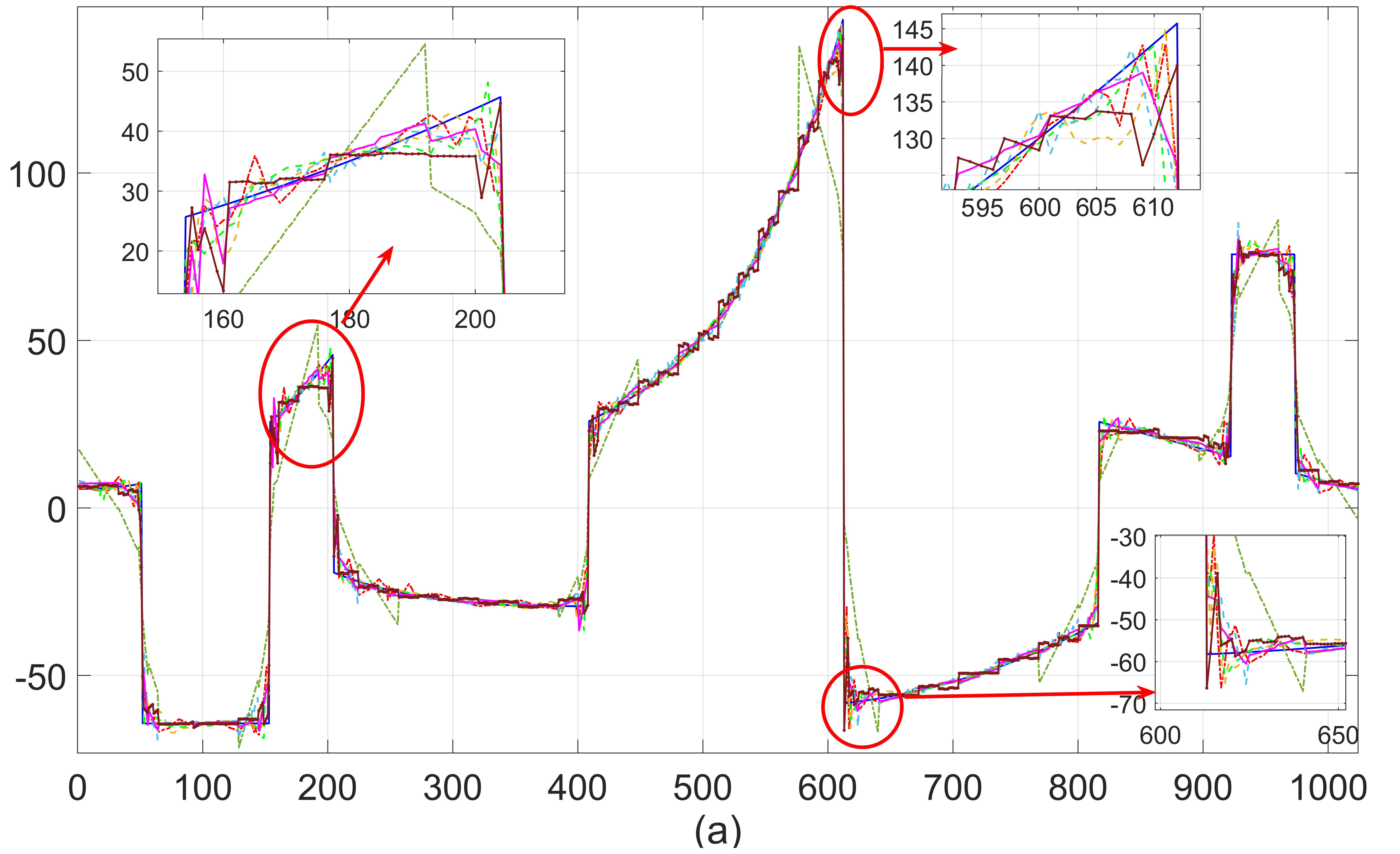}\includegraphics[height=1.95in]{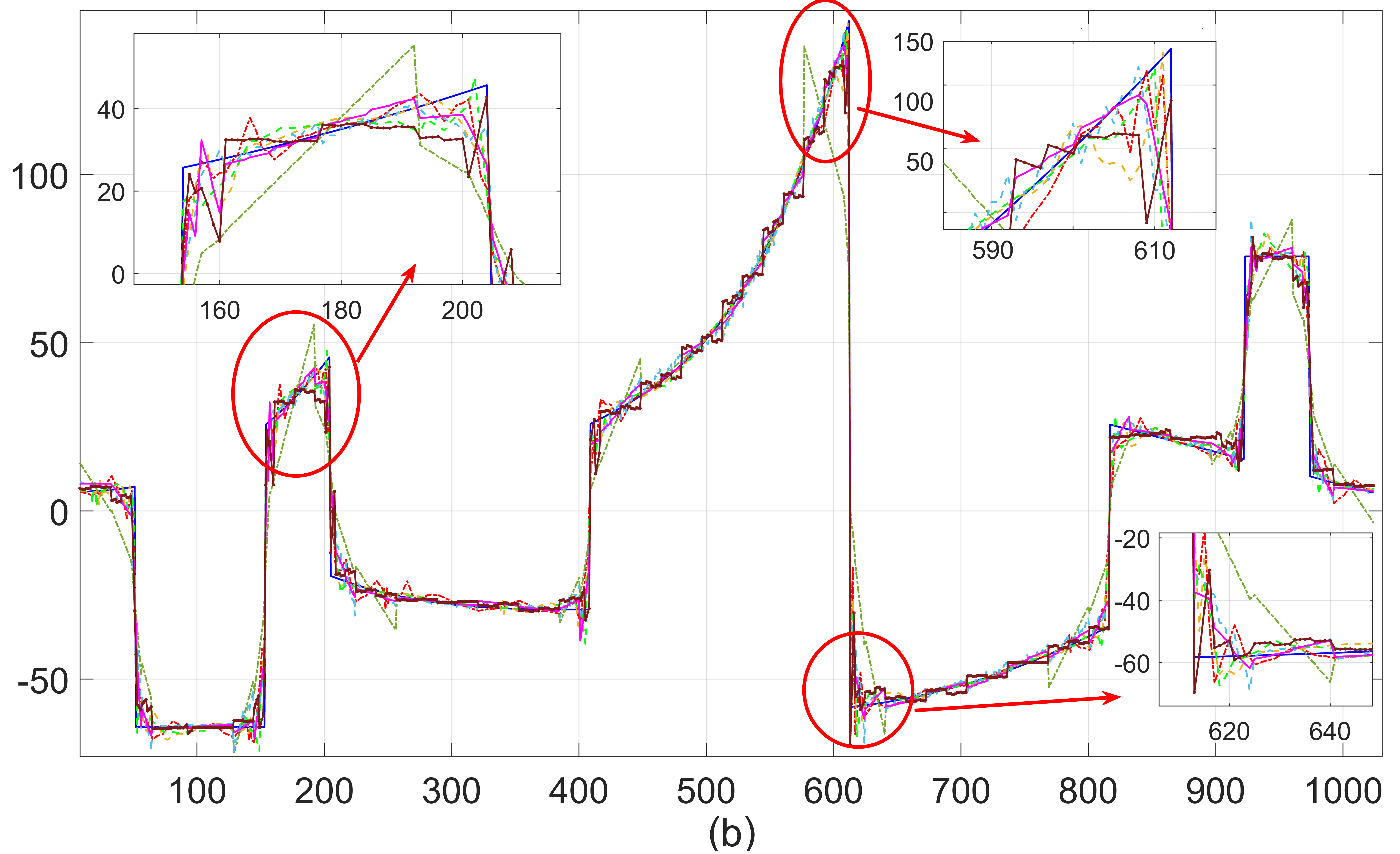}\\ 
 \includegraphics[height=1.95in]{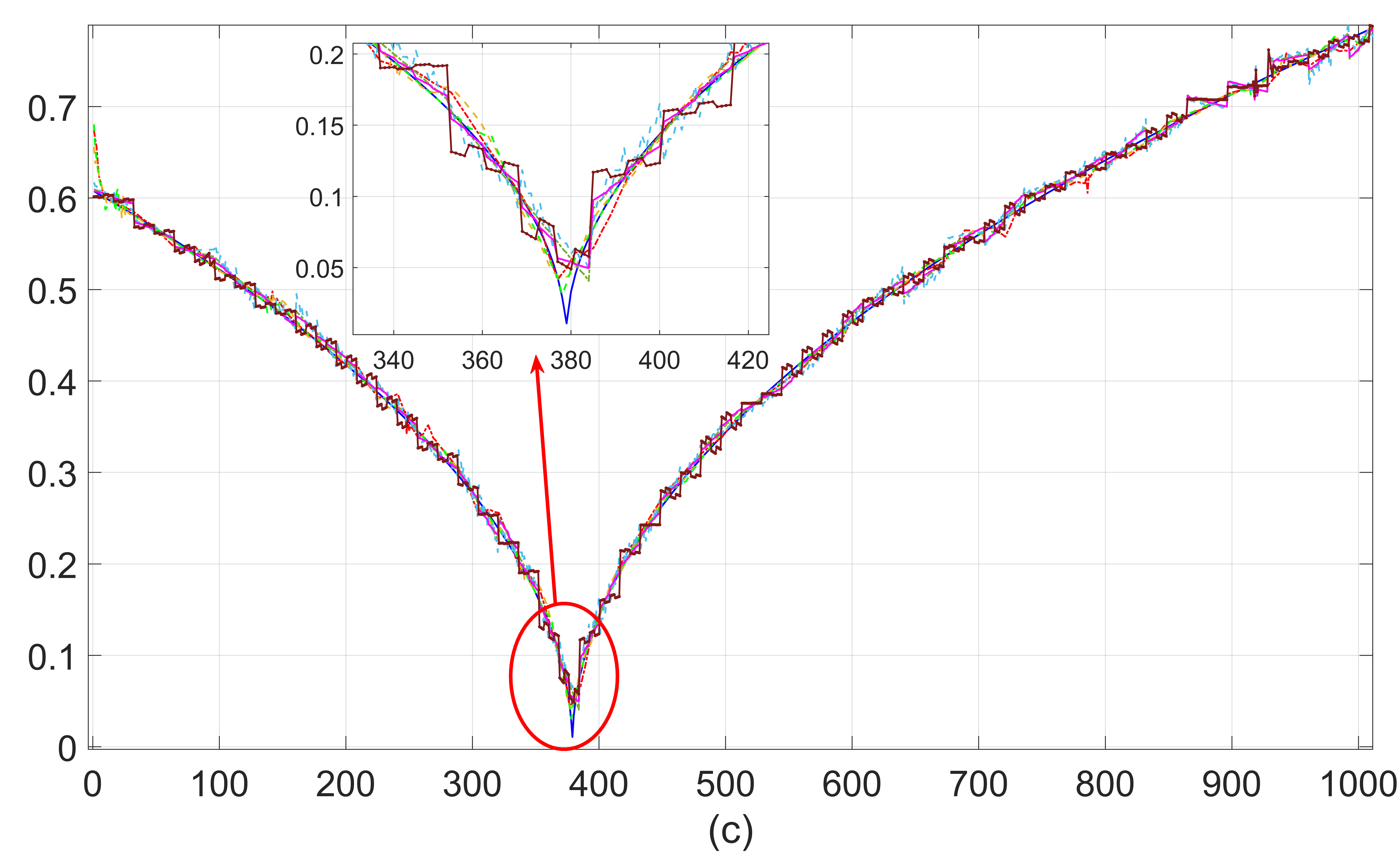}\includegraphics[height=1.95in]{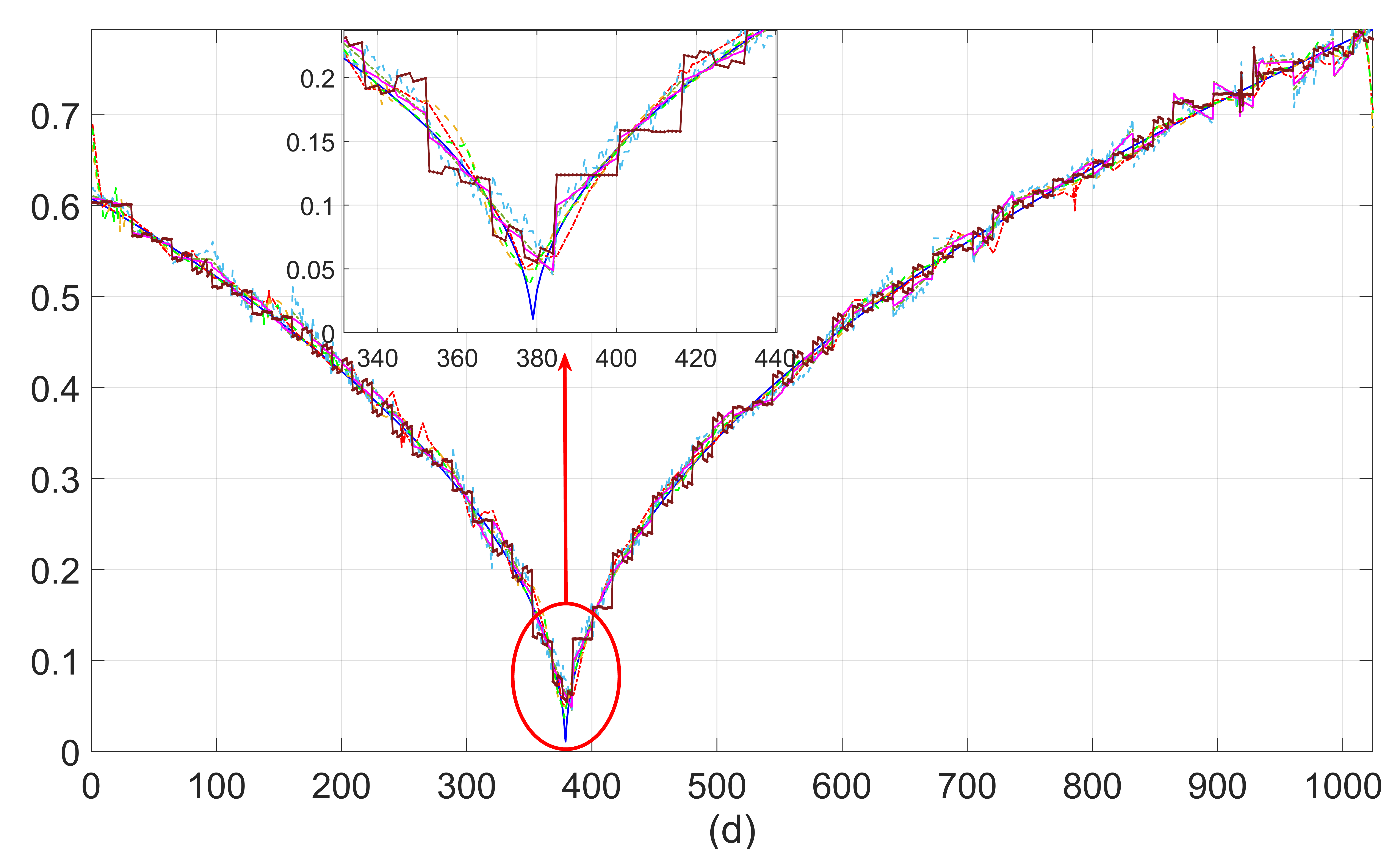}\\
 \includegraphics[height=1.95in]{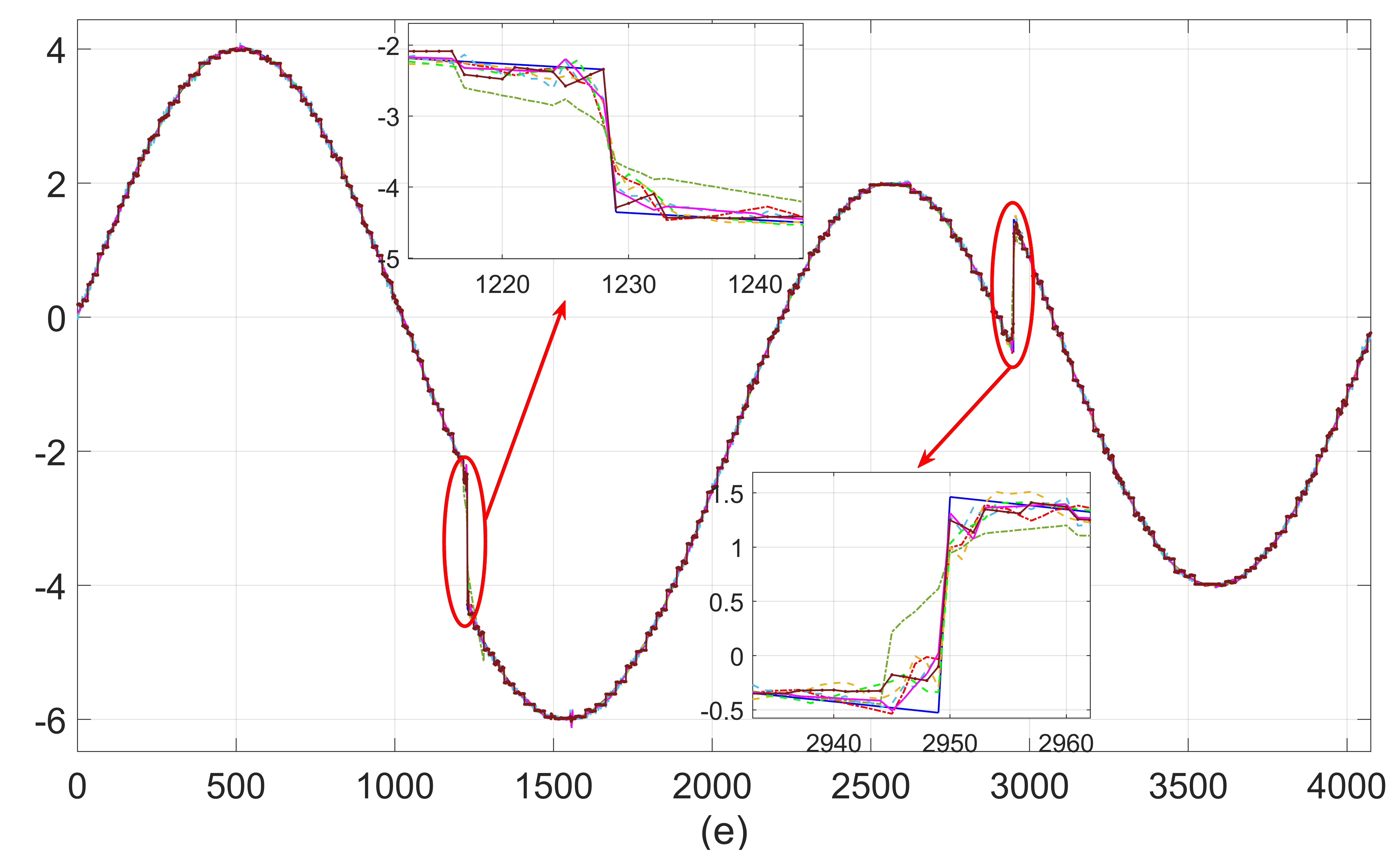}\includegraphics[height=1.95in]{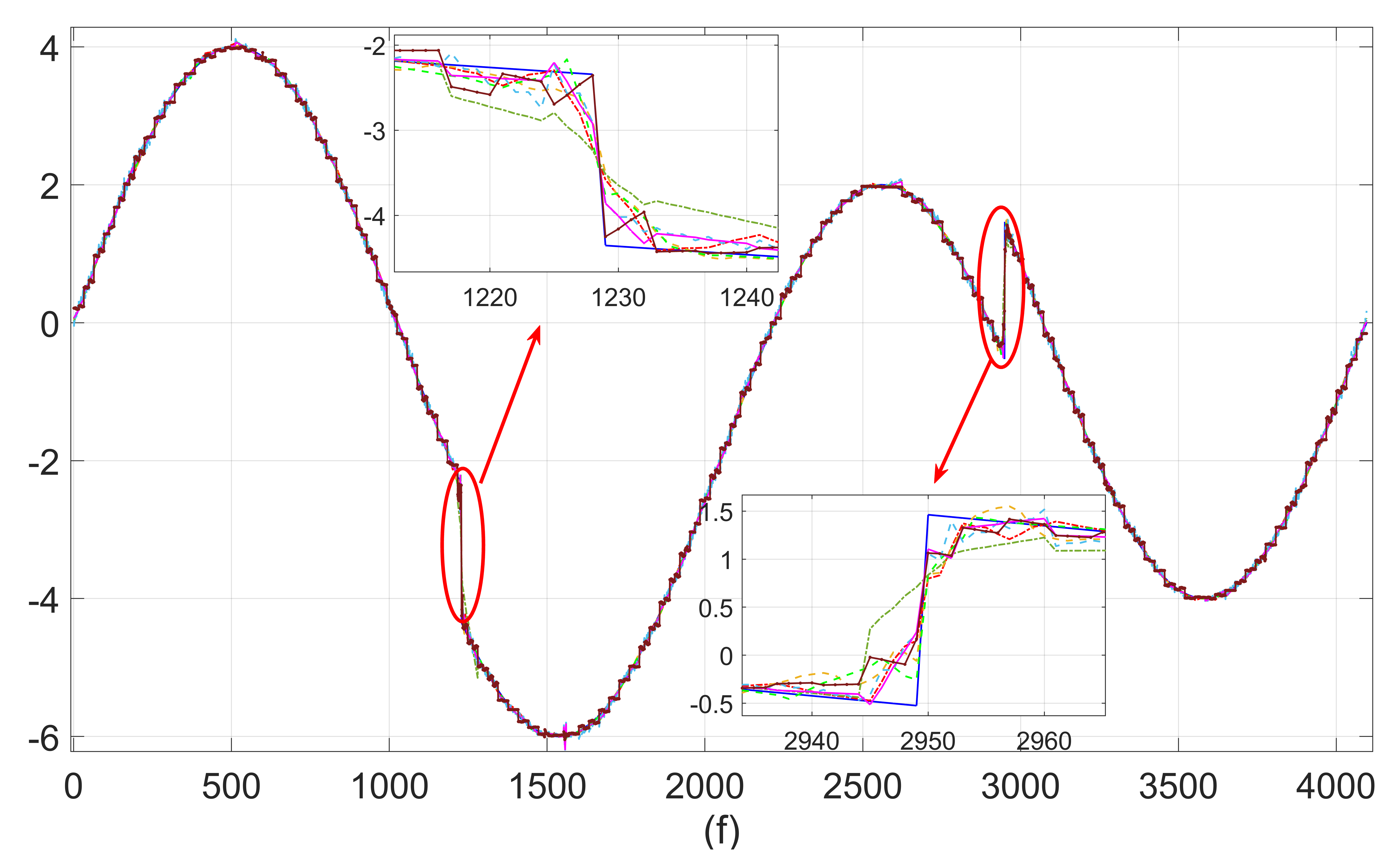}\\
\caption{Denoising of 1D signals with added AWGN with variance $\sigma$ = 10, 15 by using orthogonal multiwavelet banks and applied soft thresholding; 
(a) {\tt Piece-Polynomial} with variance $\sigma$ = 10; (b) {\tt Piece-Polynomial} with variance $\sigma$ = 15; 
(c) {\tt Cusp} with variance $\sigma$ = 10; (d) {\tt Cusp} with variance $\sigma$ = 15; 
(e) {\tt HeaviSine} with variance $\sigma$ = 10; (f) {\tt HeaviSine} with variance $\sigma$ = 15.}
  \label{fig:10}
\end{figure}

We apply denoising with seven orthogonal multifilters (GHM, SA4, CL,
Alpert, Integer Haar, New1 and New2) to three 1D three test signals
\textit{s}(t) ({\tt Piece-Polynomial} and {\tt Cusp} with a length of
1024 values and {\tt HeaviSine} with 4096), with two different additive
white Gaussian noise (AWGN) values ($\sigma$=10, 15).  The original,
noisy signals are shown in Fig.~\ref{fig:09}(a)-(f). The denoised
signals, with a zoom of some parts are shown in
Fig.~\ref{fig:10}(a)-(f).  The {\tt Piece-Polynomial} noisy signal is
more complex; it consists of exponential and others polynomial segments
with edges and jumps.

The zoom images of the simulation plots give us enough information to
recognize the quality of approximation for all multifilters.  For all
three test signals and both noise values, Fig.~\ref{fig:10}(c) and (d)
show that the New2 multifilter has the best plots for cruder numerical
approximations. The scaling part of the {\tt HeaviSine} signal shows
better approximations of sudden changes by both the New2 and New1
multifilters.  The zoom plots of {\tt Piece-Polynomial} and {\tt Cusp}
signals for $\sigma$=10, 15 show better approximation using the New1 and
GHM multifilters, as shown in Fig.~\ref{fig:10}(a),(b),(c) and (d).
Using the Alpert multifilter for {\tt Piece-Polynomial} and {\tt Cusp}
signals leads to spurious peak values.

To complete the analysis, we use the maximal absolute error (MAE):
\begin{equation}\label{eq:epsilon}
\centering 
\epsilon =||s(t)- \hat{s}(t)||_{\infty} . 
\end{equation}

The MAE values are tabulated in Table~\ref{tab:15}.

Since the {\tt Piece-Polynomial} noisy signal is more complex, its MAEs
are bigger than for other signals.  We see that the minimal MAE values
for $\sigma$=10 are obtained by both the New1 and New2 multifilters, for
$\sigma$=15 only by New1.  The noise suppression results obtained for
the {\tt Cusp} signal shows again minimal MAE values for both New1 and
New2 multifilters for $\sigma$=10, 15.  The results for the {\tt
  HeaviSine} signal show minimal MAE value using for $\sigma$=10, 15.  A
comparison of the results shows that the proposed New1 and New2
multifilters can provide more effective noise suppression and better
approximation than the other five multifilters.

% Figure 09 
% Figure 10 

% Table 15
\begin{table}
\centering 
  \renewcommand\thetable{15}
  \caption{The $\epsilon$ values for denoising of 1D {\tt Piece-Polynomial},{\tt Cusp} and {\tt HeaviSine} signals by using orthogonal multiwavelet filters.}%
    {\label{tab:15}
     \footnotesize
       \setlength\tabcolsep{5pt}
       \renewcommand{\arraystretch}{1.1}
       \begin{tabular}{|c|c|c|c|c|c|c|c|}
        \hline
         \multirow{2}{*}{Signal}
       & \multirow{2}{*}{Multiwavelet}
       & \multicolumn{2}{c|}{$\epsilon$}\\
       \cline{3-8}
       & & $\sigma = 10$ & $\sigma = 15$ \\
        \cline{5-8}
          \hline
         \hline
         \multirow{7}{*}{\rotatebox{90}{\tt Piece-Polynomial}}
         & GHM    & 28.60 & 41.67\\
		 & SA4    & 28.36 & 41.54\\
         & CL     & 31.47 & 46.49\\
         & Alpert & 69.66 & 74.91\\
         & Integer Haar & 29.52 & 36.00\\
         & New1  & {\bf 26.32} & {\bf 31.80}\\
         & New2  & {\bf 27.34} & 36.89\\
         \hline
         \multirow{7}{*}{\rotatebox{90}{\tt Cusp}}
         & GHM      & 0.0900 & 0.0887\\
		 & SA4      & 0.0974 & 0.1086\\
         & CL       & 0.0905 & 0.0919\\
         & Alpert   & 0.0495 & 0.0534\\
         & Integer Haar & 0.0562 & 0.0687\\
         & New1     & {\bf 0.0440} & {\bf 0.0506}\\
         & New2     & {\bf 0.0398} & {\bf 0.0464}\\
         \hline
         \multirow{7}{*}{\rotatebox{90}{\tt HeaviSine}}
         & GHM    & 0.7560 & 0.8731\\
		 & SA4    & 0.6690 & 0.8544\\
         & CL     & 0.7089 & 0.8032\\
         & Alpert & 1.1430 & 1.2355\\
         & Integer Haar & 0.5114 & 0.7404\\
         & New1   & 0.5496 & 0.7658\\
         & New2   & {\bf 0.4250} & {\bf 0.6944}\\
         \hline
       \end{tabular}
    }
\end{table}
\paragraph{2D denoising}

Image denoising is the process of removing noise from a noisy image
while preserving the structure of the original image.  The best way to
test the effect of noise on a standard grayscale image is to add AWGN.
We apply thresholding methods to suppress noise from grayscale images of
size $256\times 256$ and $512\times 512$ pixels shown in
Fig.\ref{fig:04} and Fig.\ref{fig:05} and the multiwavelet decomposition
for seven orthogonal multiwavelets GHM, SA4, CL, Alpert, Integer Haar,
New1 and New2. The AWGN has variances $\sigma$=10, 20.  In order to
suppress the noise coefficients in all test images, soft or hard vector
thresholding is applied at five decomposition levels. Best results are
shown in boldface.

Since a single one criterion is not enough to compare denoising quality,
four well-defined criteria are introduced: MSE, PSNR, SSIM, and MS-SSIM.

Table~\ref{tab:16} gives denoising results for the soft and hard
thresholds of a noisy {\tt Cameraman} image with $\sigma$=10. The SA4
multifilter has max PSNR, SSIM and MS-SSIM values for all measures only
for a soft threshold at the first decomposition level.  The New1
multifilter achieves highest PSNR, SSIM and MS-SSIM values from the
second to the fifth decomposition level for both thresholding methods,
while the New2 multifilter achieves only for the hard thresholding
method.
 
Table~\ref{tab:17} gives denoising results with the soft and hard
thresholds and $\sigma$=10 for the {\tt Couple} image.  The results
shows that for all decomposition levels the highest PSNR and SSIM values
are achieved by the SA4 multifilter with soft thresholding, while New1
achieves the highest MS-SSIM values. With hard thresholding, the New1
multifilter achieves the highest PSNR, SSIM and MS-SSIM values for all
decomposition levels.

Table~\ref{tab:18} gives denoising results with soft and hard
thresholding and $\sigma$=10 for the {\tt Kiel} image.  The results show
that, except for the first level of decomposition, the multifilter New1
has highest PSNR, SSIM and MS-SSIM values for both thresholding methods.

Table~\ref{tab:19} gives denoising results with soft and hard
thresholding and $\sigma$=20 for the {\tt House} image.  The New1
multifilter achieves the highest PSNR, SSIM and MS-SSIM values for soft
thresholding from the second to fifth decomposition levels, as well as
for hard thresholding from the third to the fifth level .

Table~\ref{tab:20} gives denoising results with soft and hard
thresholding and $\sigma$=20 forthe {\tt Kiel} image.  The results show
that for soft thresholding the highest PSNR, SSIM, and MS-SSIM values
are achieved by New1 from the second to the fifth decomposition levels,
and for hard thresholding from the third to the fifth level.

From the results, we can see that in the case of small noise variance
$\sigma$=10 the New1 multifilter has the best denoising. Finally,
MS-SSIM, which generalizes SSIM over all decomposition levels, gives us
information that the New1 multiwavelet filter has better HVPs for image
quality and structure.
  
% Table 16
\begin{table}
\centering 
  \renewcommand\thetable{16}
  \caption{Comparative results of MSE,PSNR, SSIM and MS-SSIM of denoised image '{\tt Cameraman}' of size $256 \times 256$ pixels for seven orthogonal multiwavelets through $J=5$ decomposition levels with AWGN and variance $\sigma = 10$.}
  \label{tab:16}
{
\footnotesize
       \setlength\tabcolsep{6pt}
       \renewcommand{\arraystretch}{1.0}
    \begin{tabular}{|c|c|c|c|c|c|c|c|}
      \hline
      \multirow{2}{*}{Image} &
      \multirow{2}{*}{Threshold} &
      Multi- &
      \multicolumn{5}{c|}{Decomposition Levels}\\
      \cline{4-8} 
      & & wavelet & 1 & 2 & 3 & 4 & 5 \\
      \hline
      \multirow{57}{*}{\parbox{1.5in}{\includegraphics[width=1.5in]{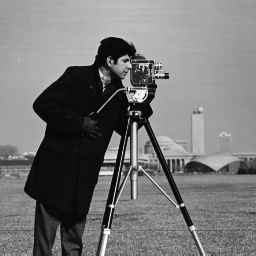}
         \\\centering {\tt Cameraman} }} & \multirow{28}{*}{Soft} &
                          \multirow{4}{*}{GHM} & 
                          71.08 & 72.02 & 75.95 & 78.16 & 78.99\\
                          & & & 29.61 & 29.56 & 29.33 & 29.2 & 29.16\\
                          & & & 0.7803 & 0.8450 & 0.8539 & 0.8542 & 0.8542\\
                          & & & 0.8077 & 0.8125 & 0.8101 & 0.8043 & 0.8026 \\ 
                          \cline{3-8}        
                         & & \multirow{4}{*} {SA4} & 
                         {\bf 66.5} & 70.00 & 74.23 & 76.64 & 77.24\\
                          & & & {\bf 29.90} & 29.68 & 29.42 & 29.29 & 29.25\\
                          & & & {\bf 0.7892} & 0.8495 & 0.8566 & 0.8563 & 0.8565\\
                          & & & {\bf 0.8025} & 0.8165 & 0.8149 & 0.7991 & 0.7960\\ 
                           \cline{3-8}        
                          & & \multirow{4}{*}{CL} & 
                          69.57 & 72.49 & 76.70 & 78.86 & 79.48\\
                          & & & 29.71 & 29.53 & 29.28 & 29.16 & 29.13\\
                          & & & 0.7851 & 0.8467 & 0.8553 & 0.8554 & 0.8558\\
                          & & & 0.8117 & 0.8100 & 0.8104 & 0.7972 & 0.7953\\ 
                           \cline{3-8}   
                          & &  \multirow{4}{*}{Alpert} &                          
                         129.73 & 212.88 & 318.79 & 431.91 & 550.35\\
                          & & & 27.00 & 24.85 & 23.10 & 21.78 & 20.72\\
                          & & & 0.7508 & 0.751 & 0.7010 & 0.6738 & 0.6599\\
                          & & & 0.7972 & 0.7809 & 0.7246 & 0.6410 & 0.5467 \\ 
                           \cline{3-8} 
                         & & \multirow{4}{*}{New1} &       
                         67.77 & {\bf 69.45} & {\bf 73.07} & {\bf 75.02} & {\bf 75.66}\\
                          & & & 29.82 & {\bf 29.71} & {\bf 29.49} & {\bf 29.38} & {\bf 29.34}\\
                          & & & 0.7855 & {\bf 0.8458} & {\bf 0.8552} & {\bf 0.8558} & {\bf 0.8559}\\
                          & & & 0.8013 & {\bf 0.8140} &{\bf 0.8277} & {\bf 0.8207} & {\bf 0.8168}\\ 
                           \cline{3-8} 
                        & &  \multirow{4}{*}{New2} &   
                        67.77 & 72.416 & 76.08 & 77.99 & 78.77\\
                          & & & 29.82 & 29.53 & 29.32 & 29.21 & 29.17\\
                          & & & 0.7855 & 0.8397 & 0.8494 & 0.8508 & 0.8513\\
                          & & & 0.8013 & 0.8043 & 0.8016 & 0.7870 & 0.7862\\ 
                            \cline{3-8} 
                         & &   \multirow{4}{*}{Integer Haar} & 
                        72.33 & 74.82 & 78.29 & 80.28 & 80.97\\
                          & & & 29.54 & 29.39 & 29.19 & 29.08 & 29.05\\
                          & & & 0.7784 & 0.8351 & 0.8451 & 0.8459 & 0.8451\\
                          & & & 0.8044 & 0.8174 & 0.8190 & 0.8088 & 0.8033\\ 
                             \cline{2-8} \\
                             \cline{2-8}
             & \multirow{28}{*}{hard} &
                          \multirow{4}{*}{GHM} & 
                          62.96 & 56.45 & 55.52 & 55.43 & 55.46\\
                          & & & 30.14 & 30.61 & 30.69 & 30.69 & 30.69\\
                          & & & 0.7733 & 0.8337 & 0.8430 & 0.8435 & 0.8435\\
                          & & & 0.8121 & 0.8270 & 0.8352 & 0.8374 & 0.8370\\ 
                          \cline{3-8}        
                         & & \multirow{4}{*} {SA4} & 
                           59.7 & 53.38 & 52.64 & 52.65 & 52.65\\
                          & & & 30.37 & 30.86 & 30.92 & 30.92 & 30.92\\
                          & & & 0.7806 & 0.8393 & 0.8463 & 0.8464 & 0.8464\\
                          & & & 0.8033 & 0.8285 & 0.8336 & 0.8326 & 0.8324\\ 
                           \cline{3-8}        
                          & & \multirow{4}{*}{CL} & 
                            59.84 & 53.12 & 52.14 & 52.06 & 52.07\\
                          & & & 30.36 & 30.88 & 30.96 & 30.97 & 30.97\\
                          & & & 0.7783 & 0.8390 & 0.8480 & 0.8483 & 0.8483\\
                          & & & 0.8170 & 0.8228	& 0.8311 & 0.8320 & 0.8318 \\ 
                          \cline{3-8} 
                         & &   \multirow{4}{*}{Alpert} &                          
                            95.12 & 146.3 & 228.29 & 326.6 & 435.01\\
                          & & & 28.35 & 26.48 & 24.55 & 22.99 & 21.75\\
                          & & & 0.7715 & 0.8003 & 0.7649 & 0.7392 & 0.7255\\
                          & & & 0.8004 & 0.8006 & 0.7708 & 0.7074 & 0.6176\\ 
                           \cline{3-8} 
                         & & \multirow{4}{*}{New1} & 
                          {\bf 59.42} & {\bf 52.45} & {\bf 51.35} & {\bf 51.36} & {\bf 51.36}\\
                          & & & {\bf 30.39} &{\bf  30.93} &{\bf  31.03} & {\bf 31.03} & {\bf 31.02}\\
                          & & & {\bf 0.7797} & {\bf 0.8388} &{\bf  0.8491} & {\bf 0.8494} & {\bf 0.8493}\\
                          & & & {\bf 0.8028} & {\bf 0.8201} & {\bf 0.8383} & {\bf 0.8367} &{\bf  0.8364}\\ 
                          \cline{3-8}        
                         & & \multirow{4}{*} {New2} & 
                         {\bf 59.42} & {\bf 52.49} & {\bf 51.57} & {\bf 51.49} & {\bf 51.49}\\
                          & & & {\bf 30.39} & {\bf 30.93} & {\bf 31.01} & {\bf 31.01} & {\bf 31.01}\\
                          & & & {\bf 0.7797} & {\bf 0.8346} & {\bf 0.8449} & {\bf 0.8457} & {\bf 0.8457}\\
                          & & & {\bf 0.8028} & {\bf 0.8142} & {\bf 0.8207} & {\bf 0.8244} & {\bf 0.8241} \\ 
                           \cline{3-8}        
                          & & \multirow{4}{*}{Integer Haar} & 
                            61.24 & 54.26 & 53.17 & 53.21 & 53.2\\
                          & & & 30.26 & 30.79 & 30.87 & 30.87 & 30.87\\
                          & & & 0.7727 & 0.8292 & 0.8385 & 0.8389 & 0.8389\\
                          & & & 0.8075 & 0.8264 & 0.8342 & 0.8311 & 0.8311 \\ 
      \hline
     \end{tabular}
  }
\end{table}

% Table 17
\begin{table}
\centering 
  \renewcommand\thetable{17}
  \caption{Comparative results of MSE,PSNR, SSIM and MS-SSIM of denoised image '{\tt Couple}' 
   of size $256 \times 256$ pixels for seven orthogonal multiwavelets through $J=5$ 
   decomposition levels with AWGN and variance $\sigma$=10.}
  \label{tab:17}
{
\footnotesize
       \setlength\tabcolsep{6pt}
       \renewcommand{\arraystretch}{1.0}
    \begin{tabular}{|c|c|c|c|c|c|c|c|}
      \hline
      \multirow{2}{*}{Image} &
      \multirow{2}{*}{Threshold} &
      Multi- &
      \multicolumn{5}{c|}{Decomposition Levels}\\
      \cline{4-8} 
      & & wavelet & 1 & 2 & 3 & 4 & 5 \\
           \hline
            \multirow{57}{*}{\parbox{1.5in}{\includegraphics[width=1.5in]{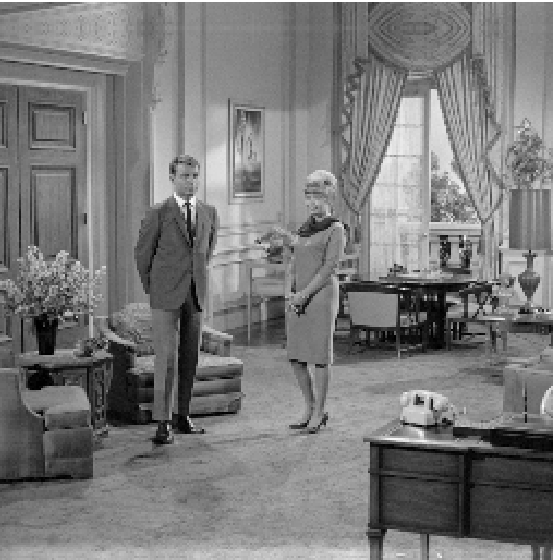}
         \\\centering {\tt Couple} }} & \multirow{28}{*}{Soft} &
                          \multirow{4}{*}{GHM} & 
                          92.29 & 100.35 & 107.19 & 110.23 & 111.19\\
                          & & & 28.48 & 28.12 & 27.83 & 27.71 & 27.67\\
                          & & & 0.8162 & 0.8049 & 0.7955 & 0.7936 & 0.7935\\
                          & & & 0.9488 & 0.9370 & 0.9238 & 0.9181 & 0.9181 \\ 
                          \cline{4-8}        
                         & & \multirow{4}{*} {SA4} & 
                         {\bf 87.91} & {\bf 98.62} & {\bf 105.69} & {\bf 108.22} & {\bf 109.12}\\
                          & & & {\bf 28.69} & {\bf 28.19} & {\bf 27.89} & {\bf 27.79} & {\bf 27.75}\\
                          & & & 0.8206 & {\bf 0.8060} & {\bf 0.7967} & {\bf 0.7953} & {\bf 0.7951}\\
                          & & & 0.9505 & 0.9398 & 0.9275 & 0.9241 & 0.9236\\ 
                           \cline{4-8}        
                          & & \multirow{4}{*}{CL} & 
                          91.44 & 102.99 & 110.09 & 112.76 & 113.7\\
                          & & & 28.52 & 28.00 & 27.71 & 27.61 & 27.57\\
                          & & & 0.8169 & 0.7998 & 0.7895 & 0.7885 & 0.7883\\
                          & & & 0.9460 & 0.9339 & 0.9181 & 0.9133 & 0.9130\\ 
                           \cline{4-8}   
                         &   &   \multirow{4}{*}{Alpert} &                          
                           155.72 & 239.46 & 320.22 & 394.98 & 479.76\\
                          & & & 26.21 & 24.34 & 23.08 & 22.17 & 21.32\\
                          & & & 0.7496 & 0.6423 & 0.5611 & 0.5394 & 0.5347\\
                          & & & 0.9435 & 0.8982 & 0.7995 & 0.7082 & 0.6526 \\ 
                           \cline{4-8} 
                         & & \multirow{4}{*}{New1} &       
                         89.19 & 100.26 & 106.99 & 109.52 & 110.37\\
                          & & & 28.63 & 28.12 & 27.84 & 27.74 & 27.70\\
                          & & & {\bf 0.8164} & 0.8002 & 0.7904 & 0.7896 & 0.7895\\
                          & & & {\bf 0.9509} & {\bf 0.9425} & {\bf 0.9290} & {\bf 0.9245} & {\bf 0.9238}\\ 
                           \cline{4-8} 
                        & &  \multirow{4}{*}{New2} &   
                        89.19 & 106.46 & 113.41 & 116.14 & 117.09\\
                          & & & 28.63 & 27.86 & 27.58 & 27.48 & 27.45\\
                          & & & {\bf 0.8164} & 0.7885 & 0.7820 & 0.7828 & 0.7832\\
                          & & & {\bf 0.9509} & 0.9339 & 0.9162 & 0.9116 & 0.9113\\ 
                              \cline{4-8} 
                         & &   \multirow{4}{*}{Integer Haar} & 
                         95.04 & 107.04 & 114.19 & 116.87 & 117.73\\
                          & & & 28.35 & 27.84 & 27.55 & 27.45 & 27.42\\
                          & & & 0.8051 & 0.7864 & 0.7782 & 0.7775 & 0.7777\\
                          & & & 0.9493 & 0.9364 & 0.9197 & 0.9135 & 0.9130\\ 
                         \cline{2-8} \\
                             \cline{2-8}
             & \multirow{28}{*}{hard} &
                          \multirow{4}{*}{GHM} & 
                          77.31 & 75.38 & 75.63 & 75.66 & 75.66\\
                        & & & 29.25 & 29.36 & 29.34 & 29.34 & 29.34\\
                        & & & 0.8324 & 0.8384 & 0.8379 & 0.8378 & 0.8378\\
                        & & & 0.9542 & 0.9549 & 0.9539 & 0.9538 & 0.9538\\ 
                          \cline{4-8}        
                         & & \multirow{4}{*} {SA4} & 
                         {\bf 73.9} & 72.06 & 72.1 & 72.12 & 72.12\\
                        & & & {\bf 29.44} & 29.55 & 29.55 & 29.55 & 29.55\\
                        & & & {\bf 0.8366} & 0.8420 & 0.8421 & 0.8420 & 0.8420\\
                        & & & 0.9529 & 0.9547 & 0.9546 & 0.9545 & 0.9545\\ 
                           \cline{4-8}        
                          & & \multirow{4}{*}{CL} & 
                          74.33 & 72.47 & 72.65 & 72.68 & 72.68\\
                        & & & 29.42 & 29.53 & 29.52 & 29.52 & 29.52\\
                        & & & 0.8357 & 0.8408 & 0.8402 & 0.8402 & 0.8402\\
                        & & & 0.9550 & 0.9539 & 0.9522 & 0.9520 & 0.9520\\ 
                          \cline{4-8} 
                       & & \multirow{4}{*}{Alpert} &                          
                           109.94 & 155.68 & 210.01 & 269.04 & 344.7\\
                        & & & 27.72 & 26.21 & 24.91 & 23.83 & 22.76\\
                        & & & 0.7959 & 0.7447 & 0.6960 & 0.6792 & 0.6743\\
                        & & & 0.9488 & 0.9268 & 0.8716 & 0.8082 & 0.7615\\ 
                           \cline{4-8} 
                         & & \multirow{4}{*}{New1} & 
                        {\bf 73.91} & {\bf 71.85} & {\bf 71.88} &{\bf  71.893} & {\bf 71.89}\\
                        & & & {\bf 29.44} &{\bf  29.57} & {\bf 29.56} & {\bf 29.56} & {\bf 29.56}\\
                        & & & 0.8360 & {\bf 0.8423} & {\bf 0.8425} & {\bf 0.8424} & {\bf 0.8424}\\
                        & & & {\bf 0.9533} & {\bf 0.9547} &{\bf  0.9548} & {\bf 0.9548} & {\bf 0.9548}\\ 
                          \cline{4-8}        
                         & & \multirow{4}{*} {New2} & 
                         73.91 & 73.08 & 73.29 & 73.34 & 73.34\\
                        & & & 29.44 & 29.49 & 29.48 & 29.48 & 29.48\\
                        & & & 0.8360 & 0.8382 & 0.8384 & 0.8383 & 0.8383\\
                        & & & 0.9533 & 0.9521 & 0.9522 & 0.9521 & 0.9521\\ 
                           \cline{4-8}        
                          & & \multirow{4}{*}{Integer Haar} & 
                         76.616 & 74.57 & 74.75 & 74.78 & 74.78\\
                        & & & 29.29 & 29.4 & 29.39 & 29.39 & 29.39\\
                        & & & 0.8308 & 0.8361 & 0.8359 & 0.8358 & 0.8358\\
                        & & & 0.9541 & 0.9548 & 0.9540 & 0.9540 & 0.9540 \\ 
      \hline
     \end{tabular}
  }
\end{table}

% Table 18
\begin{table}
\centering 
  \renewcommand\thetable{18}
  \caption{Comparative results of MSE,PSNR, SSIM and MS-SSIM of denoised image '{\tt Kiel}' 
   of size $512 \times 512$ pixels for seven orthogonal multiwavelets through $J=5$ 
   decomposition levels with AWGN and variance $\sigma$=10.}
  \label{tab:18}
{
\footnotesize
       \setlength\tabcolsep{6pt}
       \renewcommand{\arraystretch}{1.0}
    \begin{tabular}{|c|c|c|c|c|c|c|c|}
      \hline
      \multirow{2}{*}{Image} &
      \multirow{2}{*}{Threshold} &
      Multi- &
      \multicolumn{5}{c|}{Decomposition Levels}\\
      \cline{4-8}
      & & wavelet & 1 & 2 & 3 & 4 & 5 \\
           \hline
          \multirow{57}{*}{\parbox{1.5in}{\includegraphics[width=1.5in]{kiel}
          \\\centering {\tt Kiel}}} & \multirow{28}{*}{Soft} &
                          \multirow{4}{*}{GHM} & 
                          105.02 & 114.96 & 121.17 & 123.68 & 124.52\\
                         & & & 27.92 & 27.53 & 27.3 & 27.21 & 27.18\\
                         & & & 0.9139 & 0.9247 & 0.9236 & 0.9221 & 0.9220\\
                         & & & 0.8155 & 0.8069 & 0.7988 & 0.7941 & 0.7924 \\ 
                          \cline{4-8}        
                         & & \multirow{4}{*} {SA4} & 
                         {\bf 102.14} & {\bf 114.43} & 120.9 & 123.5 & 124.3\\
                         & & &{\bf 28.04 }& {\bf 27.55 }& 27.31 & 27.21 & 27.19\\
                         & & & {\bf 0.9026} & {\bf 0.9271} & 0.9268 & 0.9250 & 0.9249\\
                         & & & {\bf 0.8183} & 0.8085 & 0.7996 & 0.7949 & 0.7922\\ 
                           \cline{4-8}        
                          & & \multirow{4}{*}{CL} & 
                         104.94 & 116.99 & 123.39 & 125.82 & 126.59\\
                         & & & 27.92 & 27.45 & 27.22 & 27.13 & 27.11\\
                         & & & 0.9178 & 0.9200 & 0.9175 & 0.9164 & 0.9162\\
                         & & & 0.8120 & 0.8042 & 0.7944 & 0.7924 & 0.7894\\ 
                           \cline{4-8}   
                           & &   \multirow{4}{*}{Alpert} &                          
                          248.68 & 369.09 & 470.63 & 555.71 & 630.52\\
                         & & & 24.17 & 22.46 & 21.40 & 20.68 & 20.13\\
                         & & & 0.9000 & 0.8484 & 0.7684 & 0.7140 & 0.6970\\
                         & & & 0.8081 & 0.7658 & 0.6911 & 0.6106 & 0.5539 \\ 
                           \cline{4-8} 
                         & & \multirow{4}{*}{New1} &       
                         103.02 & {\bf 114.26} & {\bf 120.64} &{\bf 122.91} & {\bf 123.66}\\
                         & & & 28.00 & {\bf 27.55} & {\bf 27.32} & {\bf 27.23} & {\bf 27.21}\\
                         & & & 0.9069 & 0.9263 & {\bf 0.9261} & {\bf 0.9252} & {\bf 0.9252}\\
                         & & & 0.8184 & {\bf 0.8121} & {\bf 0.8085} & {\bf 0.8095} & {\bf 0.8109}\\ 
                           \cline{4-8} 
                        & &  \multirow{4}{*}{New2} &   
                       103.02 & 118.5 & 124.74 & 127.12 & 127.87\\
                         & & & 28.00 & 27.39 & 27.17 & 27.09 & 27.06\\
                         & & & 0.9069 & 0.9192 & 0.9195 & 0.9192 & 0.9198\\
                         & & & 0.8184 & 0.8063 & 0.7977 & 0.7937 & 0.7878\\ 
                              \cline{4-8} 
                         & &   \multirow{4}{*}{Integer Haar} & 
                         107.76 & 119.93 & 126.47 & 128.7 & 129.46\\
                         & & & 27.81 & 27.34 & 27.11 & 27.03 & 27.01\\
                         & & & 0.9162 & 0.9227 & 0.9191 & 0.9181 & 0.9181\\
                         & & & 0.8164 & 0.8091 & 0.8058 & 0.8047 & 0.8057\\ 
                             \cline{2-8} \\
                             \cline{2-8}
                         & \multirow{28}{*}{hard} &
                         \multirow{4}{*}{GHM} & 
                         75.26 & 71.62 & 71.21 & 71.12 & 71.11\\
                         & & & 29.37 & 29.58 & 29.61 & 29.61 & 29.61\\
                         & & & 0.9209 & 0.9426 & 0.9504 & 0.9517 & 0.9518\\
                         & & & 0.8244 & 0.8296 & 0.8345 & 0.8384 & 0.8385\\ 
                          \cline{4-8}        
                         & & \multirow{4}{*} {SA4} & 
                         72.91 & 69.49 & 69.17 & 69.16 & 69.16\\
                         & & & 29.50 & 29.71 & 29.73 & 29.73 & 29.73\\
                         & & & 0.9032 & 0.9400 & 0.9482 & 0.9487 & 0.9487\\
                         & & & 0.8222 & 0.8291 & 0.8306 & 0.8312 & 0.8316\\ 
                           \cline{4-8}        
                          & & \multirow{4}{*}{CL} & 
                        73.15 & 69.15 & 68.64 & 68.57 & 68.59\\
                         & & & 29.49 & 29.73 & 29.76 & 29.77 & 29.77\\
                         & & & 0.9322 & 0.9416 & 0.9490 & 0.9502 & 0.9501\\
                         & & & 0.8257 & 0.8277 & 0.8308 & 0.8350 & 0.8341\\ 
                          \cline{4-8} 
                         & & \multirow{4}{*}{Alpert} &                          
                        167.83 & 241.1 & 314.31 & 384.3 & 451.25\\
                         & & & 25.88 & 24.31 & 23.16 & 22.28 & 21.59\\
                         & & & 0.9063 & 0.8913 & 0.8462 & 0.8069 & 0.7918\\
                         & & & 0.8147 & 0.7932 & 0.7522 & 0.6975 & 0.6503\\ 
                           \cline{4-8} 
                        & & \multirow{4}{*}{New1} & 
                        {\bf 72.757} & {\bf 68.89} & {\bf 68.35} & {\bf 68.26} & {\bf 68.24}\\
                         & & & {\bf 29.51} &{\bf  29.75} & {\bf 29.78} & {\bf 29.79} & 29.79\\
                         & & & {\bf 0.9088} & {\bf 0.9408} & {\bf 0.9503} & {\bf 0.9517} & {\bf 0.9518}\\
                         & & & {\bf 0.8221} & {\bf 0.8249} & {\bf 0.8323} & {\bf 0.8374} & {\bf 0.8379}\\ 
                          \cline{4-8}        
                       & & \multirow{4}{*} {New2} & 
                       72.76 & 69.58 & 69.00 & 68.91 & 68.9\\
                         & & & 29.51 & 29.71 & 29.74 & 29.75 & 29.75\\
                         & & & 0.9088 & 0.9384 & 0.9481 & 0.9496 & 0.9497\\
                         & & & 0.8221 & 0.8223	& 0.8270 & 0.8289 & 0.8283\\ 
                           \cline{4-8}        
                       & & \multirow{4}{*}{Integer Haar} & 
                        74.41 & 70.74 & 70.22 & 70.13 & 70.11\\
                         & & & 29.41 & 29.63 & 29.67 & 29.67 & 29.67\\
                         & & & 0.9234 & 0.9445 & 0.9512 & 0.9525 & 0.9526\\
                         & & & 0.8227 & 0.8272 & 0.8354 & 0.8384 & 0.8386\\ 
      \hline
     \end{tabular}{}
  }
\end{table}

% Table 19
\begin{table}
\centering 
  \renewcommand\thetable{19}
  \caption{Comparative results of MSE,PSNR, SSIM and MS-SSIM of denoised image '{\tt House}' 
  of size $256 \times 256$ pixels for seven orthogonal multiwavelets through $J=5$ 
   decomposition levels with AWGN and variance $\sigma$=20.}
  \label{tab:19}
{
\footnotesize
       \setlength\tabcolsep{6pt}
       \renewcommand{\arraystretch}{1.0}
    \begin{tabular}{|c|c|c|c|c|c|c|c|}
      \hline
      \multirow{2}{*}{Image} &
      \multirow{2}{*}{Threshold} &
       Multi- &
      \multicolumn{5}{c|}{Decomposition Levels}\\
      \cline{4-8} 
       & & wavelet & 1 & 2 & 3 & 4 & 5 \\
           \hline
           \multirow{57}{*}{\parbox{1.5in}{\includegraphics[width=1.5in]{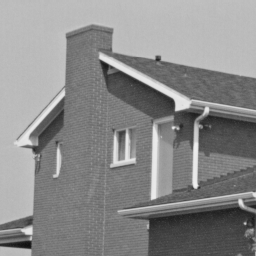}
           \\\centering {\tt House} }} & \multirow{28}{*}{Soft} &
                          \multirow{4}{*}{GHM} & 
                          142.45 & 96.18 & 96.29 & 101.75 & 104.51\\
                         & & & 26.59 & 28.30 & 28.30 & 28.06 & 27.94\\
                         & & & 0.5531 & 0.7296 & 0.7803 & 0.7856 & 0.7872\\
                         & & & 0.6539 & 0.6569 & 0.6459 & 0.6264 & 0.6176\\ 
                          \cline{4-8}        
                         & & \multirow{4}{*} {SA4} & 
                         {\bf 130.61} & 87.73 & 90.28 & 95.79 & 98.45\\
                         & & & {\bf 26.97} & 28.70 & 28.57 & 28.32 & 28.20\\
                         & & & {\bf 0.5651} & {\bf 0.7388} & 0.7842 & 0.7878 & 0.7887\\
                         & & & 0.6466 & 0.6643 & 0.6473 & 0.6351 & 0.6320\\ 
                           \cline{4-8}        
                          & & \multirow{4}{*}{CL} & 
                         138.84	& 100.21 & 99.7 & 105.1 & 107.66\\
                         & & & 26.71 & 28.12 & 28.14 & 27.91 & 27.81\\
                         & & & 0.5577 & 0.7272 & 0.7799 & 0.7862 & 0.7875\\
                         & & & {\bf 0.6554} & 0.6531 & 0.6477 & 0.6297 & 0.6172\\ 
                           \cline{4-8}   
                         &    &   \multirow{4}{*}{Alpert} &                          
                          141.88 & 132.45 & 200.02 & 283.49 & 384.37\\
                         & & & 26.61 & 26.91 & 25.12 & 23.61 & 22.28\\
                         & & & 0.5528 & 0.6944 & 0.6912 & 0.6868 & 0.6927\\
                         & & & 0.6480 & 0.6505 & 0.5989 & 0.5288 & 0.4256 \\ 
                           \cline{4-8} 
                         & & \multirow{4}{*}{New1} &       
                         133.63 & {\bf 87.54} & {\bf 85.925} & {\bf 89.234} & {\bf 91.48}\\
                         & & & 26.87 & {\bf 28.71} & {\bf 28.79} & {\bf 28.63} & {\bf 28.52}\\
                         & & & 0.5597 & 0.7335 & {\bf 0.7837} & {\bf 0.7917} & {\bf 0.7932}\\
                         & & & 0.6476 & {\bf 0.6644} & {\bf 0.6758} & {\bf 0.6801} & {\bf 0.6740}\\ 
                           \cline{4-8} 
                        & &  \multirow{4}{*}{New2} &   
                       133.63 & 100.62 & 101.43 & 105.88 & 108.33\\
                         & & & 26.87 & 28.10 & 28.07 & 27.88 & 27.78\\
                         & & & 0.5597 & 0.7095 & 0.7612 & 0.7754 & 0.7771\\
                         & & & 0.6476 & 0.6520 & 0.6508 & 0.6391 & 0.6247\\ 
                              \cline{4-8} 
                         & &   \multirow{4}{*}{Integer Haar} & 
                         143.67 & 101.04 & 100.53 & 104.21 & 106.35\\
                         & & & 26.56 & 28.09 & 28.11 & 27.95 & 27.86\\
                         & & & 0.5475 & 0.7072 & 0.7585 & 0.7668 & 0.7706\\
                         & & & 0.6517 & 0.6625 & 0.6673 & 0.6627 & 0.6537\\ 
                              \cline{2-8} \\
                             \cline{2-8}
                        & \multirow{28}{*}{hard} & 
                         \multirow{4}{*}{GHM} & 
                         166.77 & 118.57 & 110.57 & 109.79 & 109.77\\
                         & & & 25.91 & 27.39 & 27.69 & 27.73 & 27.73\\
                         & & & 0.5216 & 0.6514 & 0.6826 & 0.6850 & 0.6850\\
                         & & & 0.6563 & 0.6721 & 0.6790 & 0.6796 & 0.6801\\ 
                          \cline{4-8}        
                         & & \multirow{4}{*} {SA4} & 
                         156.48 & 109.25 & 102.48 & 102.17 & 102.2\\
                         & & & 26.19 & 27.75 & 28.02 & 28.04 & 28.04\\
                         & & & 0.5336 & 0.6620 & 0.6876 & 0.6884 & 0.6884\\
                         & & & 0.6457 & 0.6739 & 0.6786 & 0.6792 & 0.6795\\ 
                           \cline{4-8}        
                        & & \multirow{4}{*}{CL} & 
                        161.72 & 113.67 & 105.08 & 104.22 & 104.28\\
                         & & & 26.04 & 27.57 & 27.92 & 27.95 & 27.95\\
                         & & & 0.5262 & 0.6575 & 0.6907 & 0.6930 & 0.6930\\
                         & & & 0.6589 & 0.6681 & 0.6772 & 0.6768 & 0.6767\\ 
                          \cline{4-8} 
                        &   & \multirow{4}{*}{Alpert} &                          
                        {\bf 137.42} & {\bf 101.72} & 139.34 & 200.79 & 284.32\\
                         & & & {\bf 26.75} & {\bf 28.06} & 26.69 & 25.10 & 23.59\\
                         & & & {\bf 0.5559} & {\bf 0.721} & 0.7417 & 0.7385 & 0.7405\\
                         & & & {\bf 0.6485} & {\bf 0.6607} & 0.6396 & 0.5943 & 0.5093\\ 
                           \cline{4-8} 
                        & & \multirow{4}{*}{New1} & 
                        157.12 & 107.16 & {\bf 99.04} & {\bf 98.1} & {\bf 98.07}\\
                         & & & 26.17 & 27.83 & {\bf 28.17} & {\bf 28.21} & {\bf 28.22}\\
                         & & & 0.5307 & 0.6669 & {\bf 0.6996} & {\bf 0.7034} & {\bf 0.7035}\\
                         & & & 0.6463 & 0.6640 & {\bf 0.6785} & {\bf 0.6861}& {\bf 0.6870}\\ 
                          \cline{4-8}        
                       & & \multirow{4}{*} {New2} & 
                       157.12 & 113.38 & 106.22 & 105.33 & 105.27\\
                         & & & 26.17 & 27.59 & 27.87 & 27.91 & 27.91\\
                         & & & 0.5307 & 0.6502 & 0.6859 & 0.6915 & 0.6918\\
                         & & & 0.6463 & 0.6579 & 0.6674 & 0.6665 & 0.6646\\ 
                           \cline{4-8}        
                       & & \multirow{4}{*}{Integer Haar} & 
                       163.3 & 115.52 & 107.55 & 106.88 & 106.86\\
                         & & & 26.00 & 27.50 & 27.81 & 27.84 & 27.84\\
                         & & & 0.5218 & 0.6471 & 0.6795 & 0.6830 & 0.6831\\
                         & & & 0.6509 & 0.6662 & 0.6805 & 0.6826 & 0.6806\\ 
      \hline
     \end{tabular}{}
  }
\end{table}

% Table 20
\begin{table}
\centering 
  \renewcommand\thetable{20}
  \caption{Comparative results of MSE,PSNR, SSIM and MS-SSIM of denoised image '{\tt Kiel}' 
  of size $512 \times 512$ pixels for seven orthogonal multiwavelets through $J=5$ 
  decomposition levels with AWGN and variance $\sigma$=20.}
  \label{tab:20}
{
\footnotesize
       \setlength\tabcolsep{6pt}
       \renewcommand{\arraystretch}{1.0}
     \begin{tabular}{|c|c|c|c|c|c|c|c|}
      \hline
      \multirow{2}{*}{Image} &
      \multirow{2}{*}{Threshold} &
       Multi- &
      \multicolumn{5}{c|}{Decomposition Levels}\\
       \cline{4-8}
       & & wavelet & 1 & 2 & 3 & 4 & 5 \\
           \hline
       \multirow{57}{*}{\parbox{1.5in}{\includegraphics[width=1.5in]{kiel}
         \\\centering {\tt Kiel}}} & \multirow{28}{*}{Soft} &
                          \multirow{4}{*}{GHM} & 
                          272.71 & 268.66 & 281.78 & 289.61 & 292.65\\
                         & & & 23.77 & 23.84 & 23.63 & 23.51 & 23.47\\
                         & & & 0.7821 & 0.8254 & 0.8385 & 0.8360 & 0.8356\\
                         & & & 0.7432 & 0.7304 & 0.7120 & 0.6992 & 0.6950\\ 
                          \cline{4-8}   
                         & & \multirow{4}{*} {SA4} & 
                         {\bf 261.74}& {\bf 264.28} & {\bf 278.39 }&{\bf  286.46} & {\bf 289.35}\\
                         & & & {\bf 23.95} & {\bf 23.91} & {\bf 23.68} &{\bf  23.56} & {\bf 23.52}\\
                         & & & 0.7550 & {\bf 0.8282} & {\bf 0.8447} & {\bf 0.8416} & {\bf 0.8413}\\
                         & & & 0.7447 & 0.7331 & 0.7145 & 0.7010 & 0.7007\\ 
                           \cline{4-8}        
                          & & \multirow{4}{*}{CL} & 
                         269.97 & 271.63 & 285.83 & 293.45 & 296.15\\
                         & & & 23.82 & 23.79 & 23.57 & 23.46 & 23.42\\
                         & & & 0.8041 & 0.8211 & 0.8283 & 0.8275 & 0.8273\\
                         & & & 0.7408 & 0.7282 & 0.7063 & 0.6973 & 0.6955\\ 
                           \cline{4-8}   
                        & & \multirow{4}{*}{Alpert} &                          
                          397.34 & 507.77 & 625.36 & 723.45 & 805.9\\
                         & & & 22.14 & 21.07 & 20.17 & 19.54 & 19.07\\
                         & & & 0.7625 & 0.7572 & 0.6895 & 0.6297 & 0.6121\\
                         & & & 0.7376 & 0.6973 & 0.6087 & 0.5148 & 0.4615 \\ 
                           \cline{4-8} 
                         & & \multirow{4}{*}{New1} &       
                         266.38 & 266.5 & 281.11 & 288.16 & 290.8\\
                         & & & 23.88 & 23.87 & 23.64 & 23.53 & 23.49\\
                         & & & 0.7635 & 0.8250 & 0.8398 & 0.8394 & 0.8398\\
                         & & & 0.7448 & 0.7374 & 0.7212 & 0.7160 & 0.7216\\ 
                           \cline{4-8} 
                        & &  \multirow{4}{*}{New2} &   
                        266.38 & 282.67 & 296.43 & 303.67 & 306.29\\
                         & & & 23.88 & 23.62 & 23.41 & 23.31 & 23.27\\
                         & & & {\bf 0.7635} & 0.8061 & 0.8221 & 0.8246 & 0.8269\\
                       & & & {\bf 0.7448} & {\bf 0.7269} & {\bf 0.7024} & {\bf 0.6943} & {\bf 0.6920}\\ 
                        \cline{4-8} 
                         & &   \multirow{4}{*}{Integer Haar} & 
                         279.29 & 281.94 & 297.11 & 304.12 & 306.81\\
                         & & & 23.67 & 23.63 & 23.40 & 23.30 & 23.26\\
                         & & & 0.7870 & 0.8229 & 0.8262 & 0.8249 & 0.8254\\
                         & & & 0.7435 & 0.7312 & 0.7130 & 0.7051 & 0.7106\\ 
                              \cline{2-8} \\
                             \cline{2-8}
             & \multirow{28}{*}{hard} &
                         \multirow{4}{*}{GHM} & 
                         242.43 & 215.75 & 212.99 & 212.68 & 212.67\\
                         & & & 24.28 & 24.79 & 24.85 & 24.85 & 24.85\\
                         & & & 0.7892 & 0.8453 & 0.8722 & 0.8773 & 0.8775\\
                         & & & 0.7518 & 0.7577 & 0.7605 & 0.7642 & 0.7644\\ 
                          \cline{4-8}        
                         & & \multirow{4}{*} {SA4} & 
                         232.79 & 206.91 & 204.37 & 204.5 & 204.49\\
                         & & & 24.46 & 24.97 & 25.03 & 25.02 & 25.02\\
                         & & & 0.7550 & 0.8418 & 0.8710 & 0.8738 & 0.8740\\
                         & & & 0.7479 & 0.7572 & 0.7577 & 0.7588 & 0.7597\\ 
                           \cline{4-8}        
                        & & \multirow{4}{*}{CL} & 
                        235.51 & 207.86 & 204.95 & 204.51 & 204.5\\
                         & & & 24.41 & 24.95 & 25.01 & 25.02 & 25.02\\
                         & & & 0.8170 & 0.8441 & 0.8682 & 0.8734 & 0.8734\\
                         & & & 0.7536 & 0.7549 & 0.7555 & 0.7610 & 0.7611\\ 
                          \cline{4-8} 
                        &    & \multirow{4}{*}{Alpert} &                          
                        303.59 & 344.75 & 413.51 & 482.18 & 548.84\\
                         & & & 23.31 & 22.76 & 21.97 & 21.30 & 20.74\\
                         & & & 0.7678 & 0.8029 & 0.7818 & 0.7474 & 0.7329\\
                         & & & 0.7428 & 0.7245 & 0.6792 & 0.6277 & 0.5912\\ 
                           \cline{4-8} 
                        & & \multirow{4}{*}{New1} & 
                        {\bf 232.7} & {\bf 205.71} & {\bf 202.83} & {\bf 202.36} & {\bf 202.26}\\
                       & & & {\bf 24.46} & {\bf 25.00}& {\bf 25.06} & {\bf 25.07} & {\bf 25.07}\\
                       & & & {\bf 0.7650} & {\bf 0.8412}& {\bf 0.8717} & {\bf 0.8780} & {\bf 0.8786}\\
                       & & & {\bf 0.7485} & {\bf 0.7533} & {\bf 0.7575} & {\bf 0.7658} & {\bf 0.7698}\\                          \cline{4-8}        
                       & & \multirow{4}{*} {New2} & 
                       {\bf 232.7} & 211.29 & 207.95 & 207.64 & 207.57\\
                         & & & {\bf 24.46} & 24.88 & 24.95 & 24.96 & 24.96\\
                         & & & {\bf 0.7650} & 0.8313 & 0.8623 & 0.8686 & 0.8694\\
                         & & & {\bf 0.7485} & 0.7482 & 0.7509 & 0.7556 & 0.7551\\ 
                           \cline{4-8}        
                       & & \multirow{4}{*}{Integer Haar} & 
                       239.53 & 213.74 & 210.94 & 210.72 & 210.63\\
                         & & & 24.34 & 24.83 & 24.89 & 24.89 & 24.90\\
                         & & & 0.7947 & 0.8507 & 0.8729 & 0.8778 & 0.8784\\
                         & & & 0.7504 & 0.7550 & 0.7594 & 0.7645 & 0.7689\\ 
      \hline
     \end{tabular}
  }
\end{table}
% ================================================
\section{Conclusions}

Two new orthogonal bases for supercompact multiscaling functions with
exact recursion coefficients are obtained from the CL matrix product
filter by the fast Bauer's method of matrix spectral factorization. The
complementary supercompact multiwavelet functions are also found by QR
decomposition. The two new multifilters, even with very short support,
show relatively good CG performance, and better Sobolev smoothness than
other well-known orthogonal and biorthogonal multiwavelets.
	
Considering the highest MSE, PSNR, SSIM, and MS-SSIM(in bold) for CR at
all decomposition levels for image compression and denoising, the new
supercompact multiwavelet New1 (in same case also New2) achieves better
performance than the other multiwavelet filters.  The measure MS-SSIM
tells us that by applying New1 multifilters, better HVPs for image
quality and structure can be achieved in some cases.
	
From comparison with the well-known orthogonal Alpert, GHM, SA4, CL and
Integer Haar multiwavelets, we can conclude that the multifilters New1
and New2 can provide more effective noise suppression of some
one-dimensional signals, can be also useful in image compression and
subband-based edge detection applications as well.
		
In the future, these new supercompact bases could be used for solving
other different mathematical and engineering applications.
	
\section{Conflict of interest}
The author declares that there is no conflict of interest.

\section{Data availability statement}
Data available on request from the authors.

\section{Acknowledgments}
The authors would like to thank the editor, two anonymous reviewers for their careful reading, their constructive and valuable comments and very helpful suggestions which helped us to improve the manuscript.
% BibTeX users please use one of
%\bibliographystyle{spbasic}
\bibliographystyle{elsarticle-num}
\bibliography{sc}   % name your BibTeX data base
\end{document}